\pdfoutput=1
\documentclass{article}
\usepackage{iclr2027_conference,times}
\usepackage{amsmath,amssymb,array,booktabs}
\usepackage{fvextra}
\usepackage{longtable,needspace}
\usepackage{placeins}
\usepackage{graphicx}
\usepackage{flafter}
\usepackage{xcolor}
\usepackage{colortbl}
\usepackage[hypertexnames=false]{hyperref}
\usepackage{url}
\usepackage{float}

\newcommand{\HeadlineRoutes}{10}
\newcommand{\ScaleRoutes}{80}
\newcommand{\StressRoutes}{5}

\title{LongBusBench: Benchmarking LLM Agents for Long-Horizon Bus Control}

\author{Ye Hu \quad Jiawei Wang \\[3pt]
Thrust of Intelligent Transportation, Systems Hub, \\
The Hong Kong University of Science and Technology (Guangzhou) \\
\texttt{yeh@hkust-gz.edu.cn, jiaweiw@hkust-gz.edu.cn}
}

\iclrfinalcopy

\begin{document}
\setlength{\textfloatsep}{10pt plus 2pt minus 2pt}
\setlength{\floatsep}{8pt plus 2pt minus 2pt}

\maketitle
\lhead{Preprint.}

\begin{abstract}
Large language model (LLM) agents are increasingly capable of multi-step decision-making, yet their ability to sustain long-horizon control in real-world dynamic systems has received limited study. In public transit, control actions have delayed and interacting consequences: for example, bus holding changes vehicle headways and passenger queues, thereby altering the conditions encountered by subsequent vehicles and the future validity of past experience. We introduce LongBusBench, a data-grounded benchmark for evaluating LLM agents in long-horizon bus control under evolving operating conditions. We further propose MEMBus, a memory-augmented LLM agent that structures experience into episodic and procedural memories and selectively retrieves, consolidates, revises, and forgets information for transit control. Experiments across steady-state and shifting operating protocols compare MEMBus with memory-free LLM agents, alternative memory designs, and classical control baselines. Results show that operational memory improves public transit service, and that selective memory updating, suspension, and reactivation sustain control performance across shifting operating regimes. These results demonstrate how long-horizon public-transit control evaluates LLM agents in persistent operational systems. The benchmark will be open-sourced upon publication.
\end{abstract}

\section{Introduction}
\label{sec:introduction}
Large foundation models support general-purpose reasoning and decision-making across a wide range of tasks~\citep{bommasani2021foundation,wei2022emergent,openai2023gpt4}. Augmenting these models with reasoning, memory, and interaction with external environments yields large language model (LLM) agents~\citep{yao2023react,wang2023voyager,liu2024agentbench}. While LLM agents perform well in digital environments such as web navigation, software engineering, and knowledge-intensive tasks~\citep{yao2023react,zhou2024webarena,liu2024agentbench}, their capability in long-horizon operational systems that directly affect large passenger populations has received limited study.

Public-transit services operate over extended horizons under changing passenger demand, traffic conditions, and operational disruptions~\citep{wang2020dynamic,wang2021asynchronous}. Unlike episodic decision-making problems with environment resets~\citep{sutton2018reinforcement}, bus operations proceed without resets so that the consequences of an intervention propagate over extended periods. In bus bunching, a small delay increases passenger accumulation and dwell time, further delaying the affected bus while allowing following buses to approach with fewer passengers until local disturbances grow into fleet-level instability. Because holding actions reshape future operating conditions by modifying vehicle headways, passenger distributions, and subsequent decision contexts encountered by other buses, effective bus-control agents must reason over delayed consequences, coordinate across interacting vehicles, and adapt as operating conditions evolve. These dynamics differ from those in existing long-horizon agent tasks: current LLM-agent benchmarks mainly evaluate interactive tasks in digital environments~\citep{liu2024agentbench} or assess long-term memory over conversational histories~\citep{wu2025longmemeval}, whereas public-transit agents must maintain operational experience generated through their own previous interventions, identify when historical knowledge remains applicable under changing conditions, and preserve rare regime-specific experiences without allowing outdated information to degrade routine decisions.

Existing bus-control studies primarily optimize specialized reinforcement learning policies within episodic simulations under relatively stationary operating distributions~\citep{wang2020dynamic,wang2021asynchronous,tang2024robust}, whereas existing LLM-agent benchmarks do not measure downstream service outcomes in long-horizon physical systems, such as passenger waiting time, headway regularity, and fleet stability. This leaves open whether LLM agents can acquire, maintain, and apply long-term operational experience to control dynamic public transportation systems.

  To address these gaps, we first introduce a benchmark for long-horizon LLM-agent control of public-transit services, with bus-bunching mitigation as its central task. The benchmark evaluates service-level
  control performance alongside experience retention, adaptation to changing operating regimes, knowledge recovery when earlier regimes recur, and management of stale memories. We further introduce MEMBus, a memory-augmented LLM-agent framework that organizes operational experience into structured episodic and procedural memories and is designed to support selective retrieval,
  consolidation, revision, and forgetting according to estimated relevance, regime-dependent validity, and downstream control utility. Our main contributions are summarized as follows:

  \begin{itemize}
  \item \textbf{A new operational benchmark for long-horizon LLM agent control.}
  We formulate public transit management as a long-horizon agentic control problem, where decisions have delayed consequences, affect
  future operating states, and must be made under evolving service regimes.

  \item \textbf{A data-grounded testbed for long-horizon transit agents.}
  We introduce LongBusBench, a testbed built upon real stop sequences, road distances, and timetable headways that evaluates LLM agents in an evolving operational environment.

  \item \textbf{MEMBus: a memory-augmented bus control agent.}
We propose MEMBus, a memory framework designed for operational control. Unlike generic retrieval-based agent memory, MEMBus explicitly models context relevance, regime-dependent validity, and intervention utility to support long-horizon operational decision-making.

  \item \textbf{Empirical and diagnostic evaluation.}
We benchmark representative LLM agents, memory strategies, and classical
control baselines across steady-state and shifting operating conditions, paired with mechanism ablations quantifying how memory updating, suspension, and reactivation govern long-horizon control.

  \end{itemize}
  
 \section{Related Work}
  \label{sec:related-work}

\paragraph{LLM Agents for Traffic Control}

Recent studies have explored LLM agents for transportation decision-making. TrafficGPT connects natural-language queries with traffic-analysis tools for decision support~\citep{zhang2023trafficgpt}. LLMLight uses traffic observations to guide signal-phase selection with an LLM~\citep{lai2023llmlight}. CoLLMLight further investigates cooperative reasoning among multiple intersections with asynchronous LLM-based controllers~\citep{yuan2026collmlight}. In public transit, LLMs have been used as auxiliary components for bus control, such as designing reward functions for reinforcement learning controllers~\citep{yu2024llmbus}. However, existing approaches either focus on short-horizon control or use LLMs as supporting modules rather than autonomous agents. LongBusBench addresses this gap by evaluating whether LLM agents can perform long-horizon bus control through sustained interaction, experience reuse, and adaptation to evolving service conditions.

  \paragraph{Long-Horizon LLM Agents}
Long-horizon LLM agents increasingly rely on externalized memory to maintain information beyond a single interaction. ReAct interleaves reasoning, actions, and environmental observations to enable iterative decision making~\citep{yao2023react}, while Reflexion stores textual reflections from previous attempts to improve future behavior~\citep{shinn2023reflexion}. Generative
Agents extract useful memory from accumulated experiences~\citep{park2023generative}, and MemGPT manages information across memory tiers under limited context windows~\citep{packer2023memgpt}. Voyager further demonstrates lifelong skill acquisition by storing reusable skills during embodied exploration~\citep{wang2023voyager}. Recent long-horizon LLM-agent benchmarks evaluate whether agents can apply such long-term information. AgentBench assesses LLM agents across interactive digital environments~\citep{liu2024agentbench}, while LongMemEval evaluates memory retrieval, temporal reasoning, and knowledge updates over extended conversation histories~\citep{wu2025longmemeval}. LifelongAgentBench studies learning across interdependent tasks~\citep{zheng2025lifelongagentbench}, and Gaia2 considers asynchronous environments with temporal constraints~\citep{froger2026gaia2}. However, these benchmarks primarily focus on digital tasks or knowledge-based interactions, where the consequences of agent actions are often abstracted from physical system dynamics. LongBusBench extends long-horizon agent evaluation to a new operational problem.

\section{Preliminary}
  \label{sec:problem}

\begin{figure}[t]
\centering
\includegraphics[width=0.8\linewidth]{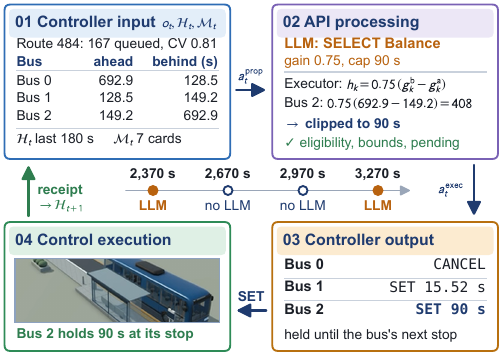}
\vspace{-6pt}
\caption{One logged MEMBus review (route 484, $t=2370$\,s) in the notation of Section~\ref{sec:problem}; the LLM is called only at reviews (orange). The Unity still is illustrative;
fields: Appendix~\ref{app:io-example}.}
\label{fig:worked-example-main}
\vspace{-8pt}
\end{figure}

  Consider a bus fleet $\mathcal{B}$ operating over a service horizon
  $[0,T]$. The state $s_t$ includes vehicle positions and passenger loads,
  stop queues, traffic conditions, operating resources, and the execution status of pending interventions.
   Controller $i$ receives observations authorized by the protocol and issues action proposals at decision times $\mathcal{T}_i$. These times need not coincide across controllers. We use one fleet-level controller ($i=1$). Let $r_j^i$ denote an observation, received message, prior action, or execution-feedback record that becomes available to controller $i$ at time $t_j^{i,\mathrm{avail}}$. Its available history is $\mathcal{H}_t^i = \bigl(r_j^i\bigr)_{j:\,t_j^{i,\mathrm{avail}}\leq t},$ with records ordered by availability time. Delayed feedback enters this history when it becomes available. Here $o_t^i$ is the latest observation-type record of $\mathcal H_t^i$.
  Controllers may condition on a selected subset of the available history $\mathcal{H}_t^i$ at each step. The retained memory $\mathcal{M}_t^i$ contains the experience maintained under the protocol's initialization and update rules. Given its latest available observation $o_t^i$, history $\mathcal{H}_t^i$, and retained memory $\mathcal{M}_t^i$, controller $i$ proposes \begin{equation}
  \mathbf{a}_t^{i,\mathrm{prop}}
  =
  \pi_i\!\left(
  o_t^i,\mathcal{H}_t^i,\mathcal{M}_t^i
  \right),
  \qquad t\in\mathcal{T}_i .
  \label{eq:prelim-proposal}
  \end{equation}
  Here $\pi_i$ includes the controller's selection and use of available
  information.
  Controllers may differ in their observation access, retained experience,
  and decision schedules.
A proposed holding command may be delayed, clipped, modified, or rejected by
operational constraints before execution.
  We therefore distinguish proposed actions from the joint fleet
  interventions that physically take effect,
  $\mathbf{a}_t^{\mathrm{exec}}$.
  The transition equation indexes consecutive simulation steps,
  whereas controller $i$ only acts at times in
  $\mathcal{T}_i$. Action execution accounts for inference delay: a proposal made at $t$ takes effect at the first executor tick at or after $t+d$, where $d$ is the answer delay, and its feedback enters $\mathcal H_t^i$ later (Appendix~\ref{app:async}).
  Under operating regime $z_t$ and exogenous variation $\xi_t$,
  the state evolves as:
  \begin{equation}
  s_{t+1}
  \sim
  P\!\left(
  \,\cdot\mid
  s_t,\mathbf{a}_t^{\mathrm{exec}},z_t,\xi_t
  \right).
  \label{eq:prelim-dynamics}
  \end{equation}
The regime $z_t$ captures demand, traffic, and capacity conditions, and is
observable only when explicitly exposed by the protocol. Holding alters
departure times, subsequent headways, passenger accumulation, and thus the future observations of the fleet. Hence the
problem is long-horizon: later service states and operational experience depend on earlier holding decisions.
A regime change within a service trajectory does not reset either the service
state or the memory retained under the protocol.
Figure~\ref{fig:worked-example-main} shows a logged example.
The objective is to use holding interventions to maintain service regularity
and passenger service under operational constraints. We report passenger
waiting time, headway variability, and served fraction separately to expose
the resulting trade-offs.

\begin{figure}[ht]
  \centering
  \IfFileExists{picture/pic/longbusbench_illustrated_agentbench_style.png}{%
  \includegraphics[width=\linewidth]{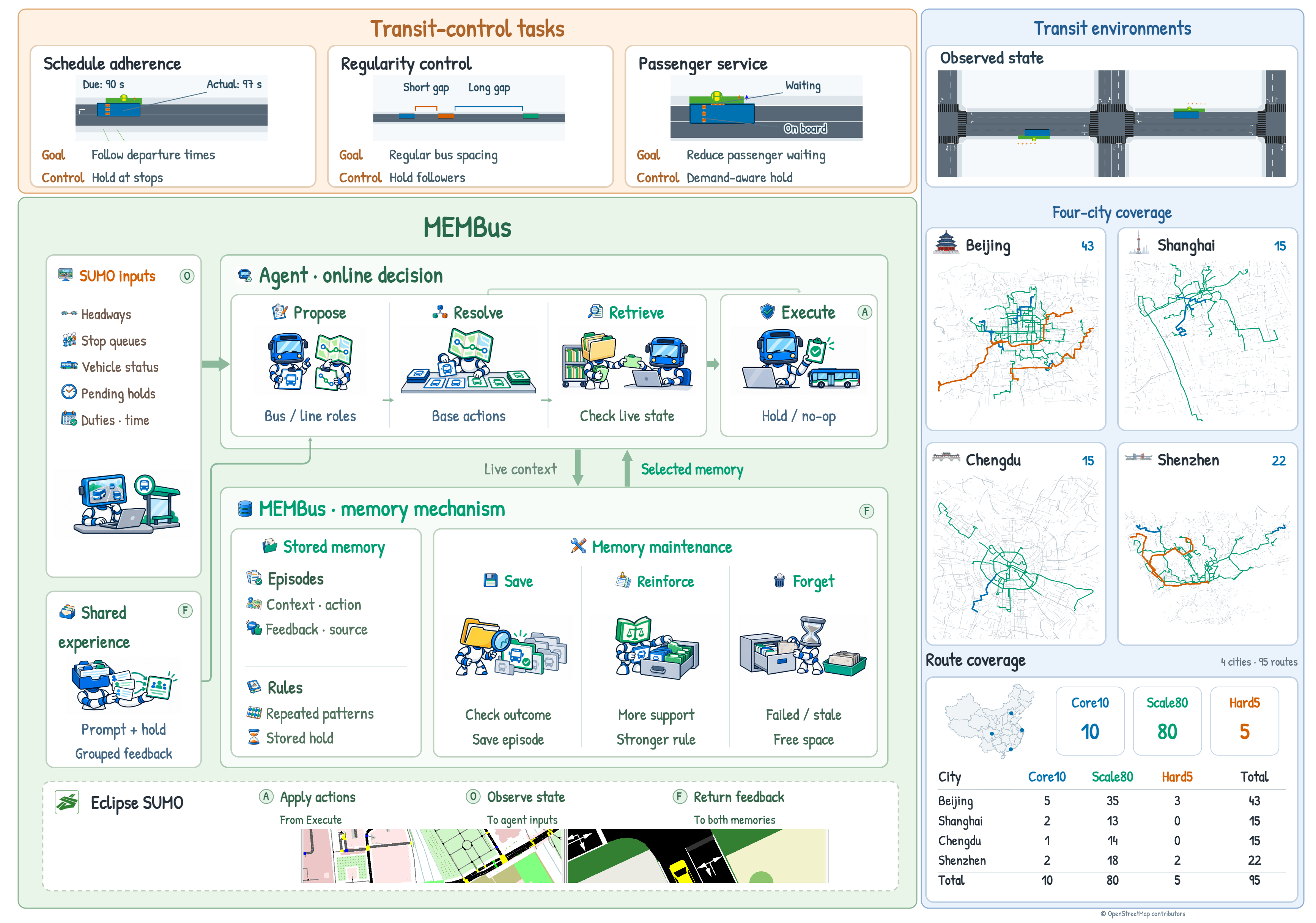}%
  }{%
  \fbox{\parbox[c][3cm][c]{0.94\linewidth}{\centering
  Overview figure placeholder\\[0.5em]
  \small The author-supplied figure will appear here.}}%
  }
  \vspace{-8pt}
  \caption{Overview of LongBusBench and the proposed MEMBus framework.}
  \label{fig:overview}
\end{figure}

  \section{LongBusBench}
  \label{sec:benchmark}
  LongBusBench combines real-route assets, simulated bus service,
  and a common agent interface.
  It supports long-horizon evaluation of operational history, changing service
  conditions, and recurring disruptions.

\subsection{Route and Demand Assets}
  The benchmark includes 95 bus routes from four major cities in China:
  Beijing, Shanghai, Shenzhen, and Chengdu (Fig.~\ref{fig:overview}). We use the vector route geometries and stop-level records from CPTOND~\citep{wang2026cptond} to construct stop sequences and inter-stop distances. We map each route onto real road networks, yielding 92 strictly validated network bindings and three reconstructed bindings. Detailed binding criteria and provenance are provided in Appendix~\ref{app:evidence-inventory}. We derive stop-level demand weights and conditional destination
  distributions from TransitLM route-planning records
  \citep{guo2026transitlm} to construct passenger-demand scenarios.
  Other operational parameters, including arrival rates, temporal
  demand profiles, fleet sizes, target headways, and traffic conditions,
  are specified separately in route and scenario configurations.
  Table~\ref{tab:routes} summarizes the route suites.

  \begin{table}[H]
    \centering
    \small
    \setlength{\tabcolsep}{4pt}
    \caption{Route suites and planning-demand evidence}
    \label{tab:routes}
    \begin{tabular}{@{}ccccc@{}}
    \toprule
    Suite & Size & Horizon & Stops
      & \shortstack{Mapped planning legs\\per stop}  \\
    \midrule
    Core10
      & \HeadlineRoutes{} routes
      & 16 h
      & 31
      & 42.44 \\
    Scale80
      & \ScaleRoutes{} routes
      & 16 h
      & 34
      & 61.40 \\
    Hard5
      & \StressRoutes{} routes
      & 16 h
      & 67
      & 41.79 \\

    \bottomrule
    \end{tabular}
  \end{table}
  
\subsection{Environment Composition}

The route collection defines the set of benchmark environments: we use SUMO~\citep{krajzewicz2012recent} to simulate traffic dynamics on OSM-derived road networks and couple it with a transit service layer that models passenger queues and vehicle states. 


\paragraph{Service dynamics.}
  Passenger demand is generated from the declared base rate,
  stop weights, and temporal profile.
  Depending on the service model, arrivals are accumulated as
  fractional mass or sampled as seeded Poisson counts.
  Passengers who cannot board remain in the queue.
  Boarding and alighting affect dwell time, while holding changes
  later departures, headways, and queues.
  Destination-based configurations use the declared destination
  distribution; other configurations use the specified fractional
  alighting rule.
  Detailed transition rules are provided in Appendix~\ref{app:service-dynamics}.

\paragraph{Agent interface.}
Agents interact with the environment through a unified API that provides
structured observations and receives control actions. Observations include the
current decision context, controllable vehicles, and available operational
information, while actions correspond to feasible holding decisions. We
support both event-driven control based on vehicle stop arrivals and
periodic fleet-level control based on system snapshots across different decision cadences. Execution details are provided in
Appendix~\ref{app:agent-interface}.

\paragraph{Benchmark scenarios and evaluation.}
Each benchmark instance specifies a route, fleet configuration, demand process,
disruption schedule, observation setting, and random seed. Additional history and recurrence protocols evaluate how agents apply prior experience under evolving conditions. Performance is measured using service-level metrics,
including passenger waiting time, headway regularity, and served demand, under
standardized evaluation windows.

  \section{MEMBus}
  \label{sec:methods}

  For this long-horizon setting, MEMBus uses historical interventions and corresponding outcomes to guide LLM-based bus control, as illustrated in Fig.~\ref{fig:overview}. It compiles related records into procedural cards and allows the controller to inspect the original evidence during decisions. A centralized LLM controller ($i=1$ in Section~\ref{sec:problem}, so we omit the controller superscript $i$ hereafter) uses this memory to generate holding plans for the fleet.

  \subsection{Memory Representation}

  An episodic record is represented by
  $e_k=(c_k,\mathbf{a}_k^{\mathrm{exec}},
  [t_k,t_k+\Delta_k],\mathbf y_k,p_k)$.
  Here $c_k$ is the pre-intervention context,
  $\mathbf{a}_k^{\mathrm{exec}}$ the executed fleet action, and $\mathbf y_k$
  the service measurements over the feedback interval.
  The provenance $p_k$ identifies the source, vehicles, stops,
  measurement protocol, and feedback availability.
  Paired records additionally specify the reference action
  and continuation policy. Then the procedural cards are compiled deterministically from records
  grouped by source conditions and comparison protocol.
  They summarize alternative actions and their outcomes,
  represent actions by vehicle roles, and retain links to
  supporting and adverse evidence.  The original records remain in an evidence archive indexed
  by a source catalog.
  Compilation requires no additional LLM call.

  \subsection{Retrieval and Execution}

  The query $c_t$ contains the current
  observation, recent history and execution feedback.
  Retrieval first removes records that are not yet available or
  whose cards are inactive.
  It then ranks the remaining cards by comparison-protocol match,
  agreement between the recorded and current operating conditions,
  and record recency.
  When a non-exact match is retained, the differing and unknown
  conditions are shown to the model.
  The LLM may request additional source records before generating
  a plan:
  \begin{equation}
  \begin{array}{@{}l@{\quad}c@{\quad}l@{}}
  \mathcal C_t
  & = &
  \operatorname{Retrieve}(\mathcal M_t,c_t),\\[4pt]
  \mathbf a_t^{\mathrm{prop}}
  & = &
  f_\theta(c_t,\mathcal C_t,\mathcal F_t),
  \end{array}
  \label{eq:memory-selection}
  \end{equation}
  where $\mathcal C_t$ contains the selected cards and
  $\mathcal F_t$ the additional source records.

  In the primary fleet interface, all seven route cards enter $\mathcal C_t$ directly ($\mathcal F_t=\varnothing$; two-stage $\mathcal F_t$ queries are evaluated in Appendix~\ref{app:membus-protocols}), and every 900\,s the controller selects one route program (Appendix~\ref{app:io-example}) from a menu of seven: no holding and six headway-balancing programs that differ in gain and cap. Each program has one card (with seven cards per route, $\mathcal C_t$ retains all active route cards without truncation), and the memory layer recommends the program with the lowest development waiting, subject to a served fraction no lower and a headway CV at most 5\% higher than without holding; the LLM may follow or override this recommendation. A deterministic executor turns the selected program into KEEP, SET, or CANCEL commands for the vehicles at each 300\,s tick.
  These operations preserve, replace, or clear the pending
  holding duration, respectively.
  The controller may revise a pending command before consumption.
  The executor checks vehicle eligibility and holding bounds
  and tracks accepted commands through consumption and dwell.
  Changing a pending command does not undo dwell already served.

  \subsection{Utility and Memory Updates}

  Candidate holding action $a$ and reference $a^0$ are compared from
  the same traffic state under a common feedback horizon
  $\Delta$, continuation policy $\pi^+$, and measurement
  protocol $\mathcal P$:
  \begin{equation}
  \Delta\mathbf y_{\mathcal P}
  (a,a^0\mid c;\Delta,\pi^+)
  =
  \bigl(
  W_a-W_{a^0},\,
  C_a-C_{a^0},\,
  S_a-S_{a^0}
  \bigr)^\top .
  \label{eq:intervention-utility}
  \end{equation}


  Here, $W$, $C$, and $S$ denote passenger waiting time, headway CV, and served
fraction, respectively. These outcomes are kept separate so that service trade-offs remain visible, and each
comparison retains the operating conditions under which it was observed.
When memory updates are enabled, newly available paired comparisons update action support in arrival order, counting each source trajectory once. Fixed rules classify paired comparisons as supporting, adverse, or neutral, while execution receipts and unpaired observations serve as contextual feedback. If a card's support reaches zero, it is suspended from active retrieval while remaining archived, leaving the underlying action legal and eligible for reactivation upon new supporting evidence. Deleting a procedural card removes the compiled summary while retaining raw archive records for reconstruction.

  \section{Experiments}
  \label{sec:protocol}

  All experiments use an 8\,h horizon. Supply uses real stop sequences, road distances, and timetable headways, which also determine route fleet sizes (Appendix~\ref{app:s11-construction}). Eight catalog routes form the main tables and 17 further routes serve as confirmation (Appendix~\ref{app:s11-construction}). Appendices~\ref{app:s11} and~\ref{app:final} document the primary timetable-grounded protocols and extended diagnostics, while Appendices~\ref{app:membus-protocols}--\ref{app:compute} report an auxiliary fixed-fleet per-bus suite.

    \subsection{Experimental Setup}

  MEMBus uses a centralized fleet controller based
  on \texttt{gpt-6-luna} with reasoning effort \texttt{none}; all LLM arms use this model.
  At each review the controller selects one route program in one call; an executor turns it into the fleet action plan at every decision (Section~\ref{sec:methods}). The plan uses the KEEP, SET, and
  CANCEL operations defined by the execution interface. Requested
  holding durations and total physical dwell are bounded by 90\,s;
  commands are also checked against vehicle eligibility and the
  pending-command execution rules. The current-observation control uses the same model, observations, action
  permissions, and decision schedule as MEMBus. Our framework exposes
  current vehicle states, pending commands, legal actions,
  and execution information, while in the current-observation control we remove the historical observations,
  procedural cards, source records, and queryable memory.
  This setup isolates the effect of MEMBus memory under an identical model and action space.

  \paragraph{Evaluation protocol.}
  Each main comparison uses eight routes and five seeds per route. Control begins at the case-specific time $t_0$ after warmup and lasts 28,800\,s: the controller makes 96 decisions at 300\,s intervals, and every third decision is a program review. Between decisions the simulation runs without further control; outcomes are scored over $[t_0, t_0+28800)$ and reported at the cumulative hours $h=1,2,4,8$ to show how
effects accumulate over the horizon.  The simulator pauses while the model answers; Section~\ref{sec:stale-experience} and Appendix~\ref{app:async} repeat the comparison with the clock running.

\paragraph{Metrics.}
Our primary metric is fixed-cohort restricted mean waiting time $W$. We additionally report the boarded (served) fraction $S$, the queue remaining at the horizon, and headway regularity measured by the coefficient of variation, $C\equiv\mathrm{CV}_{\mathrm{headway}}=s_g/\bar g$, where $s_g$ and $\bar g$ are the sample standard deviation and mean of pooled within-stop arrival intervals, respectively. Lower $C$ indicates more regular arrivals. Arrival intervals are computed between consecutive observed arrivals at the same stop and pooled over the scoring window. We also report token usage and
bunching-event counts, aggregated using the same route-weighted procedure.

\paragraph{Baselines.}
We compare MEMBus with eighteen baselines spanning four categories: LLM-based, memory-augmented LLM, learning-based, and traditional control methods.
\textbf{LLM-based baselines} include \textit{DiLu}, which retrieves prior examples without reflection~\citep{wen2024dilu}; \textit{ReAct}, which interleaves reasoning and tool use over the current bus state~\citep{yao2023react}; \textit{Recent-history}, which retains 180\,s of observations,
  actions, and execution feedback but excludes external experience and procedural memory; \textit{Self-Refine}, which performs one draft--feedback--refinement cycle per decision~\citep{madaan2023selfrefine}; and a decision-level \textit{Reflexion} adaptation that maintains a bounded buffer of verbal reflections~\citep{shinn2023reflexion}.
\textbf{Memory-augmented LLM baselines} (same development evidence as MEMBus) include \textit{RAG over records}~\citep{lewis2020rag}, \textit{ExpeL}~\citep{zhao2024expel}, \textit{Generative Agents}~\citep{park2023generative}, and \textit{MemGPT-style}~\citep{packer2023memgpt}; the MemGPT- and Voyager-style agents are re-implementations, with the differences listed in Appendix~\ref{app:vgmg}.
\textbf{Learning-based baselines} include \textit{DQN}~\citep{mnih2015human} and \textit{CQL}~\citep{kumar2020conservative}; four more are in Appendix~\ref{app:rl}.
\textbf{Rule-based baselines} include \textit{TwoWay}, a station-holding adaptation of cooperative two-way headway control~\citep{daganzo2011cooperation,xuan2011dynamic}, \textit{ForwardHeadway}~\citep{daganzo2009bunching}, \textit{BackwardHeadway}~\citep{bartholdi2012self}, and \textit{Headway feedback} (symmetric two-sided balance tuned on development seeds)~\citep{daganzo2009bunching,xuan2011dynamic}, together with a tuned forward slack-offset \textit{Daganzo rule}~\citep{daganzo2009bunching}, a per-route tuned rule over the 18-program stationary rule library, and \textit{Zero additional holding}, which removes control-induced holding while preserving passenger-service dwell.
All learned policies are frozen during evaluation; DQN and CQL are trained on the development seeds of each route. All methods share the same execution interface while retaining their specified action constraints and decision schedules. Additional implementation details are provided in Appendices~\ref{app:supporting-evaluations}, \ref{app:s11-sameif}, and~\ref{app:vgmg}.

  \subsection{Comparison with Baseline Controllers}
  \label{sec:baseline-comparison}

  \begin{table}[!t]
    \centering
    \caption{Controller comparisons. Each cell reports results at $h=1,2,4,8$.}
    \label{tab:baseline-comparison}
    \scriptsize
    \setlength{\tabcolsep}{2pt}
    \renewcommand{\arraystretch}{1.0}
    \begin{tabular}{@{}l@{\hspace{3pt}}rrrr@{\hspace{4pt}}rrrr@{\hspace{4pt}}rrrr@{\hspace{4pt}}rr@{}}
      \toprule
      & \multicolumn{4}{c}{Wait (s)$\downarrow$} & \multicolumn{4}{c}{CV$\downarrow$} & \multicolumn{4}{c}{Served (\%)$\uparrow$} & Tok. & Bunch. \\
      \cmidrule(lr){2-5}\cmidrule(lr){6-9}\cmidrule(lr){10-13}
      Controller & 1 & 2 & 4 & 8 & 1 & 2 & 4 & 8 & 1 & 2 & 4 & 8 & (M)$\downarrow$ & $\downarrow$ \\
      \midrule
      MEMBus & 582$^{*}$ & \textbf{550}$^{*}$ & \textbf{483}$^{*}$ & 425$^{*}$ & 0.65$^{*}$ & \textbf{0.55}$^{*}$ & 0.46$^{*}$ & 0.38$^{*}$ & \textbf{86.8}$^{*}$ & \textbf{94.5}$^{*}$ & \textbf{97.6}$^{*}$ & 98.8$^{*}$ & 1.94 & 18.0$^{*}$ \\
      \midrule
      \multicolumn{15}{@{}l}{\textit{LLM-based controllers}} \\
      DiLu & 633 & 757 & 923 & 1068 & 1.10 & 1.22 & 1.34 & 1.41 & 81.5 & 87.7 & 92.2 & 95.9 & 1.75 & 619.7 \\
      ReAct & 633 & 759 & 932 & 1072 & 1.09 & 1.21 & 1.34 & 1.41 & 81.4 & 87.5 & 92.2 & 95.9 & 6.93 & 617.2 \\
      Recent-history & 638 & 764 & 922 & 1064 & 1.10 & 1.23 & 1.34 & 1.41 & 81.3 & 87.4 & 92.4 & 95.9 & 3.80 & 625.8 \\
      Self-Refine & 631 & 755 & 922 & 1073 & 1.10 & 1.22 & 1.34 & 1.43 & 81.5 & 87.6 & 92.2 & 95.9 & 14.10 & 625.7 \\
      Reflexion & 632 & 757 & 920 & 1059 & 1.10 & 1.22 & 1.34 & 1.41 & 81.4 & 87.5 & 92.3 & 96.0 & 9.48 & 622.5 \\
      \multicolumn{15}{@{}l}{\textit{Memory-augmented LLM agents}} \\
      RAG over records & 618 & 699 & 810 & 901 & 0.93 & 1.00 & 1.10 & 1.15 & 83.1 & 89.3 & 93.6 & 96.6 & 1.38 & 444.4 \\
      ExpeL & 608 & 646 & 698 & 745 & 0.88 & 0.86 & 0.90 & 0.92 & 84.3 & 91.2 & 95.0 & 97.3 & 1.32 & 294.7 \\
      Generative Agents & 643 & 708 & 814 & 951 & 0.96 & 1.01 & 1.12 & 1.24 & 82.6 & 89.6 & 93.5 & 96.1 & 1.47 & 528.7 \\
      MemGPT-style & 632 & 760 & 930 & 1068 & 1.10 & 1.23 & 1.35 & 1.41 & 81.4 & 87.4 & 92.2 & 96.0 & \textbf{1.23}$^{*}$ & 623.1 \\
      \midrule
      \multicolumn{15}{@{}l}{\textit{Rule-based controllers}} \\
      TwoWay & 595 & 586 & 558 & 526 & 0.67 & 0.61 & 0.57 & 0.52 & 86.2 & 93.3 & 96.8 & 98.4 & \textemdash & 45.5 \\
      ForwardHeadway & 667 & 668 & 584 & 500 & 0.71 & 0.61 & 0.50 & 0.41 & 83.6 & 92.8 & 97.3 & 98.7 & \textemdash & 47.5 \\
      BackwardHeadway & 683 & 810 & 990 & 1167 & 1.07 & 1.16 & 1.33 & 1.54 & 79.8 & 86.9 & 91.9 & 95.4 & \textemdash & 557.7 \\
      Headway feedback (tuned) & 595 & 565 & 487 & \textbf{424} & \textbf{0.64} & \textbf{0.55} & \textbf{0.45} & \textbf{0.37} & 86.5 & 94.4 & \textbf{97.6} & \textbf{98.9} & \textemdash & \textbf{10.8} \\
      Daganzo rule (tuned) & 661 & 650 & 561 & 484 & 0.68 & 0.57 & 0.48 & 0.40 & 83.9 & 93.3 & 97.4 & 98.7 & \textemdash & 43.9 \\
      Per-route tuned rule (18 programs) & 617 & 602 & 530 & 448 & 0.65 & 0.56 & 0.46 & \textbf{0.37} & 85.7 & 93.4 & 97.5 & 98.8 & \textemdash & 17.6 \\
      Zero additional holding & 631 & 756 & 916 & 1066 & 1.10 & 1.22 & 1.35 & 1.43 & 81.5 & 87.6 & 92.3 & 95.9 & \textemdash & 621.0 \\
      \midrule
      \multicolumn{15}{@{}l}{\textit{Learning-based controllers}} \\
      DQN & 577 & 557 & 509 & 473 & 0.74 & 0.66 & 0.57 & 0.51 & 86.5 & 94.0 & 97.2 & 98.5 & \textemdash & 104.8 \\
      CQL (offline RL) & \textbf{575} & 565 & 519 & 461 & 0.76 & 0.71 & 0.62 & 0.53 & 86.3 & 93.6 & 97.2 & 98.7 & \textemdash & 126.1 \\
      \bottomrule
    \end{tabular}
    \par\smallskip{\footnotesize 8 routes $\times$ 5 seeds; columns 1/2/4/8 are hours $h$. Tok.: M per case. Bunch.: per case at $h=8$. Bold: best among all methods; $^{*}$: best among LLM-based methods.}
  \end{table}

\begin{figure}[t]
    \centering
    \includegraphics[width=\linewidth,trim=0 3 0 3,clip]{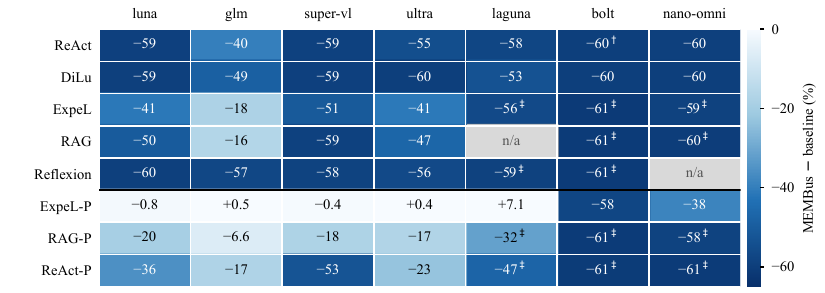}
    \vspace{-8pt}
    \caption{MEMBus minus baseline waiting (\%, 17 routes, $h=8$; darker: lower); rows below the line: same-interface agents. Gray: excluded (failed calls); $\dagger$: 16 routes; $\ddagger$: one seed; failure rates in Appendices~\ref{app:xmodel} and~\ref{app:xmodel-fig}.}
    \label{fig:heatmap}
  \end{figure}

\begin{figure}[t]
    \centering
    \includegraphics[width=\linewidth,trim=0 3 0 0,clip]{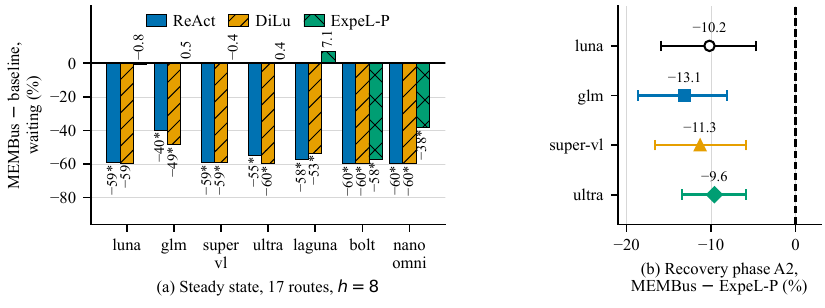}
    \vspace{-10pt}
    \caption{Cross-model comparison. (a) MEMBus minus baseline waiting (\%, 17 routes, $h=8$; negative: MEMBus lower); $*$: Holm-significant (\texttt{gpt-6-luna} DiLu reaches Holm $p=1.4{\times}10^{-4}$ in Table~\ref{tab:s11-main}, unstarred in panel a). (b) Recovery-phase gap to ExpeL-P with 95\% route-bootstrap intervals. MEMBus waits the same on every backbone because each model followed every correct recommendation (Appendix~\ref{app:xmodel}).}
    \label{fig:xmodel-main}
  \end{figure}

  The advantage grows with the horizon: against DiLu, MEMBus waits 8\% less at $h=1$ and 60\% less at $h=8$ (425 vs.\ 1068\,s), with fewer bunching events than every other LLM controller (Table~\ref{tab:baseline-comparison}). At $h=8$ it waits less than every learned baseline and rule in the table except the tuned headway-feedback rule, which is level (424 vs.\ 425\,s). Under the shared program interface (suffix -P; Appendix~\ref{app:s11-sameif}), MEMBus reduces waiting by 37\% relative to memory-free ReAct-P and 23\% relative to RAG-P, matches ExpeL-P in stationary service, and outperforms ExpeL-P when operating regimes shift (Section~\ref{sec:stale-experience}). MEMBus beats the nine original-interface LLM baselines on every service metric on all 8 routes (Appendix~\ref{app:permetric}) and is comparable to four offline-trained RL baselines (Appendix~\ref{app:rl}). On all seven backbones (17 routes) MEMBus waits less than DiLu and ReAct (Figures~\ref{fig:heatmap} and~\ref{fig:xmodel-main}).

  We next test each metric separately. Among original-interface LLM agents, MEMBus ranks first under all 286 multi-metric weightings (Appendices~\ref{app:permetric} and~\ref{app:xmodel-fig}).
  For metric $k\in\{W,C\}$ and baseline $X$, we report the route-mean relative difference
  \begin{equation}
  \delta_{k,X}=\frac{\bar Y_{k,\mathrm{MEMBus}}-\bar Y_{k,X}}{\bar Y_{k,X}},
  \label{eq:rel-diff}
  \end{equation}
  while served fraction $S$ is reported in percentage points as $\Delta S_X = \bar S_{\mathrm{MEMBus}} - \bar S_X$. Across all endpoints, we test paired route-level differences $d_{r,k,X} = \bar Y_{r,k,\mathrm{MEMBus}} - \bar Y_{r,k,X}$ with an exact two-sided route-level sign-flip test, Holm-corrected within each pre-specified hypothesis family.
  \subsection{Effect of Operational Memory}
  \label{sec:memory-effect}

  Full MEMBus has access to the fixed paired source records,
  procedural summaries, and experience acquired during its own
  trajectory.
  It can use these records when selecting actions and update the
  memory according to the lifecycle rules declared for the
  experiment.
The current-observation control has none of these sources,
  while retaining the same model, current-state inputs, action
  permissions, and execution interface; on these routes it keeps only the program descriptions and the current observation. At $h=8$ it waits 879\,s (CV 1.06, 96.8\% served) against 425\,s (0.38, 98.8\%) for MEMBus.
   MEMBus improves mean waiting, headway regularity, and served
  fraction over the current-observation LLM at every hour, and the gap widens with the horizon (20.5\% less waiting at $h=2$, 38.5\% at $h=4$; 8 of 8 routes at $h=8$; exact sign-flip protocol in Appendix~\ref{app:s11-stats}, fixed-fleet counterpart in Appendix~\ref{app:oldtb}). Zero additional holding waits 1066\,s (Table~\ref{tab:baseline-comparison}), confirming that the gain stems from selective holding rather than reduced holding volume.
  In the matched case of route 598 (Figures~\ref{fig:memory-trajectories} and~\ref{fig:memory-trajectories-8h}), memory-enabled
  control gives more evenly spaced trajectories
  and less bunching than zero additional holding.

  \subsection{Incremental Memory Ablations}
  \label{sec:membus-ablations}

  To connect short-window source-release switches ($\mathrm{B1},\mathrm{B12},\mathrm{B123}$ in the auxiliary 1.5\,h suite, Appendix~\ref{app:ablation-details}) with full-horizon lifecycle management on the 8\,h benchmark, we define for an endpoint $Y$
  \begin{equation}
  \Delta_{\mathrm{update}}=Y_{\mathrm{B12}}-Y_{\mathrm{B1}},\qquad
  \Delta_{\mathrm{exit}}=Y_{\mathrm{B123}}-Y_{\mathrm{B12}},\qquad
  \Delta_{\mathrm{mgmt}}=Y_{\mathrm{full}}-Y_{\mathrm{static}},
  \label{eq:ablation-increments}
  \end{equation}
  where $\Delta_{\mathrm{update}}$ and $\Delta_{\mathrm{exit}}$ are evaluated in Appendix~\ref{app:ablation-details} and $Y_{\mathrm{static}}$ denotes the same-interface arm with online lifecycle management disabled. On the 8\,h recurring-regime benchmark, Table~\ref{tab:ablation-main} evaluates an additive mechanism ladder from a DiLu-style baseline, a leave-one-out bundle contrast, and $\Delta_{\mathrm{mgmt}}$ across 25 unscreened routes (Appendix~\ref{app:stale}).

  \begin{table}[t]
    \centering
    \footnotesize
    \caption{Memory ablations (relative difference in waiting, \%; negative: lower waiting).}
    \label{tab:ablation-main}
    \setlength{\tabcolsep}{3pt}
    \begin{tabular}{@{}>{\raggedright\arraybackslash}p{0.47\linewidth}rrr@{}}
      \toprule
      Configuration & Wait (s) & Day \%   & A2 \%  \\
      \midrule
      \multicolumn{4}{@{}l}{\textit{Ladder from a DiLu-style agent (each row adds one mechanism)}} \\
      DiLu-style start & 701.4 & --- & --- \\
      + static cards, recommendation & 669.8 & $-$2.64  & --- \\
      + update & 646.3 & $-$3.29  & $-$1.82   \\
      + suspend & 637.7 & $-$0.99   & --- \\
      + reactivate & 630.8 & $-$1.12  & $-$7.12  \\
      + consolidate (full system) & 632.9 & $+$0.38  & --- \\
      \midrule
      Leave-one-out ($-$susp., $-$react., $-$cons.) & --- & $+$2.08  & $+$9.31  \\
      Full $-$ management off (25 routes) & --- & $-$4.07  & $-$6.65  \\
      \bottomrule
    \end{tabular}
\par\smallskip
{\footnotesize
Ladder: 16 routes $\times$ 5 seeds; percentage changes compare
with the preceding configuration.
Leave-one-out: removal of suspension, reactivation, and
consolidation relative to the full system.
Management: full system versus management off,
25 routes $\times$ 5 seeds.
A2 denotes the recovery phase.
Statistical tests and Holm-adjusted $p$-values are reported
in Appendix~\ref{app:stale}.
}
  \end{table}

  Updating drives the main whole-day reduction, whereas suspension and reactivation govern recovery when an earlier regime returns (percentages in Sections~\ref{sec:membus-ablations}--\ref{sec:stale-experience} report route means of per-route relative differences). Removing the suspension, reactivation, and consolidation bundle increases recovery-phase waiting by 9.31\% in the leave-one-out ablation across all 16 routes (Appendix~\ref{app:ladder}), and suspension and reactivation act complementarily in the $2^4$ factorial (Appendix~\ref{app:factorial}).

\subsection{Stale Experience}
  \label{sec:stale-experience}

\begin{figure}[t]
    \centering
    \includegraphics[width=\linewidth,trim=0 7 0 3,clip]{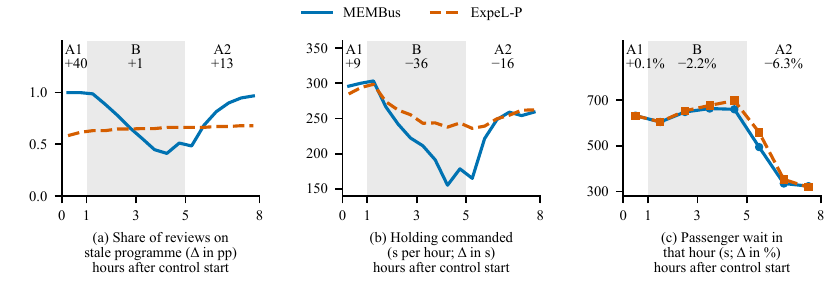}
    \vspace{-12pt}
    \caption{Stale-experience lifecycle (25 routes, \texttt{gpt-6-luna}; shaded: regime B); labels: MEMBus $-$ ExpeL-P per segment (Appendix~\ref{app:stale}).}
    \label{fig:regime-story}
  \end{figure}

  Over a long horizon the operating regime can change and later return within one episode. Each episode is A (1\,h), B (4\,h), A (3\,h), with windows
  \begin{equation}
  \mathcal{W}_{\mathrm{A1}}=[t_0,\,t_0{+}3600),\quad \mathcal{W}_{\mathrm{B}}=[t_0{+}3600,\,t_0{+}18000),\quad \mathcal{W}_{\mathrm{A2}}=[t_0{+}18000,\,t_0{+}28800),
  \label{eq:regime-windows}
  \end{equation}
  and the recovery-phase effect is computed on $\mathcal{W}_{\mathrm{A2}}$. In B the program stored for A is at least 10\% worse than the best alternative (stale) and applies again when A returns (Figure~\ref{fig:regime-story}). The change time is not told to the agent. On 25 routes not screened by any development rule, MEMBus waits 3.00\% less than a same-interface ExpeL-style agent over the day, 5.88\% less after A returns, and 6.65\% less in that phase than with management off.

MEMBus leaves the stored program after 2.47\,h against 3.75\,h for the ExpeL-style agent and, having left it, returns to it in A2 in 98\% of cases against 12\%. Equalizing total holding volume on the 11 lower-holding routes still leaves the ExpeL-style agent with 4.9\% higher whole-day waiting than MEMBus (Appendix~\ref{app:holding}). With the clock running and each answer applied after its measured delay (median 48--55\,s), MEMBus waits 3.05\% more than with the simulator paused (not significant) and the ordering is unchanged (Appendix~\ref{app:async}). With every answer applied one 300\,s tick late (97--98\% of delays) the recovery-phase gains remain: $-$7.95\% against the ExpeL-style agent and $-$7.87\% against management off (25 routes), and $-$10.1\% to $-$12.3\% on three open models (Appendix~\ref{app:async-data}).

\FloatBarrier
\section{Conclusion and Limitations}

\textbf{Conclusion.} We formulate bus control as a long-horizon LLM-agent problem in which past actions reshape future operating conditions and the value of experience changes over time. We introduce LongBusBench, a data-grounded benchmark for this setting, and MEMBus, a memory framework that organizes operational experience by relevance, regime validity, and intervention utility. MEMBus achieves lower passenger waiting than original-interface LLM baselines, with the gap growing over the evaluation horizon. Under a shared program interface, compiled procedural cards outperform memory-free and raw-retrieval baselines in stationary service, and online updating, suspension, and reactivation improve recovery when operating regimes shift and recur (Section~\ref{sec:stale-experience}). Together, these results show how selective experience management supports long-horizon LLM control in dynamic operational systems.

\textbf{Limitations.} Our evaluation operates in closed-loop SUMO simulation with planning-derived stop-demand weights, so on-road deployment requires calibration against agency ridership and automated vehicle location records.

\clearpage

\section*{Reproducibility Statement}
Section~\ref{sec:methods} describes the MEMBus memory representation,
retrieval process, action interface, and update rules.
Section~\ref{sec:protocol} specifies the primary timetable-grounded comparisons,
control schedules, and measurement windows, with statistical protocols detailed in Appendix~\ref{app:s11-stats}.
Appendices~\ref{app:evidence-inventory} and~\ref{app:execution-records} document the benchmark route assets and simulation execution interface.
Appendices~\ref{app:membus-protocols}, \ref{app:ablation-details}, \ref{app:supporting-evaluations}, \ref{app:comparison-design}, and~\ref{app:compute} provide auxiliary fixed-fleet per-bus protocols, nested ablations, baseline implementations, metric definitions, and compute accounting, while Appendices~\ref{app:s11} and~\ref{app:final} document the primary timetable-grounded benchmark, prompts, parameter tables, request/response samples, and diagnostic analyses (Appendices~\ref{app:io-example}, \ref{app:samples}, and~\ref{app:params}).

Reproduction requires the corresponding simulator and asset versions,
controller configuration, initial memory, random seeds,
and metric definitions.
Archived model responses support deterministic decision replay alongside live API replication checks (Appendix~\ref{app:s11-sameif}).

  \bibliographystyle{iclr2027_conference}
  \bibliography{final}

  \appendix
  \raggedbottom

  \textbf{Scope of the appendices.} Appendices~\ref{app:evidence-inventory} and~\ref{app:execution-records} document the 95-route LongBusBench asset catalog and SUMO simulation interface. Appendices~\ref{app:membus-protocols}--\ref{app:compute} specify the auxiliary 10-route fixed-fleet per-bus protocols, nested source-release ablations, baseline implementations, formal metric definitions, and compute accounting. Appendices~\ref{app:s11} and~\ref{app:final} provide full construction details, input/output schemas, parameter tables, multi-model evaluations, regime-shift ablations, simulator-perturbation checks, and running-clock latency analyses for the primary timetable-grounded benchmark of Section~\ref{sec:protocol}.

   \section{Route and Demand Assets}
  \label{app:evidence-inventory}

  \paragraph{Route suites.}

  LongBusBench contains 95 routes organized into three disjoint
  suites: Core10, Scale80, and Hard5, containing 10, 80, and
  5 routes, respectively.
  CPTOND provides the route and stop records used to construct
  stop sequences and inter-stop distances~\citep{wang2026cptond}.
  The mean numbers of stops per route are 30.6, 33.9375, and
  67.0; Table~\ref{tab:routes} reports these values rounded
  to integers.
  The listed 16-hour horizons describe the full-service
  benchmark protocols.
  The control and scoring windows of the MEMBus experiments
  are specified separately in
  Appendix~\ref{app:membus-protocols}.

  \paragraph{Road-binding categories.}

  The road-network catalog contains 92 strictly validated
  executable bindings and three reconstructed bindings.
  Strict validation checks stop accessibility on bus-permitted
  lanes, stop-to-road snapping distances of at most 60\,m,
  route-length deviation of at most 15\%, successful SUMO
  traversal, and zero teleports.
  Distance-constrained map matching additionally checks that
  the 90th-percentile segment-relative-distance error does
  not exceed 50\%.
  The reconstructed category permits snapping distances
  of up to 150\,m and route-length deviation of up to 20\%,
  subject to the corresponding matching and execution checks.

  The three reconstructed routes are Shanghai 71 interval
  (Fenghong Road--Huangpi North Road),
  Chengdu K12
  (Zhonghai International Station--Chengdu East Railway Station),
  and Shenzhen M359
  (Guanlan Dashuitian Terminal--Xinghe Shengjing Bus Terminal).
  All belong to Scale80.
  Core10 and Hard5 contain only strict bindings, while
  Scale80 contains 77 strict and three reconstructed bindings.
  Reconstructed bindings are available for coverage and
  sensitivity analyses and are excluded from the strict
  primary road-network evaluation.

  These bindings define executable paths on OSM-derived
  topology constrained by the source route records.

  \paragraph{Planning-demand evidence.}

  TransitLM route-planning records provide mapped public-transit
  legs for constructing relative stop-demand weights and
  conditional destination distributions~\citep{guo2026transitlm}.
  Each mapped leg identifies a boarding origin and a downstream
  destination on a selected route from public-transit route-planning queries.

  Let $n_{r,i}$ denote the number of mapped forward planning
  legs originating at stop $i$ on route $r$.
  The stored stop weight is
  \[
  w_{r,i}=n_{r,i}+1.
  \]
  The unit smoothing term assigns positive weight to stops
  without mapped origins.
  Weights are normalized to have a within-route mean of one:
  \[
  \widetilde w_{r,i}
  =
  \frac{w_{r,i}}
       {N_r^{-1}\sum_{j=1}^{N_r}w_{r,j}},
  \]
  where $N_r$ is the number of stops on route $r$.
  These weights specify relative demand across stops;
  absolute arrival rates and temporal profiles are configured
  separately.

  Table~\ref{tab:planning-demand-counts} lists the mapped-leg
  counts before smoothing.
  The final column, also reported in Table~\ref{tab:routes},
  is the total number of mapped legs divided by the number
  of route--stop entries in each suite.
  A physical stop served by multiple routes contributes
  a separate entry for each route.

  \begin{table}[t]
    \centering
    \small
    \caption{Mapped planning-demand evidence before smoothing.}
    \label{tab:planning-demand-counts}
    \begin{tabular}{@{}lrrr@{}}
      \toprule
      Suite & Route--stop entries & Mapped legs & Legs per stop \\
      \midrule
      Core10  & 306   & 12,988  & 42.44 \\
      Scale80 & 2,715 & 166,713 & 61.40 \\
      Hard5   & 335   & 14,001  & 41.79 \\
      \bottomrule
    \end{tabular}
  \end{table}

  Conditional destination distributions are constructed
  from mapped downstream destinations for each origin,
  without OD smoothing.
  Origins without an available downstream distribution use
  the synthetic fallback specified by the selected service
  configuration.

  \paragraph{Auxiliary fixed-fleet route subset.}
  Whereas the primary timetable-grounded benchmark (Section~\ref{sec:protocol}; Appendix~\ref{app:s11}) evaluates 25 catalog routes (8 main routes and 17 confirmation routes) sized from timetable AM-peak headways, the auxiliary fixed-fleet per-bus protocol (Appendices~\ref{app:membus-protocols}--\ref{app:supporting-evaluations}) evaluates a separate ten-route subset (Beijing 16, 135, \textit{Zhuan}183; Shanghai 17, 96, 152, 167; Chengdu 1; Shenzhen M401, M425). Two of these routes (16 and 96) overlap with the Core10 suite file (563, 598, 16, 306, M512, 96, 50, 546, M182, 484), six (17, 135, 152, 167, 1, M401) belong to Scale80, and two (\textit{Zhuan}183, M425) are auxiliary research reconstructions. Five routes (16, 152, \textit{Zhuan}183, 96, M425) use uniform stop weights, whereas the other five (17, 135, 167, 1, M401) use research-line stop lists and stop weights (19, 32, 19, 16, and 35 stops versus 21, 39, 22, 31, and 44 in the full catalog). On this fixed six-bus subset, MEMBus minus zero additional holding at $h=8$ is $-190.9$\,s on the five uniform-weight routes (mean wait under zero holding 678\,s) and $-11.9$\,s on the other five routes (4401\,s), reflecting differences in fleet saturation across the two subgroups (Appendix~\ref{app:oldtb}).

  \paragraph{Relation to the evaluated environments.}

  The benchmark distinguishes OSM-derived road-network
  bindings from generated route environments.
  Generated environments preserve route stop order and
  inter-stop distances in a service path with a return
  connection.
  The road-binding validation results above apply to the
  OSM-derived catalog.

  Catalog demand weights characterize the benchmark suites, whereas experiment-specific route geometries, stop weights, and service settings are specified in Section~\ref{sec:protocol} and Appendices~\ref{app:service-dynamics}, \ref{app:membus-protocols}, and~\ref{app:s11}.

  \section{Simulation and Control Interface}
  \label{app:execution-records}

  SUMO simulates vehicle movement~\citep{krajzewicz2012recent},
  while a separate service layer maintains passenger queues,
  boarding, alighting, onboard loads, and execution feedback.
  Each experiment specifies its network construction,
  arrival process, passenger-service model, and control interface.
  The benchmark supports multiple backend configurations;
  their observation and execution semantics are distinguished
  below.

  \subsection{Passenger Service and Cross-Day Operation}
  \label{app:service-dynamics}

  \paragraph{Passenger arrivals.}

  The demand model combines a base rate, normalized stop
  weights, a temporal profile, and disruption multipliers:
  \[
  \lambda_{r,i}(t)
  =
  \lambda_{0,r}\,
  \widetilde w_{r,i}\,
  m_r(t)\,
  d_{r,i}(t).
  \]
  Here $\lambda_{0,r}$ is measured in passengers per stop
  per second, $\widetilde w_{r,i}$ is the normalized stop
  weight defined in Appendix~\ref{app:evidence-inventory},
  $m_r(t)$ is the temporal multiplier, and $d_{r,i}(t)$
  is the disruption multiplier.
  An absent temporal profile or demand disruption contributes
  a multiplier of one.

  The fractional-arrival model accumulates
  $\lambda_{r,i}(t)\Delta t$ passenger mass at one-second
  intervals and retains fractional remainders.
  The public fleet backend instead uses seeded Poisson
  integer arrivals with the corresponding expected rate.
  Each protocol explicitly specifies its arrival process.

  \paragraph{Configuration used in the MEMBus experiments.}

  Both experimental suites use the fractional-arrival model: the primary 25-route timetable-grounded benchmark (Section~\ref{sec:protocol}; Appendix~\ref{app:s11}) sets each route's base rate from its AM-peak timetable headway as $\lambda_{0,r}=12.25/H_{\mathrm{am},r}$ passengers per stop per second ($0.0136$--$0.0357$), whereas the auxiliary 10-route fixed-fleet suite (Appendices~\ref{app:membus-protocols}--\ref{app:supporting-evaluations}) uses a uniform nominal base rate of $\lambda_{0,r}=0.05$ passengers per stop per second.
  Available stop weights are normalized within each route;
  reconstructed research instances use uniform weights.
  Alighting follows the configured fractional rule.
  These settings are distinct from the public fleet backend's
  Poisson arrival and directional-destination configuration.

  The planning-demand construction in
  Appendix~\ref{app:evidence-inventory}
  describes the benchmark catalog.
  Auxiliary protocols use their assigned stop weights and fractional alighting rules.
  Route assignments, source-memory conditions, and evaluation windows are specified in Section~\ref{sec:protocol} and Appendices~\ref{app:membus-protocols} and~\ref{app:s11}.

  \paragraph{Boarding, alighting, and dwell.}

  Passengers alight before boarding.
  Under finite capacity, passengers who cannot board remain
  in the stop queue.
  FIFO arrival records track waiting times and boarded
  passenger mass.
  Alighting follows either destination assignments or the
  configured fractional rule.
  The directional destination model restricts destinations
  to downstream stops and uses its specified fallback
  when destination evidence is unavailable.

  Natural dwell depends on boarding and alighting volumes,
  service times, crowding penalties, and dwell perturbations.
  At a normally served stop, ordinary pending-hold execution
  adds the control hold and any event-induced hold to
  natural dwell.
  The configured total-dwell limit caps the resulting
  duration, bounding the realized dwell increment.

  Boarding and alighting are updated at the handled arrival
  event.
  Passengers arriving during a hold board at the next arrival event.

  \paragraph{Cross-day extension.}

  The separate seven-day corridor protocol contains seven
  16-hour service windows and seven 8-hour depot windows,
  spanning 604,800\,s.
  Physical vehicle identity, block identity, cumulative trip
  counts, and battery state are carried across service days.

  In the terminal-clearing service configuration, trips and
  passenger accounting are closed before the next dispatch.
  Each service day begins with empty queues and terminal-positioned vehicles.
  Daily simulator instances may restart while the specified
  operational state is carried forward.

  The agent runtime supports retaining the same agent
  instance across days and delivering day-start and day-end
  events without an automatic agent reset.
  This cross-day capability is separate from the reported
  8\,h MEMBus evaluation, which covers sustained
  service within a single run.

  \subsection{Agent Interface and Execution}
  \label{app:agent-interface}

  \paragraph{Observations.}

  The API exchanges structured observations and action batches
  rather than simulator-owned objects.
  Observations identify the episode, decision, simulation
  time, visible vehicles, eligible vehicles, and available
  action capabilities.
  Backend-specific fields provide vehicle positions, loads,
  queues, pending commands, dwell status, and execution
  feedback.
  Unavailable common fields are represented as \texttt{null}.

  The road interface supplies decisions at vehicle stop
  events, whereas the fleet interface supplies periodic
  snapshots.
  Road headways are arrival-time gaps at shared stops;
  fleet control headways are derived from spatial separation.
  Evaluation headway CV is computed separately from recorded
  arrival intervals.
  Controllers receive only the fields authorized by their
  observation configuration.

  \paragraph{Action validation.}

  Let $\mathcal B_t^{\mathrm{cmd}}$ denote the vehicles
  receiving submitted holding commands and
  $\mathcal B_t^{\mathrm{elig}}$ the eligible set.
  For requested duration $h_{b,t}$ and permitted upper bound
  $\bar h_{b,t}$, valid holding commands satisfy
  \begin{equation}
  \begin{aligned}
  \mathcal B_t^{\mathrm{cmd}}
  &\subseteq \mathcal B_t^{\mathrm{elig}},\\
  0\leq h_{b,t}
  &\leq\bar h_{b,t},
  \qquad b\in\mathcal B_t^{\mathrm{cmd}}.
  \end{aligned}
  \label{eq:prelim-holding-bounds}
  \end{equation}

  Public API batches must match the current decision
  identifier and API version and contain only permitted
  fields with finite, in-range values.
  Invalid batches are rejected before backend execution.

  The public API and research controllers use their
  respective action contracts.
  The public API permits holding bounds of up to 30\,s,
  whereas the reported MEMBus research configuration uses
  a 90\,s command limit and a 90\,s total physical dwell
  limit.
  Vehicle-specific eligibility and holding bounds still
  apply within these limits.

  \paragraph{Pending commands.}

  In the MEMBus plan interface, KEEP preserves a pending
  command, SET replaces its duration, and CANCEL clears
  the pending duration.
  Where exact-zero normalization is enabled, a SET command
  with duration zero is interpreted as CANCEL; otherwise,
  it is handled by the configured plan validator.

  Under the research controller's ownership rules,
  cancellation retains ownership of the zero-duration
  command until consumption or replacement.
  The issuing controller may revise its pending command
  before consumption.
  Waiting for execution feedback does not by itself
  disable this permission.

  For ordinary fleet holding, a pending command is consumed
  at the next handled stop.
  Omitting a vehicle from an accepted backend action batch
  preserves its pending command, whereas an explicit zero
  clears the pending duration.
  A controller's response schema may nevertheless require
  an explicit operation for every listed vehicle.

  Road holding is scheduled at the current stop event.
  Separately enabled current-stop extensions and dwell
  reductions follow their own execution rules.
  No operation reverses dwell already served.

  Let $\mathbf q_t$ denote accepted commands awaiting
  execution.
  The execution interface maps proposals to physical
  interventions:
  \begin{equation}
  \left(
  \mathbf a_t^{\mathrm{exec}},\mathbf q_{t+1}
  \right)
  =
  \mathcal E_t
  \left(
  s_t,\mathbf q_t,\mathbf a_t^{\mathrm{prop}}
  \right).
  \label{eq:prelim-execution}
  \end{equation}
  Current-stop interventions need not pass through a
  next-stop command queue.

  \paragraph{Execution records and timing.}

  Execution records distinguish proposal validation,
  command acceptance, stop-level consumption, and observed
  departure.
  Command identifiers link subsequent feedback to the
  issuing decision.
  Realized dwell increments are recorded upon stop-level consumption.

  The evaluated simulation loop waits for model inference
  before advancing simulated time.
  Wall-clock inference latency is therefore accounted for
  as computational cost rather than simulated operating
  delay.
  Vehicle-event timing and controller-internal concurrency
  are distinct from advancing the simulation during
  inference.

\section{Auxiliary Fixed-Fleet Per-Bus Protocols}
  \label{app:membus-protocols}

  Alongside the primary 25-route timetable-grounded program-selection benchmark of Section~\ref{sec:protocol} (documented in Appendix~\ref{app:s11}), we document here the auxiliary centralized per-bus evaluation protocols on the ten-route fixed-fleet subset (1.5\,h short-window and 8\,h full-window comparisons, five seeds per route).
  MEMBus and the current-observation control each contribute
  50 completed episodes per protocol, yielding 50 matched
  pairs in each window.
  Case records associate each route and seed with its
  control-entry time, scoring boundaries, and result version.

  \subsection{Route and Seed Assignments}

  Table~\ref{tab:short-seed-plan} lists the route--seed
  assignments and corresponding control-entry times $t_0$ shared by both protocols.
  The short-window endpoint is $t_0+1800$\,s, and the
  8\,h endpoint is $t_0+28800$\,s.

  \begin{table}[!htbp]
    \centering
    \small
    \caption{Route--seed assignments (seed ranges) and control-entry times $t_0$ in seconds for both auxiliary fixed-fleet protocols.}
    \label{tab:short-seed-plan}
    \setlength{\tabcolsep}{4pt}
    \begin{tabular}{@{}llrrrrr@{}}
      \toprule
      Route & Seeds & $t_0$ seed 1 & seed 2 & seed 3 & seed 4 & seed 5 \\
      \midrule
      Beijing 16 & 2026120100--2026120104 & 2220 & 2220 & 2220 & 2220 & 2220 \\
      Shanghai 17 & 2026120110--2026120114 & 1830 & 1830 & 1830 & 1830 & 1830 \\
      Beijing 135 & 2026120120--2026120124 & 3180 & 3180 & 3360 & 3180 & 3180 \\
      Shanghai 152 & 2026120130--2026120134 & 1800 & 1800 & 1800 & 1800 & 1800 \\
      Shanghai 167 & 2026120140--2026120144 & 2490 & 2490 & 2490 & 2490 & 2490 \\
      Chengdu 1 & 2026120150--2026120154 & 2010 & 2040 & 1800 & 2010 & 1800 \\
      Shenzhen M401 & 2026120160--2026120164 & 3810 & 3810 & 3810 & 3810 & 3810 \\
      Beijing Special 183 & 2026120170--2026120174 & 1950 & 1950 & 1950 & 1950 & 1950 \\
      Shanghai 96 & 2026120180--2026120184 & 1830 & 1830 & 1830 & 1830 & 1920 \\
      Shenzhen M425 & 2026120190--2026120194 & 1800 & 1920 & 1830 & 1920 & 1920 \\
      \bottomrule
    \end{tabular}
  \par\smallskip\noindent\parbox{\linewidth}{\footnotesize Columns $t_0$ follow the five seeds in ascending order within each seed range.}
  \end{table}

  \subsection{Control and Scoring Windows}

  \paragraph{Warmup and control entry.}

  Before model evaluation, a numerical warmup determines
  the earliest permitted 30-second-aligned control-entry
  time $t_0$.
  The scan begins no earlier than 1800\,s and the
  route-specific minimum warmup.
  All six expected vehicles must be active, and every
  scheduled stop must have at least two arrivals recorded
  before the control boundary.
  This requirement concerns arrivals per stop, not per
  vehicle.
  Coverage is evaluated without inspecting candidate
  service outcomes.

  The resulting $t_0$ values range from 1800 to 3810\,s.
  Matched runs use the same route, seed, control-entry time,
  and numerical warmup.
  Each simulation is initialized independently rather than
  loaded from a shared SUMO snapshot.

  \paragraph{1.5\,h short-window protocol.}

  This protocol has 600\,s of LLM control and an 1800\,s scoring window; 1.5\,h denotes its longest simulated span including warmup (at most 93.5 minutes).
  The controllers act at
  $t_0,t_0+15,\ldots,t_0+585$\,s, giving 40 decisions over
  a 600-second control period.
  From $t_0+600$ to $t_0+1800$, the common numerical policy
  continues service while respecting outstanding LLM
  command ownership.
  The scoring interval is $[t_0,t_0+1800)$ and covers
  both control and numerical continuation. Including warmup, total simulated durations range from
  60 to 93.5 minutes.

  \paragraph{8\,h protocol.}

  Control and scoring both cover $[t_0,t_0+28800)$.
  MEMBus and the current-observation control have 96
  decision opportunities at 300-second intervals, with
  the final decision at $t_0+28500$.
  Simulation continues for the remaining 300\,s and then
  ends.
  No numerical control is inserted between these decisions,
  and no additional continuation phase follows scoring.

  Both protocols use the same route--seed set.
  The two protocols differ jointly in horizon, decision cadence, and continuation.

  \subsection{Environment and Action Settings}

  \paragraph{Route and passenger configuration.}

  The auxiliary experiments construct SUMO service networks
  directly from real-route stop sequences and
  inter-stop distances.

  Passenger demand accumulates as fractional mass at
  one-second intervals.
  In this auxiliary fixed-fleet suite, the nominal base rate is 0.05 passengers per stop per
  second, multiplied by the normalized stop weights (whereas the primary timetable-grounded benchmark uses $\lambda_{0,r}=12.25/H_{\mathrm{am},r}$; Appendix~\ref{app:s11-construction}).
  The research binding supplies no additional temporal
  demand profile or demand-surge multiplier.

  The normal passenger capacity is 70, reduced to 60
  during absolute simulation-time intervals
  720--1440\,s and 2160--2880\,s.
  These intervals are measured from simulation start,
  not from control entry.
  Their overlap with the scoring window therefore depends
  on $t_0$.
  At served stops, 25\% of onboard passenger mass alights
  before capacity-constrained boarding.

  \paragraph{Action execution.}

  The main LLM interface permits holding durations from
  0 to 90\,s per vehicle.
  SET replaces the pending duration for the next handled
  stop, and an exactly zero SET is normalized to CANCEL.
  KEEP preserves the pending command and its ownership.
  CANCEL clears its duration while retaining ownership
  until consumption or replacement.
  Total physical dwell is capped at 90\,s.

  The short-window numerical continuation resumes the
  specified two-way holding policy on its 60-second
  schedule.
  It respects unconsumed LLM-owned commands, including
  controlled zero-duration commands.
  Further execution semantics are given in
  Appendix~\ref{app:agent-interface}.

  \subsection{Current Observations and Memory Access}

  \paragraph{Program-selection versus two-stage per-bus interface.}
  In the primary program-selection architecture (Section~\ref{sec:methods}; Appendix~\ref{app:params}), MEMBus issues one program-selection call per review (every 5th tick at 1.5\,h, every 3rd tick at 8\,h). Its prompt contains the current observation, the 7-program menu of the route, the route's procedural cards, recent history, the last six reviews, and a constrained recommendation (served $\ge$ KEEP and CV $\le 1.05\times$ KEEP); the cards store paired evidence from disjoint development seeds (4 development seeds per route at 1.5\,h and 1 source seed per route at 8\,h on the auxiliary subset; 4 development seeds per route on the 25-route timetable-grounded benchmark), never target evaluation seeds. Target seeds share the route geometry and deterministic demand rate with the development seeds while varying dwell-noise realizations. By contrast, the subsections below document the two-stage per-bus evidence-selection variant used in the nested source-release ablations of Appendix~\ref{app:ablation-details}.

  \paragraph{Two-stage per-bus decision procedure.}

  In the two-stage per-bus variant, MEMBus executes an evidence-selection stage followed by
  action generation.
  The first stage may select up to two available evidence
  identifiers, including the option to select none.
  The second stage produces a complete fleet plan using
  the returned evidence.
  Evidence selection does not itself execute an action.

  Run records log per-call status, model responses, parsed decisions, token usage, retries, and fallback outcomes.
  Model and invocation versions are documented in
  Appendix~\ref{app:compute}.

  A failed evidence query is logged and returns an
  empty evidence set.
  An action-generation failure produces the all-KEEP
  fallback.
  Such failures remain part of the evaluated trajectories.

  \paragraph{Current-observation control.}

  The current-observation control receives current vehicle
  states, queues, pending commands, legal actions, and
  their execution semantics.
  Historical observations, past decisions, execution-history
  records, procedural cards, source evidence, and queryable
  runtime memory are excluded from its input.

  \paragraph{History and retrieval limits.}

  MEMBus additionally receives authorized recent history,
  external development evidence, and available experience
  from its own trajectory.
  The recent-history window is 180\,s, with observations
  sampled at 5-second intervals.
  Action records, execution feedback, and completed
  observational windows are included when available.

  The two-record query limit applies jointly to external
  and online evidence.
  The online view displays at most four procedures,
  with one case per procedure.
  External physical-source presentation is limited
  to eight rows.
  Memory context and returned archive records share an
  80,000-token budget.
  Both conditions have the same current-state observation
  fields and physical command permissions.
  Their observed values can differ after their trajectories
  diverge.
  Unexecuted numerical action recommendations are excluded
  from the memory-comparison inputs.

  \subsection{Source Evidence and Online Procedures}

  The auxiliary two-stage per-bus configuration supplies six paired external records that shorten or cancel pending holds on the fixed six-bus routes, whereas the primary benchmark compiles route-level program cards (Appendix~\ref{app:params}) and evaluates regime shifts on timetable-sized fleets (Appendix~\ref{app:stale}).

  \paragraph{External paired evidence.}

  The auxiliary per-bus memory configuration uses six paired development
  records from Route~152, seed~2026113020.
  They comprise three action comparisons from each of
  two source states at simulation times 3000 and 3120\,s.

  Each candidate changes one target vehicle's pending hold
  to zero, one-half, or three-quarters of its existing
  duration, while all other vehicles KEEP.
  The reference plan assigns KEEP to the entire fleet.
  Candidate and reference outcomes are compared over
  1800\,s under the recorded numerical continuation.

  Each record retains the candidate and reference plans,
  source conditions, feedback interval, outcomes, and
  provenance.
  These six completed development branches stem from two states of a single development run.

  \paragraph{Release and source-context exit.}

  Three records are available at control entry, and three
  additional records are scheduled for release after
  150\,s of control.
  After 300\,s, the initial source conditions leave active
  procedural presentation.
  Their underlying records remain queryable in the archive.

  At each decision, the memory update first releases
  newly available records and then applies the scheduled
  source-context exit.
  The updated view is constructed before evidence selection
  and action generation.
  Scheduled exit moves initial source conditions from active display to the archive.

  \paragraph{Online feedback.}

  Online acquisition links submitted plans to command
  consumption or non-consumption, concurrent interventions,
  and completed observation windows.
  Feedback uses predefined, non-overlapping 120-second
  windows beginning at $t_0-120$.
  Plans are assigned to windows by submission time;
  a separate window is not started after each action.

  Execution feedback is linked through \texttt{command\_id}.
  Records distinguish consumption within the window,
  later consumption, superseded commands, and commands
  whose consumption has not been observed.
  A record enters memory after its window closes and
  the corresponding feedback becomes available.
  Late receipts may produce a new version of the same
  plan--window record.

  \paragraph{Online procedural lifecycle.}

  Procedural records group experiences by their joint plan,
  route, reference plan, feedback horizon, continuation,
  measurement protocol, cohort, units, and evidence type.
  Source cases and record versions are retained.

  At a memory review, a procedure is archived after
  180\,s without an update.
  New matching evidence restores archived procedures to the active view.

  The displayed procedures are ordered by update recency,
  with procedure identifier used to resolve ties.
  At most four procedures and one case per procedure
  are displayed.

  \paragraph{Timing in the 8\,h protocol.}

  The 180-second recent-history and archival thresholds
  remain unchanged when decision spacing increases to
  300\,s. The previous decision's observation is outside the
  recent-history window at the next decision, although
  newer observations and newly available feedback may
  still be included. Online feedback windows remain 120\,s.
  External-evidence release and source-context exit are
  processed at decisions.
  Consequently, the 150-second release threshold is
  first handled at $t_0+300$ in the 8\,h protocol.

  \subsection{Case Identity and Result Reuse}

  Pairing uses route, seed, control-entry time, scoring
  boundaries, and the common numerical warmup.
  Passenger arrivals follow the deterministic demand
  schedule.
  Dwell perturbations are coupled by vehicle identity,
  stop identity, and visit number, using the noise seed
  from absolute simulation time 600\,s onward.
  Indexing perturbations by vehicle, stop, and visit aligns local dwell noise across paired runs.

  The recorded endpoint times and decision schedules
  match the specified protocols for all 200 episodes.
  Shared MEMBus endpoints across tables report the same
  underlying evaluation runs, subset comparisons use matched
  cases for both methods, and recorded-response replay
  reproduces trajectories under matching requests, prompts,
  and schemas.
 
  \section{Auxiliary Short-Window Source-Release Ablations}
  \label{app:ablation-details}

  We first compare three nested source-release configurations over 50 matched
  1.5\,h-window route--seed cases on the auxiliary 10-route subset.
  All configurations share the initial external-source
  procedures, paired evidence, action space, and numerical
  continuation policy.
  Online observational procedures and their archival and
  re-entry mechanisms are enabled in B12 and B123, whereas B1 isolates static initial source procedures without online procedure formation.
  Table~\ref{tab:ablation-switches} summarizes the configuration switches.

  \begin{table}[!htbp]
    \centering
    \small
    \caption{Mechanisms enabled in the nested short-window memory
    ablation.}
    \label{tab:ablation-switches}
    \begin{tabular}{@{}lccc@{}}
      \toprule
      Mechanism & B1 & B12 & B123 \\
      \midrule
      Initial source procedures and paired evidence
        & Yes & Yes & Yes \\
      Queryable newly released external evidence
        & Yes & Yes & Yes \\
      External-source procedure updates
        & No & Yes & Yes \\
      Initial source-context exit at 300\,s
        & No & No & Yes \\
      Online observational procedure formation/revision
        & No & Yes & Yes \\
      Online procedure archival and re-entry
        & No & Yes & Yes \\
      \bottomrule
    \end{tabular}
  \end{table}

  \paragraph{Configuration differences.}

  B1 keeps the initial external-source procedures fixed and runs without online procedure formation, archival, or re-entry (0 online procedures formed in B1 versus 398 in B12 and 400 in B123).
  The three additional external records, released after
  150\,s of control, enter the queryable archive in B1 but do
  not update active procedures.
  B12 enables external-source procedure updates from the newly released paired evidence together with online observational procedure formation, archival after 180\,s without an update, and re-entry when new matching evidence arrives.
  B123 further removes the initial source conditions from
  active presentation after 300\,s of control, while retaining
  their underlying records for querying.

  \paragraph{Contrast definitions.}

  For endpoint $Y$, the two incremental contrasts are
  \[
  \Delta_{\mathrm{update}}
  =
  Y_{\mathrm{B12}}-Y_{\mathrm{B1}},
  \qquad
  \Delta_{\mathrm{exit}}
  =
  Y_{\mathrm{B123}}-Y_{\mathrm{B12}}.
  \]
  Here $\Delta_{\mathrm{update}}$ measures the combined effect of external-source procedure updates and online procedural lifecycle activation.
  The second contrast $\Delta_{\mathrm{exit}}$ isolates the additional effect of removing
  the initial source conditions from active presentation, and
  $Y_{\mathrm{B123}}-Y_{\mathrm{B1}}$ measures their combined increment.

  \paragraph{Paired results.}

  Table~\ref{tab:ablation-paired-details} reports the paired
  contrasts computed from unrounded endpoints;
  displayed increments may differ by $0.01$ from the displayed combined contrast due to rounding.

  \begin{table}[!htbp]
    \centering
    \small
    \caption{Nested ablation contrasts over 50 matched
    short-window cases. Differences are the first configuration minus
    the second; served-fraction differences are in
    percentage points.}
    \label{tab:ablation-paired-details}
    \begin{tabular}{@{}lrrr@{}}
      \toprule
      Contrast & $\Delta W$ (s) 
        & $\Delta C$ & $\Delta S$ (pp) \\
      \midrule
      B12$-$B1
        & $-0.32$ 
        & $-0.02215$ & $+0.034$ \\
      B123$-$B12
        & $-1.06$ 
        & $+0.00979$ & $+0.169$ \\
      B123$-$B1
        & $-1.39$ 
        & $-0.01236$ & $+0.203$ \\
      \bottomrule
    \end{tabular}
  \end{table}

  Moving from B1 to B12 improves all three mean service
  metrics, and source-context exit in B123 further reduces mean waiting and
  increases served fraction with a small increase in headway CV $C$.
  B123 achieves the lowest mean waiting and highest served
  fraction, whereas B12 achieves the lowest $C$.
  Evaluated trajectories include all recorded fallbacks.

  \paragraph{Single-mechanism ablations on the timetable-grounded benchmark.}

  Because the nested configurations above bundle external-source updates with online procedural lifecycle activation on the short-window per-bus setting, Section~\ref{sec:membus-ablations} and Appendix~\ref{app:stale} evaluate individual lifecycle mechanisms on the 8\,h timetable-grounded benchmark starting from a DiLu-style baseline: a static-card control with experience management disabled (Appendix~\ref{app:ladder}), an additive ladder adding one mechanism at a time (Appendix~\ref{app:ladder}), a leave-one-out comparison (Appendix~\ref{app:ladder}), and a complete $2^4$ factorial over update, suspension, reactivation, and consolidation (Appendix~\ref{app:factorial}).

  \section{Baseline Implementations and Auxiliary Comparisons}
  \label{app:supporting-evaluations}

  This section documents the LLM-based, memory-augmented, rule-based, and learning-based controller implementations shared across our evaluations, together with extended controller and tool-use comparisons on the auxiliary ten-route fixed-fleet subset (ten routes $\times$ five seeds per route).

  Methods share an execution interface while retaining
  their specified observation access, action restrictions,
  and model-call schedules.
  Let $H$ denote the target headway, $f_b\equiv g_b^{\mathrm{ahead}}$ and $b_b\equiv g_b^{\mathrm{behind}}$ the
  forward and backward spatial headways of vehicle $b$, and
  $u_{b,t}$ the adapter-supplied headway-slack upper bound, which defines the permitted holding limit $\bar h_{b,t}=\max(0,\min(h_{\max},u_{b,t}))$ in Eq.~\eqref{eq:prelim-holding-bounds} with $h_{\max}=90$\,s.
  These inputs are derived from authorized fleet
  observations, whereas evaluation headway CV $C$ uses recorded stop-arrival intervals.

  \subsection{Shared Evaluation Protocols}

  Auxiliary comparisons in this section use the route--seed assignments and
  scoring boundaries specified in
  Appendix~\ref{app:membus-protocols}.
  The short-window protocol provides 40 execution
  opportunities over 600\,s, followed by 1200\,s of
  numerical continuation.
  The 8\,h protocol provides 96 execution opportunities
  at 300-second intervals without numerical continuation.

  Execution opportunities are distinct from fresh model
  calls.
  Retrieval queries, intermediate reasoning, refinement,
  reflection, and plan reuse are accounted for separately.
  Service outcomes use 50 cases per method and protocol (46 for RAG and 48 for ExpeL at 8\,h due to pairwise exclusion of failed runs; Appendix~\ref{app:compute}).
  Execution and token accounting rules are detailed in Appendices~\ref{app:comparison-design} and~\ref{app:compute}.

  \subsection{LLM-Based Baselines}

  \paragraph{DiLu.}

  The adaptation uses experience retrieval inspired by
  DiLu~\citep{wen2024dilu}.
  Current-state text is encoded with
  all-MiniLM-L6-v2 into a 384-dimensional vector.
  Cosine similarity selects three available development
  examples, with ties resolved by record identifier.
  The model receives the current state, the previous
  actual plan, and the retrieved examples.
  Reflection is disabled.

  Where a plan is reused, commands are aligned by vehicle
  identity and checked against current execution constraints.
  Departed vehicles are omitted and newly appearing vehicles
  receive KEEP.
  Plan reuse does not invoke the model.
  The retained call records contain 20 new-plan calls per
  short-window case and 48 per long-window case.

  \paragraph{ReAct.}

  The ReAct adaptation combines reasoning and tool use
  over the authorized current bus state~\citep{yao2023react}.
  The final holding plan is submitted through the common
  execution interface.
  Tool responses inform the plan rather than directly
  executing physical interventions.
  The call records contain one recorded model stage per
  scheduled decision.

  \paragraph{Recent-history.}

  Recent-history receives the current observation and
  180\,s of recent observations, actions, and execution
  feedback from its own trajectory.
  External development examples, procedural cards, and
  long-term runtime memory are excluded.
  It does not update procedural memory or manage its
  lifecycle.
  The comparison evaluates the broader MEMBus memory
  configuration beyond access to recent operational context.

  \paragraph{Self-Refine.}

  Self-Refine performs one draft--feedback--refinement
  cycle per decision~\citep{madaan2023selfrefine}.
  The model drafts a holding plan, critiques that draft,
  and generates a refined plan.
  Only the final plan is submitted for execution.
  The intermediate feedback is model-generated critique
  rather than newly observed traffic feedback.
  All three stages enter the call and token records.

  \paragraph{Reflexion.}

  The decision-level Reflexion adaptation maintains a
  bounded buffer of verbal reflections on available
  execution feedback~\citep{shinn2023reflexion}.
  These reflections inform later decisions without
  restarting the simulation.
  Action generation and reflection are recorded as
  separate model stages.
  The retained records contain 40 action and 39 reflection
  calls per short-window case, and 96 action and 95
  reflection calls per long-window case.

  \paragraph{Holding activity on the 10-route fixed-fleet subset.}
  Evaluated with \texttt{gpt-6-luna} across the same 10 routes $\times$ 5 seeds and decision ticks (1.5\,h: 40 ticks of 15\,s; 8\,h: 96 ticks of 300\,s), all accepted runs of the five LLM baselines and MEMBus completed with zero failed calls (gateway retry details in Appendix~\ref{app:compute}). On this fixed-fleet subset, standard per-bus LLM baselines issue few holds (8\,h positive holding submissions per case: DiLu 8.2, ReAct 10.3, Reflexion 0.5) and perform similarly to zero additional holding, whereas controllers that hold frequently without calibrated evidence degrade waiting time (Recent-history: 70 positive holds per case).

  \paragraph{ReAct with diagnostic tools.}
  Because the default per-bus ReAct prompt constrained output to direct JSON plans without invoking tools, we evaluated two tool-enabled ReAct variants on the 10-route subset. First, on a 1.5\,h subset (routes 152, 96, 17, M425, \textit{Zhuan}183 $\times$ 3 seeds), enabling tools that merely re-format prompt fields led to 1 tool call across 888 decisions ($-1.2$\,s $[-3.4,+0.9]$ vs.\ ReAct). Second, providing three informative current-state tools (headway diagnostics, a deterministic hold-effect calculator, and recent gap trends, with no simulator rollout, future information, or cross-episode memory) raised tool usage to 20\% of decisions at 1.5\,h (4.6 vs.\ 1.7 holds per case) without reducing waiting ($+2.2$\,s $[-0.3,+4.6]$ vs.\ ReAct, $+3.6$\,s $[+1.4,+5.6]$ vs.\ zero holding, and $+18.5$\,s $[+16.4,+20.6]$ vs.\ MEMBus). Over the full 8\,h horizon across all 10 routes $\times$ 5 seeds (4,800 decisions, 0 failed decisions, 465M tokens; tools invoked in 36\% of decisions, 1,879 calls), tool-augmented ReAct reduced headway CV relative to standard ReAct ($-10.9\%$ \{$-18.9,-3.8$\}) while leaving waiting unchanged ($+14.5$\,s, $+0.6\%$ \{$-19.8,+48.8$\}, W/T/L 3/0/7, served $-0.04$\,pp, n.s.). MEMBus outperformed tool-augmented ReAct on all 10 routes: waiting $-131.7$\,s ($-5.1\%$ \{$-195.0,-78.4$\}, 10/0/0), CV $-39.8\%$ \{$-60.6,-12.8$\}, and served $+0.60$\,pp \{$+0.41,+0.81$\} (10/0/0), with hourly waiting differences of $-18.3$ (n.s.), $-48.2$, $-75.2$, $-95.6$, $-110.6$, $-120.6$, $-126.5$, and $-131.7$\,s at $h=1,\dots,8$ (significant from $h=2$). Current-state diagnostic tools improve headway regularity for per-bus ReAct but do not substitute for operational memory.

  \paragraph{Per-bus agents augmented with MEMBus cards.}
  Appending MEMBus's operating memory (identical 7-program menu, route cards, and constrained recommendation rule) directly to the per-bus ReAct and DiLu prompts without providing the deterministic program executor yields the following paired contrasts at 1.5\,h (10 routes $\times$ 5 seeds; seed-within-route $[~]$ and hierarchical $\{~\}$ CIs; W/T/L for the first listed arm, tie $|\Delta|<1\%$):
  \begin{center}\small
  \resizebox{\linewidth}{!}{%
  \begin{tabular}{@{}lcccc@{}}
  \toprule
  Contrast & $\Delta W$ (s) & $\Delta C$ & $\Delta S$ (pp) & W/T/L\\
  \midrule
  ReAct+cards $-$ MEMBus & $+5.7$ [$+2.9,+9.1$] \{$-0.7,+14.4$\} & $+0.054$ \{$-0.009,+0.138$\} & $-0.83$ & 0/8/2\\
  DiLu+cards $-$ MEMBus & $+4.8$ [$+2.6,+7.5$] \{$-1.8,+13.8$\} & $+0.040$ \{$0.000,+0.093$\} & $-0.68$ & 1/7/2\\
  ReAct+cards $-$ ReAct & $-3.0$ [$-6.1,+0.6$] \{$-11.8,+3.0$\} & $-0.049$ \{$-0.107,-0.004$\} & $+0.81$ & 3/7/0\\
  DiLu+cards $-$ DiLu & $-3.7$ [$-5.9,-0.7$] \{$-10.1,+0.5$\} & $-0.065$ \{$-0.137,-0.009$\} & $+0.85$ & 2/8/0\\
  \bottomrule
  \end{tabular}}
  \end{center}
  With cards appended, per-bus agents hold roughly 16 times per 1.5\,h case (versus 1--2 without) at 1.8--3.7$\times$ MEMBus's token usage, leaving residual gaps on routes whose recommended program actively balances headways (152, \textit{Zhuan}183). At 8\,h (10 routes $\times$ 5 seeds; one ReAct+cards run on route 16 failed validation twice and is counted as a failure), ReAct+cards minus MEMBus is $+80.8$\,s [$+63.0,+98.5$] \{$+33.0,+138.5$\} (W/T/L 0/1/9, $\Delta C=-0.012$ n.s., $\Delta S=-0.44$\,pp) and DiLu+cards minus MEMBus is $+42.5$\,s [$+36.3,+48.7$] \{$+23.9,+61.9$\} ($\Delta C=+0.135$), while cards improve ReAct by $-35$\,s \{$-148,+76$\} and DiLu by $-69$\,s \{$-139,-8$\} relative to their unaugmented per-bus baselines.

  \subsection{Rule-Based Baselines}

  The following rules are implemented headway-control
  references informed by the bus-control literature
  (\citealp{daganzo2009bunching}; \citealp{daganzo2011cooperation}; \citealp{xuan2011dynamic}; \citealp{bartholdi2012self}).
  Their formulas and adapter settings define the evaluated
  configurations.

  \paragraph{TwoWay.}

  Adapted to station holding from cooperative two-way headway feedback~\citep{daganzo2011cooperation,xuan2011dynamic}, TwoWay combines forward and backward headway information:
  \[
  h_b^{\mathrm{raw}}
  =
  \max\!\left(
  0,\,
  0.6(H-f_b)-0.4\max(0,H-b_b)
  \right),\qquad h_b=\min(h_b^{\mathrm{raw}},\bar h_{b,t}).
  \]
  The baseline adapter bounds the duration by the legal
  holding limit.
  Unavailable forward-headway information produces KEEP.

  \paragraph{ForwardHeadway.}

  The forward-headway rule computes
  \[
  h_b^{\mathrm{raw}}
  =
  \max\!\left(0,\,0.5(H-f_b)\right),
  \qquad
  h_b=\min(h_b^{\mathrm{raw}},u_{b,t}).
  \]
  A zero-duration command clears pending holding duration.

  \paragraph{BackwardHeadway.}

  The implementation recorded as
  \texttt{BackwardHeadway45} computes
  \[
  h_b
  =
  \min\!\left(
  \max(0,0.15b_b),\,
  \max(0,\min(45,u_{b,t}))
  \right).
  \]
  It uses the headway field supplied by its adapter and
  may command multiple eligible vehicles.

  \paragraph{Headway feedback and tuned rule/MPC baselines.}

  The fixed slack-offset Daganzo reference applies
  \[
  h_b
  =
  \operatorname{clip}
  \left(
  15-0.6(f_b-H),\,
  0,\,
  \max(0,\min(90,u_{b,t}))
  \right),
  \]
  which is a parameterized forward-headway holding rule~\citep{daganzo2009bunching}. On the timetable-grounded benchmark (Table~\ref{tab:baseline-comparison} and Table~\ref{tab:s11-main}), \textit{Daganzo rule (tuned)} selects the per-route development-best forward gain and slack offset, \textit{Headway feedback (tuned)} selects the per-route development-best symmetric two-sided balance rule $h_b=\min(\mathrm{cap},\bar h_{b,t},\max(0,\mathrm{gain}\cdot(b_b-f_b)))$, \textit{Per-route tuned rule (18 programs)} optimizes per route over the 18-program stationary forward/backward/two-way gain--cap grid (24 programs when combined with MEMBus's 6 balance programs), and \textit{MPC (per route)} solves a receding-horizon quadratic headway-variance and queue-waiting minimization over predicted stop arrivals at each decision tick.

  \paragraph{Zero additional holding.}

  This baseline clears pending control-hold duration where
  permitted while preserving passenger-service dwell.
  It does not reverse dwell already served.
  Submitting zero is distinct from KEEP, which preserves
  the pending command.

  \paragraph{NoControl.}

  NoControl provides a separate reference without control
  interventions.
  Normal vehicle movement and passenger-service dwell
  remain active.
  Unlike Zero additional holding, it does not issue
  commands to clear pending holding durations.
  On the 10-route 8\,h protocol, the three zero-intervention references yield slightly different mean waits due to pending-command initialization: NoControl (2582\,s), the KEEP program (2547\,s; KEEP at every tick), and Zero additional holding (2540\,s; explicit zero hold at each review). In the regime-change and running-clock experiments on the timetable-grounded benchmark, no control denotes the KEEP program.

  \subsection{Learning-Based Controllers}

  All learned policies are frozen during evaluation.
  Their native action spaces may differ even though they
  use the common physical execution interface.
  The descriptions below summarize the algorithmic basis
  of each method and the available details of its
  bus-control adaptation.
  Training mechanisms are distinguished from online
  control: policy parameters are not updated during
  the reported evaluation episodes.

  \paragraph{DQN.}

  DQN selects holding durations using a frozen action-value
  network~\citep{mnih2015human}.
  Its action set is
  $\{0,5,10,15,20,25,30\}$\,s.
  Inputs contain normalized headway deviations, three
  zero-filled coordinates, vehicle load fraction,
  sine and cosine clock features, and a constant.
  The highest-value action is selected.
  The evaluated adapter returns KEEP if that action exceeds
  the legal holding limit.

  \paragraph{DDPG and SAC.}

  DDPG uses a deterministic actor for continuous
  control~\citep{lillicrap2016continuous}, whereas SAC
  uses entropy-regularized actor--critic
  learning~\citep{haarnoja2018soft}.
  The evaluated implementations are trained on a fixed
  development transition dataset and transferred to
  the evaluation cases.
  Their actors remain frozen, and proposals are processed
  by their holding adapters.

  \paragraph{TD3.}

  TD3 extends deterministic actor--critic learning with
  twin critics, delayed policy updates, and target-policy
  smoothing~\citep{fujimoto2018addressing}.
  Using the smaller target value from the two critics
  reduces overestimation during training.
  The trained actor provides continuous action proposals
  during evaluation.

  \paragraph{MADDPG.}

  MADDPG uses centralized critics during multi-agent
  training~\citep{lowe2017multiagent}.
  The evaluated actors generate holding proposals from
  their assigned observation inputs.
  Policy parameters remain fixed during evaluation.

  \paragraph{CAAC.}

  CAAC addresses credit assignment for asynchronous
  bus-control events~\citep{wang2021asynchronous}.
  Its training framework accounts for intervening actions
  by other vehicles when attributing outcomes to a decision.
  The frozen policy is evaluated through the decision
  schedule and execution interface of each protocol.

  \paragraph{MPPO.}

  The transferred MPPO implementation maps normalized
  forward and backward headways and a scaled target-stop
  queue through a frozen actor to a holding duration
  bounded by the legal action interface.
  It serves as a transferred policy-based bus-control
  reference.

  \paragraph{RLHold.}

  RLHold uses a frozen matrix to score the discrete
  actions $\{0,7.5,15,22.5,30\}$\,s.
  Its features comprise normalized forward and backward
  headway deviations, their product, and a constant.
  The selected action is processed by the holding adapter.
  Only the largest positive holding candidate is retained
  at an execution opportunity, with vehicle identity
  resolving ties.

  \paragraph{AWAC.}

  AWAC combines off-policy value learning with
  advantage-weighted policy regression
  \citep{nair2021awac}.
  Actions in the training data receive larger policy-update
  weights when their estimated advantages are higher.
  Although the original method supports subsequent online
  fine-tuning, the evaluated policy remains frozen during
  the bus-control episodes.

  \paragraph{CRR.}

  Critic Regularized Regression uses critic estimates
  to filter or weight actions in the training data during
  policy regression~\citep{wang2020crr}.
  This favors recorded actions with higher estimated
  value while anchoring policy learning to available
  data.
  The resulting policy is used without online parameter
  updates during evaluation.

  \paragraph{BCQ.}

  Batch-Constrained Q-learning restricts action selection
  to candidates supported by the training data
  \citep{fujimoto2019offpolicy}.
  In its continuous-action formulation, a learned
  generative model proposes candidate actions, a bounded
  perturbation model adjusts them, and the critic scores
  the resulting candidates.
  This reduces reliance on value estimates for actions
  far outside the data distribution.

  \paragraph{BEAR.}

  BEAR reduces bootstrapping error by constraining the
  learned policy toward the support of the behavior data
  \citep{kumar2019stabilizing}.
  Its training objective uses a maximum mean discrepancy
  constraint between policy actions and actions sampled
  from a learned behavior model.
  The constraint limits unsupported action choices
  during policy optimization.

  \paragraph{CQL.}

  Conservative Q-learning augments value learning with
  a regularizer that discourages overestimated values
  for actions insufficiently supported by the training
  data~\citep{kumar2020conservative}.
  It provides an offline-learning reference based on
  conservative value estimation.
  The learned controller remains frozen during evaluation.

  \paragraph{IQL.}

  Implicit Q-learning estimates state values through
  expectile regression and extracts a policy using
  advantage-weighted behavioral cloning
  \citep{kostrikov2022implicit}.
  Its value-learning procedure avoids explicitly
  maximizing over out-of-distribution actions.
  The extracted policy supplies actions without further
  training during the evaluation episodes.

  \paragraph{TD3+BC.}

  TD3+BC adds a behavior-cloning term to the TD3 actor
  objective~\citep{fujimoto2021minimalist}.
  The resulting update balances estimated action value
  with agreement with actions in the training dataset.
  This provides a simple offline-learning reference
  based on a regularized deterministic actor.

\subsection{Extended Controller Comparisons}
  \label{app:extended-controller-comparisons}

  Table~\ref{tab:extended-controller-results} reports extended
  rule-based and learning-based controller comparisons alongside
  an uncontrolled reference on the auxiliary ten-route fixed-fleet subset
  (ten routes $\times$ five seeds per route under both the 1.5\,h and 8\,h protocols).
  Outcomes are averaged across seeds within each route
  and then equally across routes.
  The two protocols retain their respective control and
  scoring windows.

  NoControl applies no control intervention, whereas
  Zero additional holding clears pending control holds
  while preserving passenger-service dwell.
  The table reports descriptive mean outcomes.

  \begin{table}[t]
    \centering
    \small
    \setlength{\tabcolsep}{3pt}
    \caption{Extended rule-based and learning-based controller comparisons on the auxiliary 10-route fixed-fleet subset (50 cases per method and protocol).}
    \label{tab:extended-controller-results}
    \begin{tabular}{@{}lrrr@{}}
      \toprule
      Controller
        & Wait (s)$\downarrow$
        & CV$\downarrow$
        & Served (\%)$\uparrow$ \\
      \midrule
      MEMBus
        & 591.4 / 2438.5
        & 0.627 / 0.547
        & 65.57 / 85.71 \\
      \midrule
      \multicolumn{4}{l}{\textit{Rule-based controllers}} \\
      TwoWay
        & 632.21 / 2818.84
        & 0.53792 / 0.44621
        & 62.363 / 83.276 \\
      ForwardHeadway
        & 653.90 / 3080.71
        & 0.54696 / 0.41490
        & 61.334 / 81.534 \\
      BackwardHeadway
        & 632.82 / 3037.90
        & 0.60322 / 0.72058
        & 62.279 / 81.810 \\
      Headway feedback
        & 674.68 / 3345.36
        & 0.48846 / 0.37687
        & 60.961 / 79.553 \\
      Zero additional holding
        & 599.20 / 2539.86
        & 0.72903 / 1.00151
        & 64.023 / 85.274 \\
      \midrule
      \multicolumn{4}{l}{\textit{Learning-based controllers}} \\
      MPPO
        & 632.77 / 2745.50
        & 0.46510 / 0.27897
        & 62.905 / 83.647 \\
      DQN
        & 646.73 / 2938.67
        & 0.65576 / 0.63762
        & 61.183 / 82.858 \\
      RLHold
        & 617.39 / 2607.22
        & 0.64040 / 0.75233
        & 62.936 / 84.726 \\
      CAAC
        & 621.18 / 2802.01
        & 0.63524 / 0.72797
        & 62.561 / 83.477 \\
      MADDPG
        & 679.74 / 3484.13
        & 0.69462 / 0.71238
        & 59.377 / 78.870 \\
      DDPG
        & 718.39 / 3918.69
        & 0.71730 / 0.75632
        & 57.508 / 75.936 \\
      TD3
        & 717.12 / 3885.71
        & 0.70418 / 0.72088
        & 57.585 / 76.177 \\
      SAC
        & 718.13 / 3950.08
        & 0.72567 / 0.81013
        & 57.328 / 75.753 \\
      AWAC
        & 638.47 / 3097.53
        & 0.76223 / 1.23509
        & 61.975 / 81.649 \\
      CRR
        & 709.41 / 3833.33
        & 0.74415 / 0.84872
        & 57.799 / 76.603 \\
      BCQ
        & 602.23 / 2666.22
        & 0.73760 / 1.06339
        & 64.146 / 84.289 \\
      BEAR
        & 646.01 / 3132.95
        & 0.72207 / 0.86743
        & 62.763 / 80.933 \\
      CQL
        & 600.32 / 2499.71
        & 0.58321 / 0.52438
        & 65.401 / 85.346 \\
      IQL
        & 640.62 / 2979.62
        & 0.65705 / 0.94440
        & 62.947 / 82.143 \\
      TD3+BC
        & 634.08 / 2928.76
        & 0.77364 / 1.26676
        & 61.922 / 82.794 \\
      \midrule
      \multicolumn{4}{l}{\textit{Uncontrolled reference}} \\
      NoControl
        & 616.56 / 2582.24
        & 1.14582 / 1.23594
        & 61.933 / 85.125 \\
      \bottomrule
    \end{tabular}
  \par\smallskip\noindent\parbox{\linewidth}{\footnotesize Each cell reports 1.5\,h-window / 8\,h-window means. Served fractions are percentages.}
  \end{table}

  The extended comparison shows that controller trade-offs
  depend strongly on fleet saturation.
  On this fixed-fleet subset, CQL attains a slightly lower headway CV than MEMBus in both windows, whereas MEMBus achieves lower passenger waiting ($-9$\,s at 1.5\,h and $-61$\,s at 8\,h) and a higher served fraction, and improves both waiting and served fraction over NoControl across horizons.

  \subsection{Model Usage and Execution Accounting}

  \paragraph{Model calls and tokens.}

  Model usage includes the stages implemented by each
  controller, including evidence selection, action generation,
  refinement, and reflection where applicable.
  Logical model calls are distinguished from physical
  request attempts and traffic decision opportunities.
  Reusing an existing plan does not invoke the model.
  Rule-based and learning-based controllers do not invoke
  an LLM during evaluation.

  \paragraph{Commands and physical execution.}

  Command submissions are distinguished from KEEP choices
  and subsequent physical execution.
  KEEP preserves a pending command, whereas SET and CANCEL
  replace or clear its duration.
  An accepted command may be replaced before consumption.
  Execution statistics cover the controller's intervention
  window, excluding warmup and the common short-window
  numerical continuation.

  \paragraph{Multi-metric composite rank definition.}
  To summarize multi-metric trade-offs among $M$ LLM controllers across waiting ($W$), headway CV ($C$), served fraction ($S$), token usage ($T$), and bunching ($B$), let $r_{m,h,k}$ denote controller $m$'s rank on metric $k$ in window $h$ (ascending for $W,C,T,B$; descending for $S$; average ranks for ties). Under a weight vector $\mathbf w=(w_W,w_C,w_S,w_T,w_B)$ on the unit simplex, the within-window score is $R_{m,h}(\mathbf w)=\sum_k w_k r_{m,h,k}$. To evaluate ranking stability across metric trade-offs without fixing a single weight vector, Section~\ref{sec:baseline-comparison} and Appendix~\ref{app:xmodel-fig} evaluate $R_{m,8\mathrm{h}}(\mathbf w)$ across all 286 points of the 0.1-step five-metric simplex.

  \paragraph{Result identity and reuse.}

  Each comparison retains its route--seed identities,
  controller configuration, endpoint version, and
  measurement definitions.
  Service outcomes and execution statistics use their
  respective case sets and accounting units.
  An endpoint appearing in multiple comparisons denotes
  the same underlying run.
  
 \section{Metrics and Statistical Analysis}
  \label{app:comparison-design}

  \paragraph{Statistical inference.}
  All contrasts are paired by route and seed. On the primary timetable-grounded benchmark (Section~\ref{sec:protocol}; Appendix~\ref{app:s11}), the primary hypothesis test is the exact two-sided route-level sign-flip test over route means, Holm-adjusted within each pre-specified hypothesis family, with hierarchical bootstrap confidence intervals reported as a secondary summary (Appendix~\ref{app:s11-stats}). Across our evaluations, we distinguish seed-within-route bootstrap intervals $[~]$ (20,000 resamples of seeds within each fixed route) from hierarchical bootstrap intervals $\{~\}$ (which resample routes and then seeds within each route), reported alongside per-route win/tie/loss (W/T/L) counts.

  \paragraph{Scoring cohort.}

  For scoring interval $[t_0,t_1)$, the passenger cohort $\mathcal Q$ contains
  passengers waiting immediately before events at $t_0$
  and passengers arriving during the interval.
  Passenger mass can be fractional.
  FIFO records track arrival times and boarding events.
  When a record is only partly boarded, boarded and unserved
  portions are accounted for separately.

  Let $a_p$ denote arrival time, $b_p$ boarding time,
  and $m_p$ the mass of a passenger record or portion $p\in\mathcal Q$.
  Set $b_p=\infty$ for mass remaining unserved.
  For a nonempty cohort $\mathcal Q$, restricted mean waiting is
  \begin{equation}
  W=
  \frac{
  \sum_{p\in\mathcal Q}
  m_p\left[\min(b_p,t_1)-\max(a_p,t_0)\right]
  }{
  \sum_{p\in\mathcal Q}m_p
  }.
  \label{eq:appendix-restricted-wait}
  \end{equation}
  Waiting before $t_0$ is excluded.
  Unserved mass contributes waiting through $t_1$ and
  remains in the denominator.
  The endpoint measures queue waiting rather than
  total journey time.

  \paragraph{Served fraction and endpoint queue.}

  Served fraction uses the same passenger cohort $\mathcal Q$:
  \begin{equation}
  S=
  \frac{
  \sum_{p\in\mathcal Q}
  m_p\,\mathbf 1[b_p<t_1]
  }{
  \sum_{p\in\mathcal Q}m_p
  }.
  \label{eq:appendix-served-fraction}
  \end{equation}
  Endpoint queue is the remaining waiting mass at the
  scoring boundary, before arrivals at $t_1$ enter
  the next interval.
  Tables express served fraction as a percentage and
  between-controller differences in percentage points.

  \paragraph{Headway CV.}

  For each stop, consecutive arrivals within the scoring
  window define arrival intervals.
  Both arrivals defining an interval must lie within
  $[t_0,t_1)$.
  The intervals from all stops are pooled as
  $g_1,\ldots,g_n$.
  We compute
  \[
  \bar g=\frac{1}{n}\sum_{j=1}^{n}g_j,
  \qquad
  s_g=
  \sqrt{
  \frac{1}{n-1}\sum_{j=1}^{n}(g_j-\bar g)^2
  },
  \qquad
  C=\frac{s_g}{\bar g}.
  \]
  The metric requires at least two intervals and a positive
  mean.
  It is not the average of separately computed stop-level
  CVs.
  The arrival intervals used for evaluation are distinct
  from the spatial headways supplied to controllers.

  \paragraph{Bunching events.}

  A bunching event is counted when the interval between
  two consecutive arrivals at the same stop is strictly
  less than 25\% of the target headway.
  Both arrivals must lie within the scoring window:
  1,800\,s for the 1.5\,h-window protocol and 28,800\,s
  for the 8\,h-window protocol.
  Using the pooled within-stop arrival intervals
  $g_1,\ldots,g_n$, the case-level count is
  \[
  B=\sum_{j=1}^{n}\mathbf{1}[g_j<0.25H],
  \]
  where $H$ is the route's target headway.

  Counts are computed from stop-arrival records rather
  than periodic state scans.
  Each qualifying interval contributes one event.
  Repeated bunching by the same vehicle pair at different
  stops is counted separately, rather than merged into
  one continuous episode.
  Tables report mean case-level counts, averaged within
  routes and then equally across routes.

  \paragraph{Aggregation and pairing.}

  For route $r\in\{1,\dots,R\}$ with target seed set $\mathcal S_r$ ($R=8$ or $17$ on the primary timetable-grounded benchmark, and $R=10$ on the auxiliary fixed-fleet subset, with $|\mathcal S_r|=5$ unless otherwise noted), the
  controller mean is
  \[
  \bar Y_r=
  \frac{1}{|\mathcal S_r|}
  \sum_{s\in\mathcal S_r}Y_{r,s}.
  \]
  The overall mean gives equal weight to the $R$ routes:
  \[
  \bar Y=\frac{1}{R}\sum_{r=1}^{R}\bar Y_r.
  \]
  When $|\mathcal S_r|$ is equal across routes, this coincides with
  the unweighted mean over case-level endpoints.
  Waiting and served fraction are first computed within
  each case; passengers from different cases are not pooled
  into a single aggregate ratio.

  Pairwise contrasts use matched route--seed cases
  within the same protocol.
  For baseline $k$, the case-level difference is
  \[
  d_{r,s}^{(k)}
  =
  Y_{r,s}^{\mathrm{MEMBus}}
  -
  Y_{r,s}^{(k)}.
  \]
  These differences are averaged within routes and then
  equally across routes.
  Negative waiting, CV, and bunching differences favor
  MEMBus; positive served-fraction differences favor MEMBus.
  Reported contrasts use unrounded endpoints.

  Comparisons based on a subset identify the included
  cases and use the same subset for both controllers.
  Full-suite sample sizes are never assigned to
  statistics computed from fewer cases.

  \paragraph{Usage and execution statistics.}

  Service endpoints, model usage, and command execution
  have distinct accounting units.
  Logical calls, physical request attempts, per-vehicle
  action choices, submitted commands, and consumed commands
  are distinguished.
  A command submission does not by itself establish
  additional physical dwell.
  Token usage and execution accounting are described in
  Appendices~\ref{app:supporting-evaluations} and~\ref{app:compute}.

  \paragraph{Uncertainty intervals.}

  Seed-within-route intervals $[~]$ use 20,000 resamples of seeds within each fixed route; hierarchical intervals $\{~\}$ additionally resample routes. Summary tables compute bootstrap resamples using independent Monte Carlo streams. Confidence intervals are unadjusted for multiplicity. Throughout this paper, ``pre-specified'' denotes analysis plans and hypothesis families fixed prior to executing the corresponding target-seed runs. Seed-within-route intervals describe seed variation conditional on the evaluated routes, whereas route-generalization claims rely on route-level sign-flip tests and hierarchical intervals.

  \paragraph{Comparisons across protocols.}

  On the auxiliary ten-route fixed-fleet subset, the 1.5\,h and 8\,h protocols share the same ten routes and five seeds per route, and service outcomes are computed separately under their respective control schedules and scoring windows.
  Because the two protocols differ in control duration, decision
  spacing, and numerical continuation, bunching counts cover scoring windows of different
  lengths and are compared between controllers within each
  protocol rather than across protocols. Multi-metric weighting sensitivity on the primary benchmark is evaluated across a 286-point simplex grid in Appendix~\ref{app:xmodel-fig}.

  \paragraph{Failures and dependent observations.}

  Recorded query failures, invalid action responses,
  and KEEP fallbacks remain part of completed evaluated
  trajectories, except on the auxiliary 8\,h subset where runs of memory-augmented baselines that failed twice were excluded pairwise (4 RAG runs and 2 ExpeL runs; Appendix~\ref{app:compute}).
  Query failures and action fallbacks are counted
  separately because a failed intermediate query need
  not cause an action fallback.

  Decision counts within a run are not independent
  traffic replications.
  Reusing endpoints across tables, replaying recorded
  responses, or repeating a route--seed case does not
  increase the independent sample count.
  Incomplete cases and missing execution records are
  identified rather than assigned zero outcomes.

  \section{Compute, Software, and Replay}
  \label{app:compute}

  \paragraph{Model usage.}

  The 1.5\,h short-window protocol provides 40 decision opportunities
  per run, and the 8\,h protocol provides 96.
  In the primary program-selection interface, MEMBus issues one call per review (32 reviews per 8\,h episode), whereas the auxiliary two-stage per-bus variant executes evidence selection followed by action generation.
  Other LLM controllers follow their configured retrieval,
  reasoning, refinement, or reflection procedures.

  In the auxiliary short-window per-bus protocol, the current-observation control retains
  the evidence-selection stage with an empty catalog, whereas
  the 8\,h implementation skips this empty query.
  Decision opportunities therefore do not imply identical
  call counts across all configurations.

  Usage accounting distinguishes logical model calls,
  physical request attempts, successful responses, and
  responses used for control.
  Input, cached-input, and output tokens, reference-token
  estimates, retries, and wall-clock latency are recorded
  separately where available.
  Missing usage records are not assigned zero.
  Rule-based and learning-based controllers make no LLM
  calls during evaluation.

  \paragraph{Failure accounting and gateway retries.}

  In their accepted runs on the primary timetable-grounded benchmark (Table~\ref{tab:baseline-comparison}), MEMBus and the LLM baselines completed with zero failed calls. On the auxiliary 10-route 8\,h subset, one run of ReAct+cards failed validation twice and is counted as a failure, while 4 RAG runs (routes \textit{Zhuan}183, 96) and 2 ExpeL runs (routes 1, 96) failed after one re-run and are excluded pairwise. Transient HTTP 503 and HTTP 520 gateway errors encountered on early auxiliary 8\,h batches (routes 17, 152, 96, 135, and 167) were resolved by enabling transport-level gateway retries and re-running the affected cases; paired checks across 15 cases per method confirmed negligible differences between attempts (8\,h wait differences: ReAct $+7.2$\,s [$-13.9,+27.9$], DiLu $-5.3$\,s [$-31.5,+19.8$], two-stage MEMBus $+11.4$\,s [$-5.9,+30.4$], current-observation LLM $+24.3$\,s [$-8.7,+56.0$]). In the auxiliary 50-case short-window two-stage comparison, both MEMBus and the current-observation control recorded 98 action fallbacks across 2,000 decisions (351 stage-level transport failures out of 8,000 stage calls).

  Query-stage failures and action fallbacks are separate
  events.
  A failed evidence query need not prevent action generation,
  whereas an action failure invokes the specified KEEP
  fallback.
  Fallback decisions remain in the trajectories used
  to compute service outcomes.
  Per-vehicle fallback KEEP choices are distinguished
  from failed fleet-level decisions.

  \paragraph{Software and replay.}

  Reproduction requires the simulator and asset versions,
  controller configuration, initial memory, random seeds,
  and metric definitions associated with each result.
  Stored requests, responses, actions, and execution
  records support inspection and replay.

  Exact closed-loop trajectory reproduction uses the archived request--response logs alongside live API replication checks (Appendix~\ref{app:s11-sameif}).

  Archived responses support deterministic replay of the recorded
  controller decisions.
  Response reuse requires the request, prompt, and schema
  to match the saved originals alongside simulation-state and
  execution checks.
  Replay reproduces recorded online evaluations deterministically.

  In synchronous protocols, the simulation waits for model inference before advancing
  simulated time, so wall-clock latency is recorded as computational
  cost rather than added to simulated passenger waiting; running-clock evaluations are reported in Appendices~\ref{app:async} and~\ref{app:async-data}.

 \section{\texorpdfstring{Timetable-Grounded Benchmark: Construction and Extended Results}{Timetable-Grounded Benchmark: Construction and Extended Results}}
  \label{app:s11}

  This appendix details the route construction, full baseline tables, statistical testing protocol, same-interface comparisons, and 45-route supply release for the primary timetable-grounded benchmark of Section~\ref{sec:protocol}.

  \subsection{\texorpdfstring{Construction}{Construction}}
  \label{app:s11-construction}

  Each route is modeled as a closed one-direction service loop sized from its AM-peak timetable headway (Table~\ref{tab:s11-routes}).

  \begin{table}[!htbp]
    \centering
    \caption{The eight main timetable-grounded routes.}
    \label{tab:s11-routes}
    \small
    \begin{tabular}{@{}llrrr@{}}
      \toprule
      Route & City & Stops & Buses & $H_{\mathrm{am}}$ (s) \\
      \midrule
      598 & Beijing & 29 & 4 & 655 \\
      546 & Beijing & 30 & 6 & 450 \\
      484 & Beijing & 21 & 3 & 800 \\
      SH50 & Shanghai & 26 & 6 & 514 \\
      CD306 & Chengdu & 31 & 4 & 800 \\
      CD50 & Chengdu & 40 & 11 & 360 \\
      CD99 & Chengdu & 21 & 3 & 800 \\
      K12 & Chengdu & 27 & 9 & 343 \\
      \bottomrule
    \end{tabular}
  \par\smallskip\noindent\parbox{\linewidth}{\footnotesize Buses $=$ round(simulated KEEP loop time of the one-direction ring / timetable AM-peak headway $H_{\mathrm{am}}$); demand $\lambda=12.25/H_{\mathrm{am}}$ per stop per second.}
  \end{table}

  Route specifications are reconstructed from the full route catalog. Candidates 839, 913 and 973 were excluded by the pre-specified screening rule (served 0.87--0.94 with KEEP queue growth of 39--91\% from $h=4$ to $h=8$; 913 and 973 also require more than 7200\,s of warm-up). The 17 confirmation routes are BJ2, BJ337, BJ342, BJ386, BJ402, BJ425, BJ55, BJ556, BJ58, BJ597, BJ663, BJ671, BJ820, BJ9, BJ997, CD56 and CD61 (16--43 stops, 2--11 buses, development-seed KEEP served 91--98\%); of 34 initial candidates, 2 (BJ300W, CDK2) were excluded due to imputed AM-peak departures, 2 (CD47, CD4) lay outside the candidate pool, and 13 of the remaining 30 exceeded the queue-growth threshold (Table~\ref{tab:supply-set}). The simulated headway matches the timetable AM-peak headway within 15\% and is 3--51\% shorter than the timetable 10:00--16:00 headway, because the fleet is constant all day. Dwell noise is keyed by the seed during warm-up and by the seed plus $10^9$ from the start of control; development and target seed families are disjoint. Analysis protocols and hypothesis families were specified prior to executing the target-seed runs; protocol notes are summarized in Appendix~\ref{app:s11-disclosures}.

  \subsection{\texorpdfstring{Relative Results on 8 and 17 Routes}{Relative Results on 8 and 17 Routes}}
  \label{app:s11-rel}

  \begin{table}[!htbp]
    \centering
    \caption{MEMBus against every baseline on the timetable-grounded benchmark.}
    \label{tab:s11-main}
    \scriptsize
    \setlength{\tabcolsep}{3pt}
    \resizebox{\linewidth}{!}{%
    \begin{tabular}{@{}lrcrr@{\hspace{8pt}}rcrr@{}}
      \toprule
      & \multicolumn{4}{c}{8 routes $\times$ 5 seeds} & \multicolumn{4}{c}{17 confirmation routes $\times$ 5 seeds} \\
      \cmidrule(lr){2-5}\cmidrule(l){6-9}
      Baseline & Wait \%, $h=1/2/4/8$ & W/T/L & $p$ & CV \% & Wait \%, $h=1/2/4/8$ & W/T/L & $p$ & CV \% \\
      \midrule
      \multicolumn{9}{@{}l}{\textit{No control}} \\
      Zero additional holding & $-$7.7 / $-$27.3 / $-$47.3 / $-$60.1 & 8/0/0 & 0.039 & $-$73.3 & $-$4.9 / $-$22.4 / $-$44.6 / $-$59.4 & 17/0/0 & \textemdash & $-$71.8 \\
      \multicolumn{9}{@{}l}{\textit{LLM agents (same model)}} \\
      ReAct & $-$7.9 / $-$27.6 / $-$48.2 / $-$60.4 & 8/0/0 & 0.039 & $-$72.8 & $-$5.5 / $-$23.1 / $-$44.8 / $-$59.3 & 17/0/0 & $1.4{\times}10^{-4}$ & $-$71.6 \\
      DiLu & $-$8.0 / $-$27.4 / $-$47.7 / $-$60.2 & 8/0/0 & 0.039 & $-$72.9 & $-$5.4 / $-$22.8 / $-$44.8 / $-$59.4 & 17/0/0 & $1.4{\times}10^{-4}$ & $-$71.7 \\
      \multicolumn{9}{@{}l}{\textit{Four memory-augmented agents (same development evidence) and three reflective or history agents}} \\
      ExpeL & $-$4.3 / $-$15.0 / $-$30.8 / $-$43.0 & 8/0/0 & 0.055 & $-$58.2 & $-$2.8 / $-$11.4 / $-$26.7 / $-$41.4 & 17/0/0 & $1.4{\times}10^{-4}$ & $-$57.3 \\
      RAG over records & $-$5.7 / $-$21.4 / $-$40.4 / $-$52.8 & 8/0/0 & 0.055 & $-$66.7 & $-$3.4 / $-$15.0 / $-$34.0 / $-$50.4 & 17/0/0 & $1.4{\times}10^{-4}$ & $-$64.5 \\
      MemGPT-style & $-$7.9 / $-$27.7 / $-$48.1 / $-$60.2 & 8/0/0 & 0.055 & $-$72.8 & $-$5.5 / $-$23.3 / $-$45.0 / $-$59.5 & 17/0/0 & $1.4{\times}10^{-4}$ & $-$71.7 \\
      Generative Agents & $-$9.5 / $-$22.4 / $-$40.7 / $-$55.3 & 8/0/0 & 0.055 & $-$69.1 & $-$6.8 / $-$19.0 / $-$38.3 / $-$55.4 & 17/0/0 & $1.4{\times}10^{-4}$ & $-$68.2 \\
      Reflexion & $-$7.8 / $-$27.4 / $-$47.5 / $-$59.9 & 8/0/0 & 0.055 & $-$72.9 & $-$5.2 / $-$23.1 / $-$45.1 / $-$59.6 & 17/0/0 & $1.4{\times}10^{-4}$ & $-$72.1 \\
      Self-Refine & $-$7.7 / $-$27.2 / $-$47.6 / $-$60.4 & 8/0/0 & 0.055 & $-$73.1 & $-$5.2 / $-$22.9 / $-$44.9 / $-$59.5 & 17/0/0 & $1.4{\times}10^{-4}$ & $-$71.8 \\
      Recent-history & $-$8.7 / $-$28.1 / $-$47.6 / $-$60.0 & 8/0/0 & 0.055 & $-$72.8 & $-$5.6 / $-$23.4 / $-$45.1 / $-$59.6 & 17/0/0 & $1.4{\times}10^{-4}$ & $-$71.8 \\
      \multicolumn{9}{@{}l}{\textit{Headway rules}} \\
      TwoWay & $-$2.1 / $-$6.3 / $-$13.4 / $-$19.2 & 8/0/0 & 0.008$^{u}$ & $-$25.8 & \textemdash & \textemdash & \textemdash & \textemdash \\
      ForwardHeadway & $-$12.6 / $-$17.7 / $-$17.3 / $-$15.0 & 8/0/0 & 0.008$^{u}$ & $-$7.7 & \textemdash & \textemdash & \textemdash & \textemdash \\
      BackwardHeadway & $-$14.8 / $-$32.2 / $-$51.2 / $-$63.6 & 8/0/0 & 0.008$^{u}$ & $-$75.2 & \textemdash & \textemdash & \textemdash & \textemdash \\
      Per-route tuned rule (18 programs) & $-$5.5 / $-$8.7 / $-$8.9 / $-$5.1 & 6/0/2 & 0.047 & $+$4.8 & $-$4.1 / $-$7.0 / $-$8.8 / $-$8.5 & 16/1/0 & $5{\times}10^{-5}$ & $+$0.5 \\
      \multicolumn{9}{@{}l}{\textit{Model-based and learned controllers ($\dagger$: baseline $-$ MEMBus, relative to MEMBus)}} \\
      MPC (per route)$^\dagger$ & $-$0.4 / $-$1.7 / $-$3.8 / $-$5.6 & 1/1/6 & 0.023$^{u}$ & $-$16.4 & $-$1.2 / $-$2.5 / $-$4.2 / $-$4.8 & 1/2/14 & \textemdash & $-$7.6 \\
      DQN (offline)$^\dagger$ & $-$1.0 / $+$1.3 / $+$5.3 / $+$11.2 & 6/0/2 & \textemdash & $+$33.0 & \textemdash & \textemdash & \textemdash & \textemdash \\
      CQL (offline)$^\dagger$ & $-$1.2 / $+$2.9 / $+$7.4 / $+$8.5 & 8/0/0 & \textemdash & $+$39.7 & \textemdash & \textemdash & \textemdash & \textemdash \\
      \bottomrule
    \end{tabular}}
  \par\smallskip\noindent\parbox{\linewidth}{\footnotesize Cells: MEMBus $-$ baseline, relative difference in cumulative mean wait (\%) over the first $h$ hours, $h=1$ / 2 / 4 / 8 (negative: MEMBus waits less). W/T/L: routes on which MEMBus is better / tied / worse at $h=8$. $p$: exact route-level sign-flip test at $h=8$, Holm-adjusted within each pre-specified family (5 contrasts for the 8 routes, 9 for the 17 routes; $u$: unadjusted; \texttt{gpt-6-luna} DiLu is unstarred in Figure~\ref{fig:xmodel-main}a). CV: relative difference in stop-arrival headway CV at $h=8$. $\dagger$: baseline $-$ MEMBus, relative to MEMBus (negative: baseline waits less); all other rows are MEMBus $-$ baseline, relative to the baseline, and the text compares DQN and CQL with the tuned rule. MEMBus waits 424.9\,s (8 routes) and 512.9\,s (17 routes) at $h=8$ and equals its scripted executor on every route.}
  \end{table}

\begin{figure}[!htbp]
    \centering
    \includegraphics[width=\linewidth,trim=0 188 428 0,clip]{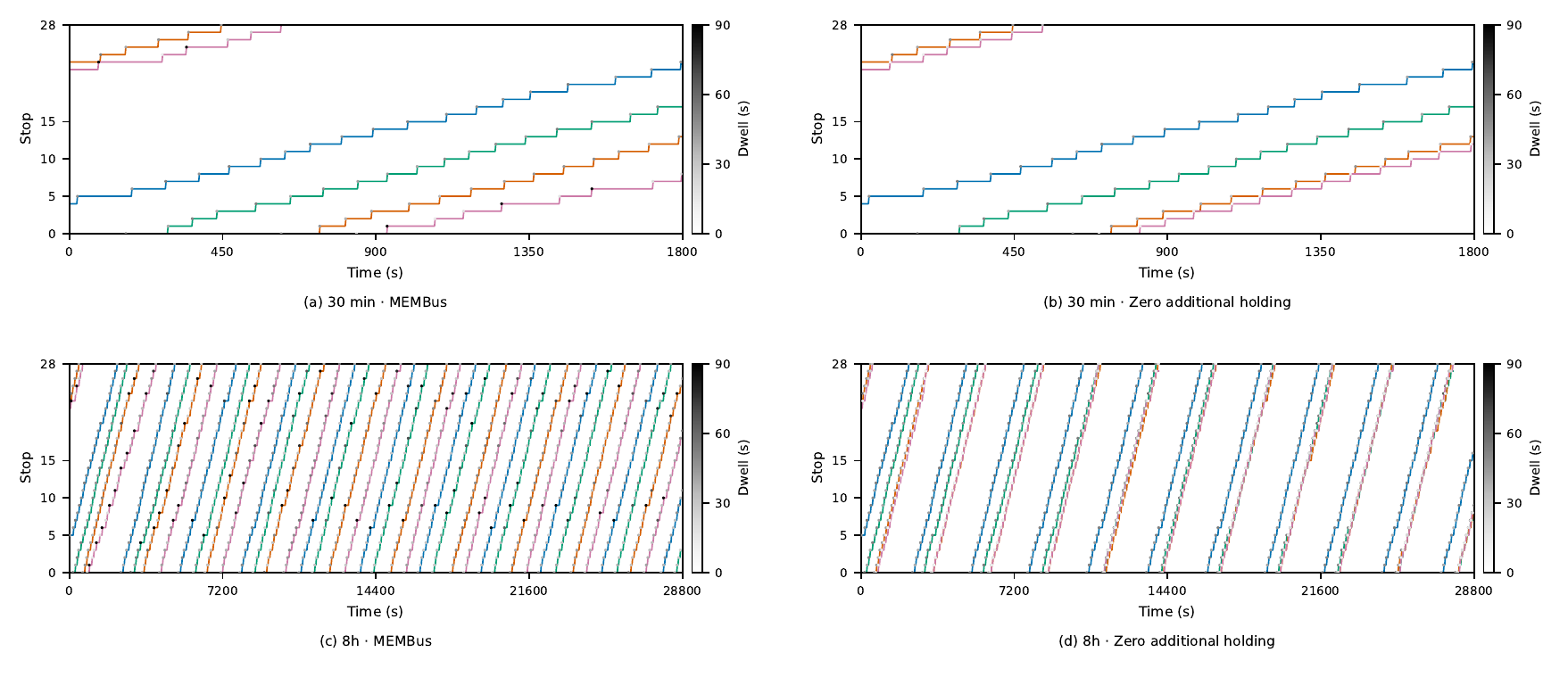}\\[2pt]
    \includegraphics[width=\linewidth,trim=420 188 0 0,clip]{picture/pic/recommended_598_s11_full8h.pdf}
    \caption{Bus trajectories on Beijing Route~598 over 1,800\,s: MEMBus (top) and zero additional holding (bottom).
  Colors identify vehicles; grayscale indicates total dwell.}
    \label{fig:memory-trajectories}
  \end{figure}

\begin{figure}[!htbp]
    \centering
    \includegraphics[width=\linewidth,trim=0 0 428 182,clip]{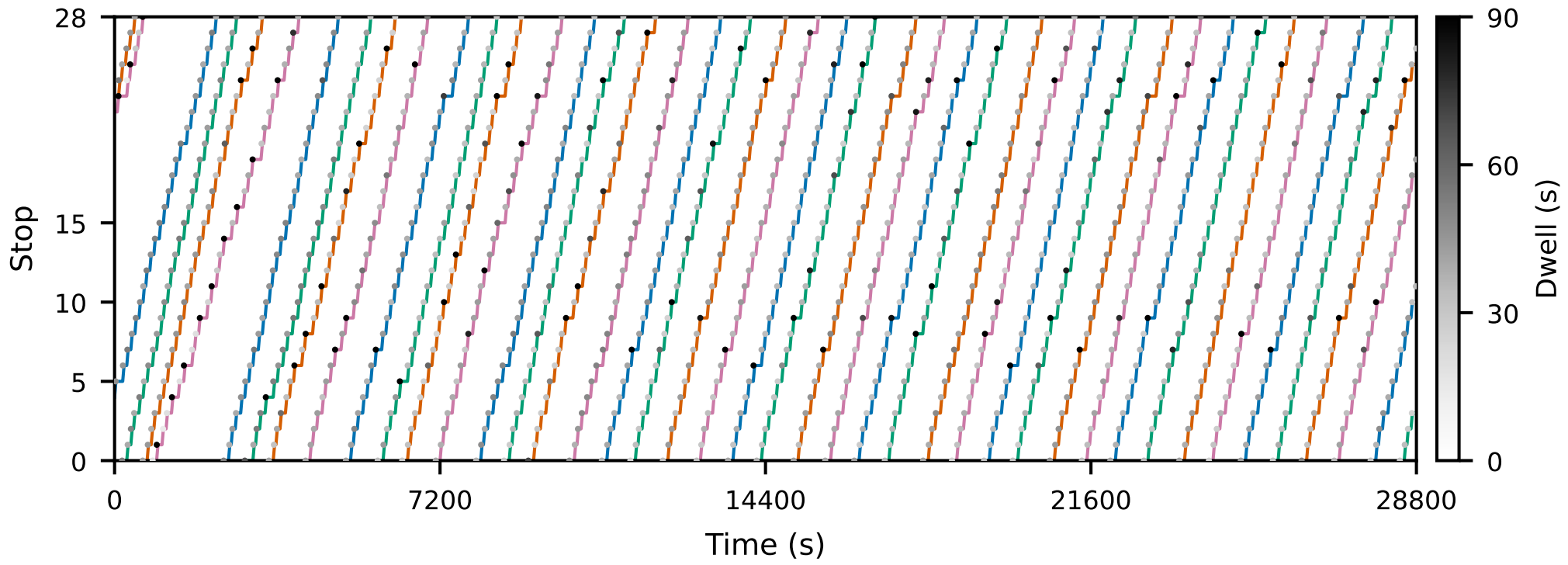}\\[2pt]
    \includegraphics[width=\linewidth,trim=420 0 0 182,clip]{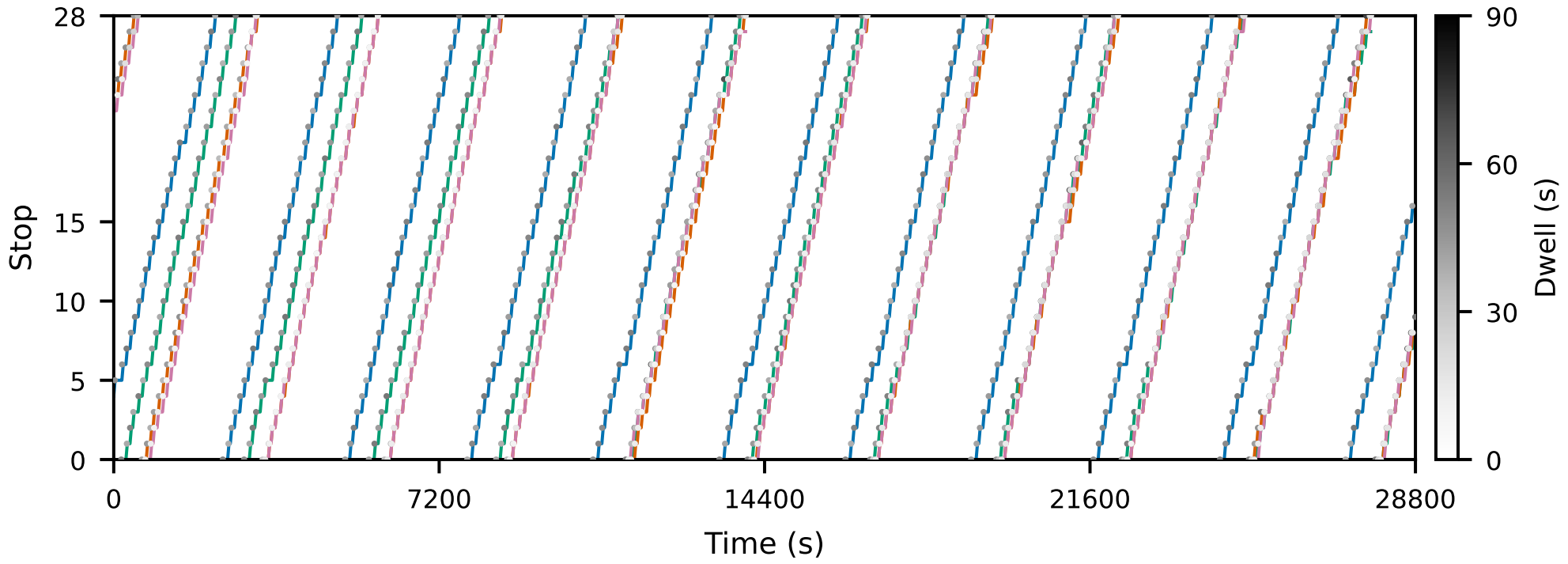}
    \caption{Bus trajectories on Beijing Route~598 over 8\,h: MEMBus (top) and zero additional holding (bottom).
  Colors identify vehicles; grayscale indicates total dwell.}
    \label{fig:memory-trajectories-8h}
  \end{figure}

  MEMBus waits 60\% less than KEEP, ReAct and DiLu at $h=8$ on all 8 routes and 59\% less on all 17 confirmation routes; ReAct and DiLu issue few holds and perform like KEEP. Against four memory-augmented agents given the same development evidence and three reflective or history agents it waits 43--60\% less on 8 routes (8/0/0, Holm $p=0.055$ across 7 contrasts), and 41--60\% less on the 17 confirmation routes (all 17/0/0, Holm $p=1.4{\times}10^{-4}$). Against the per-route tuned rule over 18 rule programs MEMBus waits 5.1\% less (6/0/2) and 8.5\% less on the confirmation routes (16/1/0); a rule tuned per route over the same 24 programs, including MEMBus's, picks MEMBus's program on all 8 routes (identical) and is level on the 17 routes ($-$0.36\%, 3/14/0; unconstrained argmin of development wait, whereas the MEMBus recommendation also requires served $\ge$ KEEP and CV $\le1.05\times$ KEEP), with no significant change in CV ($+$4.8\%, n.s., on the 8 routes). MPC is the strongest stationary controller: it waits 5.6\% less than MEMBus (better on 6 of 8 routes) and 4.8\% less on the confirmation routes, with a lower CV. DQN and CQL trained offline on the development seeds wait 5.6\% and 3.0\% longer than the tuned rule. The LLM chose the recommended program at all 1280 reviews (3840 executed decisions), and every further model that ran MEMBus on the 8 routes gave identical results (Table~\ref{tab:crossmodel-main}). Figures~\ref{fig:memory-trajectories} and~\ref{fig:memory-trajectories-8h} illustrate the resulting space--time vehicle trajectories on Beijing Route~598.

  \subsection{\texorpdfstring{Statistical Protocol}{Statistical Protocol}}
  \label{app:s11-stats}

  For each contrast, route means over target seeds are differenced and the exact two-sided sign-flip test enumerates all $2^n$ sign vectors of the $n$ route differences (the 1\,s tie threshold applies only to W/T/L counts; the sign-flip test uses the unrounded route means). Holm's procedure is applied within each pre-specified hypothesis family; families and their sizes are stated with each result. Relative differences are ratios of route-weighted levels (Eq.~\eqref{eq:rel-diff}), with hierarchical bootstrap intervals (routes, then seeds within route). With 8 routes the smallest $p$ is 0.0078; Holm over five, six or seven contrasts therefore needs 8/0/0 or 7/0/1 on most of them, giving a minimum adjusted $p=0.055$ for a seven-contrast family. The 17-route confirmation set provides higher route-level resolution for these multi-baseline families (smallest $p=1.5{\times}10^{-5}$). Contrasts outside a primary family are marked unadjusted.

  \subsection{\texorpdfstring{LLM Baselines at the Same Program Interface}{LLM Baselines at the Same Program Interface}}
  \label{app:s11-sameif}

  Table~\ref{tab:s11-sameif} compares MEMBus with the LLM baselines at the same program interface.

  \begin{table}[!htbp]
    \centering
    \caption{MEMBus against LLM baselines at the same program interface, $h=8$ wait.}
    \label{tab:s11-sameif}
    \scriptsize
    \setlength{\tabcolsep}{3pt}
    \resizebox{\linewidth}{!}{%
    \begin{tabular}{@{}lrrrr@{}}
      \toprule
      Contrast (A $-$ B) & A / B (s) & Change (\%) & W/T/L & $p$ \\
      \midrule
      \multicolumn{5}{@{}l}{\textit{Primary (Holm over 3)}} \\
      MEMBus $-$ ReAct-P & 424.9 / 676.0 & $-$37.1 & 8/0/0 & 0.023 \\
      MEMBus $-$ RAG-P & 424.9 / 553.0 & $-$23.2 & 8/0/0 & 0.023 \\
      MEMBus $-$ ExpeL-P & 424.9 / 429.8 & $-$1.1 & 3/5/0 & 0.25 \\
      \midrule
      \multicolumn{5}{@{}l}{\textit{Secondary (unadjusted)}} \\
      MEMBus $-$ Reflexion-P & 424.9 / 902.5 & $-$52.9 & 8/0/0 & 0.008 \\
      ReAct-P $-$ per-bus ReAct & 676.0 / 1072.1 & $-$36.9 & 8/0/0 & 0.008 \\
      ExpeL-P $-$ cards, no recommendation & 429.8 / 426.7 & $+$0.7 & 2/3/3 & 0.31 \\
      \midrule
      \multicolumn{5}{@{}l}{\textit{Independent API replication $-$ primary runs (unadjusted)}} \\
      MEMBus & 424.9 / 424.9 & 0.0 & 0/8/0 & 1.00 \\
      Per-bus ReAct & 1057.9 / 1072.1 & $-$1.3 & 5/2/1 & 0.16 \\
      Per-bus DiLu & 1062.1 / 1068.3 & $-$0.6 & 4/2/2 & 0.58 \\
      \bottomrule
    \end{tabular}}
    \par\smallskip{\footnotesize 8 routes $\times$ 5 seeds, \texttt{gpt-6-luna}; 280/280 runs accepted, no rejected runs. $p$: exact route-level sign-flip.}
  \end{table}

  \textbf{Design.} The four baselines (suffix -P) use MEMBus's interface: the same 7-program menu, review cadence, program executor and objective sentence. ReAct-P may call two observation tools before deciding and has no memory; ExpeL-P adds ExpeL insights extracted offline from the same development runs and the 28 raw paired (program, development day) records from which the MEMBus cards are compiled, with no summary or recommendation; RAG-P adds the 4 raw development tick records most similar to the current observation; Reflexion-P adds one in-episode reflection per review and no development memory. \textbf{Reading.} At the shared program interface, MEMBus reduces waiting by 37.1\% relative to memory-free ReAct-P (424.9 vs.\ 676.0\,s) and 23.2\% relative to RAG-P (553.0\,s), while matching ExpeL-P (429.8\,s, $-$1.1\%, n.s.) and unrecommended cards (426.7\,s, $+$0.7\%); MEMBus selects the evidence-best program in 100\% of reviews versus 71.5\% for ExpeL-P. Moving from per-bus ReAct (1072.1\,s) to ReAct-P (676.0\,s, 0 tool calls invoked) reduces waiting by 396.1\,s, and adding MEMBus memory reduces waiting by a further 251.1\,s ($-$37.1\%). \textbf{Implementation notes.} For the tool-capable -P arms, the reply parser accepts a bare JSON decision object as final when no tool call is issued, avoiding wrapper-schema mismatches. \textbf{Independent API replication.} To quantify live serving variance at temperature 0, we re-executed all 40 main-table runs of MEMBus, ReAct, and DiLu with full JSONL request/response capture (1.85\,GB archived). MEMBus is bit-exact across all 40/40 runs; ReAct and DiLu shift by $-$1.3\% and $-$0.6\% (n.s.; exact waits in 9 and 4 of 40 runs), leaving all conclusions unchanged ($-$59.8\% vs.\ ReAct and $-$60.0\% vs.\ DiLu, both 8/0/0).
  \textbf{17 confirmation routes.} On the 17 confirmation routes (\texttt{gpt-6-luna}, 5 seeds) MEMBus waits 35.8\%, 19.5\% and 53.1\% less than ReAct-P, RAG-P and Reflexion-P (17/0/0 each, Holm $p=6.1\times10^{-5}$) and 0.8\% less than ExpeL-P (11/3/3, $p=0.47$).

  \subsection{\texorpdfstring{Protocol and Implementation Notes}{Protocol and Implementation Notes}}
  \label{app:s11-disclosures}

  \begin{itemize}
  \item \textbf{Objective wording.} ReAct and DiLu receive MEMBus's objective sentence (low fixed-cohort waiting first, then low headway CV and high served fraction). On the auxiliary 10-route setting this alignment lowered ReAct's 8\,h wait by 0.4\% [$-$0.9, $-$0.1] (9/1/0, $p=0.002$) and DiLu's by 0.2\% ($p=0.22$, n.s.).
  \item \textbf{Learned baselines} were trained offline under a shared gradient budget on the timetable-grounded development seeds; on the largest-fleet routes CD50 and K12, DQN waits 523.8 and 440.4\,s versus 431.2 and 341.3\,s for the tuned rule.
  \item \textbf{Model exclusions:} Mistral Large 4, Nemotron Lightning, and DiffusionGemma did not pass initial JSON-schema or context-length pre-checks; Nemotron Bolt's per-bus ReAct is evaluated on 7 of the 8 main routes due to context overflow on one route.
  \item \textbf{Auxiliary paired records:} 12 per-bus paired records from two auxiliary routes are included in the shared prompt template of MEMBus, corresponding memory arms, and DiLu's retrieval bank; omitting them in cross-model checks leaves MEMBus unchanged at 424.9\,s.
  \item \textbf{Multi-model coverage:} on the 17-route extension, Nemotron Ultra 550B, GLM-5.3-Flash, and Nemotron-3.5-Super-VL run the full 5-seed design, whereas Laguna XS, Nemotron Bolt, and Nemotron Nano Omni are evaluated on a 17-route $\times$ 3-seed core subset, with Step 5 evaluated on the 8-route main set ($\approx 11$ tokens/s).
  \end{itemize}

  \subsection{\texorpdfstring{Released Supply Set (45 Routes)}{Released Supply Set (45 Routes)}}
  \label{app:supply-set}

  The benchmark also releases the supply data of 45 routes (stop sequence, road distance and the App timetable headways by time of day). Of these, 25 enter the experiments (8 main-table and 17 confirmation routes), 18 have complete data but were not selected under the route-screening rules, and 2 have an AM peak made up entirely of estimated departures and are excluded from simulation. Supply attributes for all 45 routes are included in the benchmark release. Headways are those of the single-day App timetable for direction A$\to$B between 07:00 and 09:00. Table~\ref{tab:supply-set} lists the routes.

  {\scriptsize
  \setlength{\tabcolsep}{4pt}
  \begin{longtable}{@{}llrrrl@{}}
  \caption{The 45 routes of the released supply set. Stops: catalog stops; length: road-network length; headway: AM-peak timetable headway (A$\to$B).}\label{tab:supply-set}\\
  \toprule
  City & Route & Stops & Length (km) & Headway (min) & Status \\
  \midrule
  \endfirsthead
  \toprule
  City & Route & Stops & Length (km) & Headway (min) & Status \\
  \midrule
  \endhead
  \bottomrule
  \endlastfoot
Shanghai & 50 & 26 & 18.5 & 8.6 & main \\
Beijing & 484 & 21 & 13.6 & 13.3 & main \\
Beijing & 546 & 30 & 16.5 & 7.5 & main \\
Beijing & 598 & 29 & 14.9 & 10.9 & main \\
Chengdu & 306 & 31 & 16.5 & 13.3 & main \\
Chengdu & 50 & 40 & 22.5 & 6.0 & main \\
Chengdu & 99 & 21 & 13.0 & 13.3 & main \\
Chengdu & K12 & 27 & 24.1 & 5.7 & main \\
Beijing & 2 & 25 & 15.0 & 7.1 & confirmation \\
Beijing & 337 & 22 & 13.9 & 7.5 & confirmation \\
Beijing & 342 & 28 & 16.5 & 12.0 & confirmation \\
Beijing & 386 & 41 & 24.8 & 15.0 & confirmation \\
Beijing & 402 & 29 & 18.1 & 12.0 & confirmation \\
Beijing & 425 & 32 & 19.1 & 8.6 & confirmation \\
Beijing & 556 & 20 & 10.7 & 8.6 & confirmation \\
Beijing & 55 & 16 & 8.2 & 13.3 & confirmation \\
Beijing & 58 & 17 & 13.5 & 8.6 & confirmation \\
Beijing & 597 & 32 & 17.6 & 8.0 & confirmation \\
Beijing & 663 & 32 & 19.5 & 9.2 & confirmation \\
Beijing & 671 & 31 & 26.3 & 9.2 & confirmation \\
Beijing & 820 & 39 & 38.2 & 10.9 & confirmation \\
Beijing & 997 & 43 & 32.3 & 15.0 & confirmation \\
Beijing & 9 & 27 & 20.2 & 10.9 & confirmation \\
Chengdu & 56 & 40 & 21.0 & 6.0 & confirmation \\
Chengdu & 61 & 40 & 21.4 & 10.9 & confirmation \\
Beijing & 117 & 28 & 14.0 & 17.1 & excluded: queue growth $>$25\% \\
Beijing & 1 (section) & 22 & 17.4 & 5.7 & excluded: queue growth $>$25\% \\
Beijing & 200 (outer ring) & 36 & 35.0 & 10.9 & excluded: queue growth $>$25\% \\
Beijing & 27 & 20 & 12.4 & 8.0 & excluded: queue growth $>$25\% \\
Beijing & 354 & 44 & 30.7 & 13.3 & excluded: queue growth $>$25\% \\
Beijing & 387 & 35 & 33.0 & 8.6 & excluded: queue growth $>$25\% \\
Beijing & 432 & 46 & 27.7 & 13.3 & excluded: queue growth $>$25\% \\
Beijing & 451 & 53 & 36.2 & 12.0 & excluded: queue growth $>$25\% \\
Beijing & 596 & 31 & 18.3 & 12.0 & excluded: queue growth $>$25\% \\
Beijing & 599 & 45 & 28.1 & 7.5 & excluded: queue growth $>$25\% \\
Beijing & 603 & 48 & 30.8 & 10.9 & excluded: queue growth $>$25\% \\
Beijing & 634 & 39 & 22.1 & 17.1 & excluded: queue growth $>$25\% \\
Beijing & 63 & 45 & 24.2 & 13.3 & excluded: queue growth $>$25\% \\
Beijing & 839 & 41 & 56.7 & 12.0 & excluded: queue growth, warm-up \\
Beijing & 913 & 92 & 62.7 & 20.0 & excluded: queue growth, warm-up \\
Beijing & 973 & 73 & 56.0 & 15.0 & excluded: queue growth, warm-up \\
Beijing & 300 (outer ring) & 36 & 48.1 & 8.0 & excluded: estimated AM peak \\
Chengdu & 47 & 47 & 24.0 & 5.7 & excluded: not in candidate pool \\
Chengdu & 4 & 30 & 19.2 & 7.1 & excluded: not in candidate pool \\
Chengdu & K2 & 30 & 30.9 & 3.9 & excluded: estimated AM peak \\
  \end{longtable}}

  \section{\texorpdfstring{Additional Diagnostic, Multi-Model, and Robustness Analyses}{Additional Diagnostic, Multi-Model, and Robustness Analyses}}
  \label{app:final}

  This appendix documents the diagnostic and robustness analyses supporting Sections~\ref{sec:protocol}, \ref{sec:membus-ablations}, and~\ref{sec:stale-experience}: the comparison between synthetic fixed fleets and timetable-grounded fleet sizing, the input/output schema and captured model calls, controller and lifecycle-learner parameters, per-metric paired tests, stale-experience and erroneous-injection scenarios, running-clock execution, offline-trained reinforcement-learning baselines, Voyager-style and MemGPT-style adaptations, multi-model replications, and simulator perturbation checks. All $p$-values are two-sided from exact route-level sign-flip tests; W/T/L counts routes on which the first listed arm is better / tied / worse; Holm adjustments apply within each pre-specified hypothesis family unless marked unadjusted.

  \subsection{\texorpdfstring{Motivation for Timetable-Grounded Fleet Sizing}{Motivation for Timetable-Grounded Fleet Sizing}}
  \label{app:oldtb}

  Comparing the auxiliary ten-route fixed-fleet subset (Appendices~\ref{app:membus-protocols}--\ref{app:supporting-evaluations}) with the primary timetable-grounded benchmark clarifies why realistic fleet sizing is essential for evaluating bus-holding agents. Under synthetic fixed six-bus fleets with TransitLM stop weights, four of the ten auxiliary routes operate in severe capacity saturation where zero additional holding is already near-optimal (599.20 vs.\ 609.53\,s at 1.5\,h and 2539.86 vs.\ 2672.54\,s at 8\,h), leaving minimal room for headway-holding policies to improve passenger waiting. Evaluating all arms under the same backbone (\texttt{gpt-6-luna}, 10 routes $\times$ 5 seeds; route-level exact sign-flip, Holm over 21 contrasts per window) confirms this structural ceiling: at 1.5\,h, two-stage MEMBus waits 600.4\,s, ReAct 600.1\,s, DiLu 599.9\,s, and zero holding 599.2\,s (Holm $p=1.0$); at 8\,h, the corresponding values are 2556.9, 2553.5, 2551.0, and 2539.9\,s ($+$0.1\% and $+$0.2\% vs.\ ReAct and DiLu; $+$0.7\% above zero holding, Holm $p=0.082$). Introducing the program-selection interface on the same fixed-fleet routes reduces 8\,h waiting to 2438.5\,s ($-$4.5\% vs.\ ReAct, $-$4.4\% vs.\ DiLu, $-$7.5\% vs.\ no-memory, and $-$2.4\% vs.\ CQL, 10/0 each, Holm $p=0.041$; $-$4.0\% vs.\ zero holding, 9/1, Holm $p=0.109$; $-$0.4\% vs.\ same-interface ExpeL-P, n.s., and $-$0.56\% vs.\ a per-route tuned rule with KEEP included in its grid, $-$13.8\,s, 6/4/0, $p=0.031$). Sizing fleets from observed AM-peak timetable headways (Appendix~\ref{app:s11-construction}) avoids artificial fleet starvation while preserving realistic bunching instability (uncontrolled KEEP waits 1066.0\,s versus 424.9\,s under MEMBus), enabling clean separation of action-interface design, static development evidence, and online experience lifecycle management.

  \subsection{\texorpdfstring{Input/Output Interface and Worked Example}{Input/Output Interface and Worked Example}}
  \label{app:io-example}

  \textbf{Scope.} Every reported MEMBus result on the primary benchmark uses \emph{one} centralized fleet controller consulted every 900\,s (every third 300\,s execution tick); between consultations the chosen program runs deterministically without an LLM call. Using the notation of Section~\ref{sec:problem}, Figure~\ref{fig:worked-example} illustrates a logged review and Table~\ref{tab:io-schema} lists all input, output, and executor fields.

\begin{figure}[!htbp]
\centering\scriptsize
\setlength{\tabcolsep}{2.5pt}\renewcommand{\arraystretch}{1.0}
\resizebox{\linewidth}{!}{%
\begin{tabular}{@{}p{0.13\linewidth}p{0.855\linewidth}@{}}
\toprule
\multicolumn{2}{@{}l}{\textbf{One MEMBus review} (route 484, seed 2061120120, $t{=}2370$\,s; real log)}\\
\midrule
$\mathcal H_t$ &
3 buses, 21 stops, target headway 411.4\,s; gaps to leader/follower: bus\_0 692.9/128.5\,s, bus\_1 128.5/149.2\,s, bus\_2 149.2/692.9\,s; 167 passengers queued, headway CV 0.81, no pending hold.\\
$\mathcal M_t$ &
7 program cards of route 484, each with 4 development-day deltas vs.\ KEEP; recommended \texttt{balance\_gain075\_cap90}: wait $-590.3$\,s, CV $-1.10$, served $+2.40$\,pp.\\
LLM out &
\texttt{\{"decision":"SELECT",}\newline\texttt{"program\_id":"balance\_gain075\_cap90",}\newline\texttt{"deviates\_from\_recommendation":false,\ldots\}}\\
$\mathbf{a}_t^{\mathrm{prop}}$, $\mathbf{a}_t^{\mathrm{exec}}$ &
Executor $h_k{=}\min(90,\max(0,0.75(g^{\mathrm{behind}}_k{-}g^{\mathrm{ahead}}_k)))$: bus\_0 CANCEL, bus\_1 SET 15.52\,s, bus\_2 SET 90\,s (capped); the same program is re-run on the updated gaps at 2670 and 2970\,s without an LLM call.\\
\bottomrule
\end{tabular}}
\caption[Worked example of one review in the notation of Section~\ref{sec:problem}.]{Worked example of one review in the notation of Section~\ref{sec:problem} (the LLM followed the recommendation; episode wait 441.9\,s against 1057.9\,s for ReAct on the same seed; schema fields in Table~\ref{tab:io-schema}). Here \texttt{target\_headway} ($411.4$\,s, giving bus\_2 \texttt{headway\_slack\_s} $=411.4-149.2=262.2$\,s) is the nominal free-flow spatial ring headway, distinct from the AM-peak timetable departure headway $H_{\mathrm{am}}=800$\,s in Table~\ref{tab:s11-routes}.}
\label{fig:worked-example}
\end{figure}

  \begin{table}[!htbp]
  \centering\scriptsize
  \caption{Input, LLM output, executor output, and episode summary fields for the program-selection interface.}
  \label{tab:io-schema}
  \setlength{\tabcolsep}{3pt}
  \begin{tabular}{@{}>{\raggedright\arraybackslash}p{0.27\linewidth}>{\raggedright\arraybackslash}p{0.13\linewidth}>{\raggedright\arraybackslash}p{0.08\linewidth}>{\raggedright\arraybackslash}p{0.46\linewidth}@{}}
  \toprule
  Field & Type & Unit & Example\\
  \midrule
  \multicolumn{4}{@{}l}{\emph{Input (one JSON user message)}}\\
  \texttt{route}, \texttt{time\_\allowbreak s} & str, int & s & "484", 2370\\
  \texttt{current\_\allowbreak programme} & str or null & -- & null (before the first selection: KEEP)\\
  \texttt{programme\_\allowbreak menu} & list of 7 objects & -- & \texttt{\{program\_\allowbreak id:\allowbreak "balance\_\allowbreak gain075\_\allowbreak cap90", gain:0.75, cap\_\allowbreak s:90\}}\\
  \texttt{previous\_\allowbreak programme\_\allowbreak reviews} & list & -- & [] (at 3270\,s: one entry with \texttt{applied}, \texttt{reason})\\
  \texttt{recommended\_\allowbreak programme} & str & -- & "balance\_\allowbreak gain075\_\allowbreak cap90"\\
  \texttt{current\_\allowbreak observation.\allowbreak buses[k]} & object & m, m/s, s & \texttt{\{id:\allowbreak "bus\_2", pos\_\allowbreak m:11350.0, speed\_\allowbreak mps:13.89, load\_\allowbreak fraction:0.303, last\_\allowbreak stop\_\allowbreak id:"S16", pending\_\allowbreak hold\_\allowbreak s:0.0, gap\_\allowbreak to\_\allowbreak ahead\_\allowbreak s:149.2, gap\_\allowbreak from\_\allowbreak behind\_\allowbreak s:692.9\}}\\
  \texttt{current\_\allowbreak observation.\allowbreak pax\_\allowbreak queues} & map stop$\to$float & pax & \{"S0":6.0, \ldots, "S20":0.0\}\\
  \texttt{legal\_\allowbreak hold\_\allowbreak menu.\allowbreak buses[k]} & object & s & \texttt{\{legal\_\allowbreak positive:true, max\_\allowbreak hold\_\allowbreak s:90.0, headway\_\allowbreak slack\_\allowbreak s:262.2\}}\\
  \texttt{programme\_\allowbreak memory[p].\allowbreak evidence[]} & list of 4 & s, --, pp & \texttt{\{evidence\_\allowbreak id:\allowbreak "DEV8H:484:\allowbreak \ldots:\allowbreak balance\_\allowbreak gain075\_\allowbreak cap90", wait\_\allowbreak s:-581.796, cv:-1.049, served\_\allowbreak pct:2.163\}}\\
  \texttt{operational\_\allowbreak history}, \texttt{physical\_\allowbreak feedback}, \texttt{recent\_\allowbreak arrival\_\allowbreak evidence}, \texttt{paired\_\allowbreak outcome\_\allowbreak memory} & structured, delta-coded & s & frames from 2190\,s; consumption receipts; 12 paired source facts\\
  \texttt{required\_\allowbreak output\_\allowbreak schema} & JSON schema & -- & 5 required keys, no extras\\
  \midrule
  \multicolumn{4}{@{}l}{\emph{LLM output (one JSON object)}}\\
  \texttt{decision} & enum & -- & "SELECT" or "CONTINUE"\\
  \texttt{program\_\allowbreak id} & enum (7) & -- & "balance\_\allowbreak gain075\_\allowbreak cap90"\\
  \texttt{deviates\_\allowbreak from\_\allowbreak recommendation} & bool & -- & false\\
  \texttt{evidence\_\allowbreak ids} & list[str] & -- & 4 \texttt{DEV8H:484:\ldots} ids\\
  \texttt{reason} & str & -- & "Follows the development recommendation: \ldots"\\
  \midrule
  \multicolumn{4}{@{}l}{\emph{Executor output $\to$ simulator (per bus)}}\\
  \texttt{operation} & enum & -- & CANCEL / SET / KEEP\\
  \texttt{hold\_\allowbreak s} & float or null & s & 15.52 (SET); null (CANCEL); $0<h\le 90$\\
  \midrule
  \multicolumn{4}{@{}l}{\emph{Episode result}}\\
  \texttt{wait\_\allowbreak s}, \texttt{cv}, \texttt{served\_\allowbreak fraction}, \texttt{terminal\_\allowbreak queue} & float & s, --, --, pax & 441.9, 0.225, 0.987, 125.7\\
  \bottomrule
  \end{tabular}
  \end{table}

  \subsection{\texorpdfstring{Captured Requests and Replies}{Captured Requests and Replies}}
  \label{app:samples}

  One real captured call for each of MEMBus, ReAct, DiLu and ExpeL-P (route 484, seed 2061120120, first call of the episode at $t=2370$\,s, \texttt{gpt-6-luna}, reasoning effort none, temperature 0, at most 4096 output tokens; JSON output enforced). Long texts are truncated and marked with the number of omitted characters; nothing else was edited. The MEMBus system prompt is shown in full; the other three are cut at 1500 characters (ReAct and DiLu share a longer protocol text, and ExpeL-P adds the ExpeL and ReAct instructions to the MEMBus prompt). ReAct may answer with a tool query before its final answer; this call answered directly.

  {\scriptsize
\textbf{MEMBus: 55342 input tokens, 168 output tokens, 15.6 s; the episode makes 32 model calls}
\begin{Verbatim}[breaklines=true,breakanywhere=true]
[system, 1475 characters]
You supervise headway control of one bus route in a SUMO simulation. You do not set per-bus holds
yourself: you choose which holding PROGRAMME the controller runs. The chosen programme is executed in
full for every bus at every scheduled tick until you change it. Objective: low fixed-cohort passenger
[... system prompt continues ...]
[user, 153577 characters; JSON]
{"controller_input":{"route":"484","time_s":2370,"protocol":"8h (28800 s, 96 decisions every 300 s, no c
ontinuation)","current_programme":null,"programme_menu":[{"program_id":"KEEP","kind":"KEEP","gain":0,"ca
[... user message continues ...]
[reply]
{"decision":"SELECT","program_id":"balance_gain075_cap90","deviates_from_recommendation":false,"evidence
_ids":["DEV8H:484:2061110120:balance_gain075_cap90","DEV8H:484:2061110121:balance_gain075_cap90","DEV8H:
484:2061110122:balance_gain075_cap90","DEV8H:484:2061110123:balance_gain075_cap90"],"reason":"Follows
the development recommendation: mean waiting delta -590.28 s, CV delta -1.095, and served fraction delta
+2.397 percentage points versus KEEP. Current headway CV is high (0.808); the transfer evidence is
route-matched development data, not a current forecast."}
\end{Verbatim}

\textbf{ReAct: 9585 input tokens, 139 output tokens, 7.6 s; the episode makes 96 model calls}
\begin{Verbatim}[breaklines=true,breakanywhere=true]
[system, 2959 characters]
Control all observed buses in the exclusive LLM control window [2370,31170). Choose KEEP, CANCEL, or any
continuous SET dose from 0 to 90 seconds for every bus. There is no fixed30 limit, no headway-slack dose
cap, and no high-load or late-ahead veto: these conditions are warnings for your judgement, not
[... system prompt continues ...]
[user, 13727 characters; JSON]
{"controller_input":{"time_s":2370,"past_trajectory":[],"current_observation":{"spatial_headways":[{"lea
der":"bus_2","follower":"bus_0","gap_s":692.9},{"leader":"bus_1","follower":"bus_2","gap_s":149.2},{"lea
[... user message continues ...]
[reply]
{"final":{"plans":{"bus_0":{"operation":"CANCEL","hold_s":null},"bus_1":{"operation":"CANCEL","hold_s":n
ull},"bus_2":{"operation":"CANCEL","hold_s":null}},"memory_refs":[],"reason":"All buses have zero
pending commands. Headways are highly uneven, with bus_0 trailing bus_2 by 692.9 s while bus_1 and bus_2
are closely spaced at 128.5 s and 149.2 s. Avoid adding discretionary dwell that could worsen bunching
or delay service; bus_1's current dwell cannot be changed by this decision."}}
\end{Verbatim}

\textbf{DiLu: 28302 input tokens, 167 output tokens, 10.2 s; the episode makes 49 model calls}
\begin{Verbatim}[breaklines=true,breakanywhere=true]
[system, 3365 characters]
Control all observed buses in the exclusive LLM control window [2370,31170). Choose KEEP, CANCEL, or any
continuous SET dose from 0 to 90 seconds for every bus. There is no fixed30 limit, no headway-slack dose
cap, and no high-load or late-ahead veto: these conditions are warnings for your judgement, not
[... system prompt continues ...]
[user, 68492 characters; JSON]
{"controller_input":{"current_observation":{"spatial_headways":[{"leader":"bus_2","follower":"bus_0","ga
p_s":692.9},{"leader":"bus_1","follower":"bus_2","gap_s":149.2},{"leader":"bus_0","follower":"bus_1","ga
[... user message continues ...]
[reply]
{"plans":{"bus_0":{"operation":"KEEP","hold_s":null},"bus_1":{"operation":"KEEP","hold_s":null},"bus_2":
{"operation":"KEEP","hold_s":null}},"memory_refs":[],"reason":"The current spatial gaps are highly
uneven, but bus_0 already has a large gap ahead and a comparatively tight gap behind, so holding it
could worsen bunching behind. Bus_1 is dwelling for 76 seconds with no pending command; its next stop
has a small queue, and an added hold could delay service. Bus_2 has a modest gap ahead and a large gap
behind, but low load and no evidence of an imminent need to hold. Keep all commands unchanged; current
geometry does not establish future stop-arrival CV or waiting outcomes."}
\end{Verbatim}

\textbf{ExpeL-P: 17502 input tokens, 176 output tokens, 5.9 s; the episode makes 32 model calls}
\begin{Verbatim}[breaklines=true,breakanywhere=true]
[system, 2184 characters]
You supervise headway control of one bus route in a SUMO simulation. You do not set per-bus holds
yourself: you choose which holding PROGRAMME the controller runs. The chosen programme is executed in
full for every bus at every scheduled tick until you change it. Objective: low fixed-cohort passenger
[... system prompt continues ...]
[user, 34188 characters; JSON]
{"controller_input":{"route":"484","time_s":2370,"protocol":"8h (28800 s, 96 decisions every 300 s, no c
ontinuation)","current_programme":null,"programme_menu":[{"program_id":"KEEP","kind":"KEEP","gain":0,"ca
[... user message continues ...]
[reply]
{"final":{"decision":"SELECT","program_id":"balance_cap90","evidence_ids":["DEV8H:484:2061110120:balance
_cap90","DEV8H:484:2061110121:balance_cap90","DEV8H:484:2061110122:balance_cap90","DEV8H:484:2061110123:
balance_cap90"],"reason":"The fleet has a severely oversized 692.9 s gap alongside compressed 128.5 s
and 149.2 s gaps, favoring corrective balancing. Across all four development days, balance_cap90 sharply
reduced waiting and CV while increasing served fraction versus KEEP; this fits the strongest insights
favoring selective correction with a larger cap. At this geometry its balancing rule acts on the bus
creating the oversized gap, not the compressed pairs."}}
\end{Verbatim}
  }

  \subsection{\texorpdfstring{Parameters of the Evaluated MEMBus Configuration}{Parameters of the Evaluated MEMBus Configuration}}
  \label{app:params}

  Tables~\ref{tab:membus-params} and~\ref{tab:learner-params} list the controller parameters and online lifecycle-learner thresholds of the evaluated MEMBus configuration.

  \begin{table}[!htbp]
  \centering\scriptsize
  \caption{Controller and memory-card parameters of the evaluated MEMBus configuration.}
  \label{tab:membus-params}
  \setlength{\tabcolsep}{3pt}
  \begin{tabular}{@{}>{\raggedright\arraybackslash}p{0.27\linewidth}>{\raggedright\arraybackslash}p{0.69\linewidth}@{}}
  \toprule
  Item & Value\\
  \midrule
  Program library (main tables) & 7 programs per route: KEEP, and balance programs with $(\mathrm{gain},\mathrm{cap}\,[\mathrm{s}])=(0.5,15),(0.5,30),(0.5,60),(0.5,90),(0.25,90),(0.75,90)$. For each bus and tick, $h=\min(\mathrm{cap},h_{\max},\max(0,\mathrm{gain}\times(g^{\mathrm{behind}}-g^{\mathrm{ahead}})))$; $h=0$ gives CANCEL; KEEP issues no command and leaves pending commands unchanged.\\
  Legal hold bound $h_{\max}$ & 90\,s (requested holding and total physical dwell, as in Section~\ref{sec:protocol}; \texttt{max\_hold\_s}\,$=90.0$ in the logged legal-hold menu)\\
  Review period / execution tick & 900\,s (LLM) / 300\,s (executor); 96 ticks and 32 reviews per 8\,h episode; scoring window $[t_0,t_0+28800)$\,s\\
  Recent history / online feedback window & 180\,s sampled every 5\,s / 120\,s starting at $t_0-120$\\
  Controllers; model & one centralized fleet controller; \texttt{gpt-6-luna}, reasoning effort none\\
  Cards & 7 per route; each carries 4 development-day paired differences against KEEP (seeds 2061110120--23) in wait ($\Delta W$), headway CV ($\Delta C$), and served fraction ($\Delta S$) with an evidence id, source seed, and route; compiled deterministically with no extra LLM call\\
  Recommendation & the program with the lowest mean development-seed wait, subject to served $\ge$ KEEP and CV $\le1.05\times$ KEEP\\
  Retrieval in the evaluated interface & all 7 cards of the route are shown without truncation or ranking (logged requests contain all seven); the general ranking rule of Section~\ref{sec:methods} applies when larger card banks exceed the display cap\\
  Paired source facts & 12 per-bus paired records from two auxiliary routes are included in the shared prompt template of MEMBus and the memory arms listed in Appendix~\ref{app:s11-disclosures}\\
  Archive timeout and active-view cap & 180\,s inactivity timeout and 4-procedure online-view cap apply in the auxiliary per-bus protocol (Appendix~\ref{app:membus-protocols}); in the 7-program interface all seven route cards remain visible\\
  \bottomrule
  \end{tabular}
  \end{table}

  In the evaluated 7-program interface, cards are fetched by route so retrieval ranking is not required; when free-form operational notices are supplied, the LLM parses them into structured operating conditions for retrieval (see Appendix~\ref{app:io-example} for the schema and worked example).

  \begin{table}[!htbp]
  \centering\scriptsize
  \caption{Default thresholds and transition rules of the online lifecycle learner.}
  \label{tab:learner-params}
  \setlength{\tabcolsep}{3pt}
  \begin{tabular}{@{}>{\raggedright\arraybackslash}p{0.16\linewidth}>{\raggedright\arraybackslash}p{0.18\linewidth}>{\raggedright\arraybackslash}p{0.13\linewidth}>{\raggedright\arraybackslash}p{0.47\linewidth}@{}}
  \toprule
  Mechanism & Parameter & Value & Meaning\\
  \midrule
  Common & window & 3 reviews & route queue $Q$ (pax) and route-wide arrival rate $\Lambda$ (pax/s) are averaged over the last 3 ticks; Little's-law wait proxy $W=Q/\Lambda$; evidence uses only current and past ticks\\
  Update & initial / maximal support & 3 / 3 & support of a card; supporting evidence adds 1 (capped), adverse evidence subtracts 1\\
  Update & $z_{\mathrm{adv}}$ & 4.0 & compiled card: standardized queue residual $z_t=|Q_t-\hat\mu_{Q,p}(t)|/\sigma_{Q,p}(t)>4.0$ from the card's predicted path counts as adverse\\
  Update & floor (relative / absolute) & 0.20 / 5.0 & effective scale $\sigma_{Q,p}(t)=\max(s_{Q,p}(t),\,0.20\,\hat\mu_{Q,p}(t),\,5.0)$\\
  Update & online-card tolerance & 0.15 & $W$ above 1.15 times the card's historical mean is adverse\\
  Update & drift tolerance & 0.10 & $W$ counts as stable evidence only if consecutive values differ by at most 10\%\\
  Suspension & trigger & support $\le0$ & the running card is not the reference, and no trial is in progress; the card is kept as SUSPENDED\\
  Probing & candidates / maximum / settle, block & route-specific list / 4 / 1, 1 review & each candidate is settled for one review and measured for one; accepted if at least 10\% below the baseline, else the best candidate if not worse than the baseline, else the reference (KEEP)\\
  Reactivation & $z_{\mathrm{react}}$ / reviews / probation & 2.0 / 2 / 2 & a suspended compiled card returns when the observed queue under the running program matches the card's own predicted path ($z\le2.0$) on 2 consecutive reviews; one adverse review within 2 reviews of return resets its support\\
  Consolidation & merge rule & as Section~\ref{sec:membus-ablations} & when an online card is created while the compiled card is suspended, both are merged into one two-member card that applies the member whose prediction matches the observed queue\\
  Deletion & archival retention & raw archive kept & removes only the compiled card while retaining raw records in the archive; neither the scripted learner nor the evaluated 8\,h episodes trigger card deletion\\
  \bottomrule
  \end{tabular}
  \end{table}

  The thresholds in Table~\ref{tab:learner-params} are the defaults of the lifecycle learner; the ablation ladder arms of Appendix~\ref{app:stale} toggle only the mechanism switches (update, suspend, reactivate, consolidate) alongside the route-specific candidate list, and the 25-route lifecycle evaluation uses identical harness, learner, and memory defaults.

  \paragraph{Reproduction notes.} The environment uses SUMO 1.25.0 with the TraCI interface; matching the SUMO version is required for bit-exact trajectory reproduction. In deterministic replay checks, our release package replayed six full simulations bit-exactly (decisions, waiting, and CV identical) and reproduced all 40 of 40 MEMBus main-table cells. Fresh online LLM calls are subject to provider-side nondeterminism: when the main-table cases were re-executed live, waiting matched identically in 9 of 40 cases for ReAct and 4 of 40 for DiLu, whereas MEMBus's program-selection path remained identical across all 40 of 40 cases. Archived requests and replies allow deterministic replay of individual decisions (Appendices~\ref{app:io-example} and~\ref{app:samples}).

  \subsection{\texorpdfstring{Per-Metric Paired Tests on the Timetable-Grounded Benchmark}{Per-Metric Paired Tests on the Timetable-Grounded Benchmark}}
  \label{app:permetric}

  Table~\ref{tab:permetric} reports per-metric paired tests across waiting, headway CV, served fraction, and bunching events.

  \begin{table}[!htbp]
    \centering
    \caption{MEMBus $-$ baseline per metric at $h=8$ (8 routes $\times$ 5 seeds, \texttt{gpt-6-luna}).}
    \label{tab:permetric}
    \scriptsize
    \setlength{\tabcolsep}{3pt}
    \resizebox{\linewidth}{!}{%
    \begin{tabular}{@{}lrrrr@{}}
      \toprule
      Baseline & Wait \% (W/T/L) & CV \% & Served (pp) & Bunching \% \\
      \midrule
      \multicolumn{5}{@{}l}{\textit{Original-interface LLM baselines (per-bus plans)}} \\
      ReAct & $-$60.4 (8/0/0) & $-$72.8 & $+$2.89 & $-$97.1 \\
      DiLu & $-$60.2 (8/0/0) & $-$72.9 & $+$2.89 & $-$97.1 \\
      ExpeL & $-$43.0 (8/0/0) & $-$58.2 & $+$1.48 & $-$93.9 \\
      RAG over records & $-$52.8 (8/0/0) & $-$66.7 & $+$2.21 & $-$96.0 \\
      Reflexion & $-$59.9 (8/0/0) & $-$72.9 & $+$2.82 & $-$97.1 \\
      Self-Refine, Recent-history, Generative Agents, MemGPT-style & $-$55.3 to $-$60.4 (8/0/0) & $-$69.1 to $-$73.1 & $+$2.70 to $+$2.90 & $-$96.6 to $-$97.1 \\
      \multicolumn{5}{@{}l}{\textit{Same-interface baselines (program selection, same menu, cadence, executor and evidence)}} \\
      ExpeL-P & $-$1.1 (3/4/1) & $-$3.5 & $+$0.01 & $-$23.4 \\
      RAG-P & $-$23.2 (8/0/0) & $-$37.2 & $+$0.49 & $-$86.1 \\
      ReAct-P & $-$37.1 (8/0/0) & $-$49.2 & $+$1.01 & $-$93.2 \\
      Reflexion-P & $-$52.9 (8/0/0) & $-$66.0 & $+$2.22 & $-$96.0 \\
      \multicolumn{5}{@{}l}{\textit{Learned baselines and rules tuned per route offline (foreknowledge reference)}} \\
      DQN & $-$10.1 (6/0/2) & $-$24.8 & $+$0.32 & $-$82.9 \\
      CQL & $-$7.8 (8/0/0) & $-$28.4 & $+$0.09 & $-$85.8 \\
      Headway feedback (tuned) & $+$0.3 (3/1/4) & $+$4.5 & $-$0.05 & $+$66.6 \\
      Tuned rule (18 programs) & $-$5.1 (6/0/2) & $+$4.8 & $-$0.03 & $+$2.1 \\
      \bottomrule
    \end{tabular}}
    \par\smallskip{\footnotesize Relative difference in the route-weighted mean (negative: MEMBus is lower; for served, percentage points, positive: MEMBus is higher). Route-level exact sign-flip test, two-sided; with 8 routes the smallest attainable $p$ is 0.0078, reached by every 8/0/0 cell. On waiting, CV, served fraction and bunching MEMBus is 8/0/0 (served: 7/0/1 for RAG-P) against every original-interface baseline and against RAG-P, ReAct-P and Reflexion-P; it is level with ExpeL-P and with the tuned rules on all four. Holm correction over all nine original-interface baselines yields smallest adjusted $p=0.070$ on 8 routes (versus Holm $p=0.039$ for the 5-contrast primary family in Table~\ref{tab:s11-main}); on the 17 confirmation routes across three additional backbones, the smallest attainable $p$ is $1.5{\times}10^{-5}$ (Holm $7.6{\times}10^{-5}$, Appendix~\ref{app:xmodel}). For bunching, the tuned headway-feedback rule has a lower point estimate (10.8 against 18.0), not significant at route level.}
  \end{table}

  \subsection{\texorpdfstring{Stale-Experience Scenarios, Ablation Ladder, and Mechanisms}{Stale-Experience Scenarios, Ablation Ladder, and Mechanisms}}
  \label{app:stale}

  \subsubsection{\texorpdfstring{Regime-Shift Design}{Regime-Shift Design}}
  \label{app:stale-design}

  \textbf{Notation.} The stored program is the program stored for regime A (the program that is best in the development data of the route; ``pA'' in Figure~\ref{fig:regime}); the regime-B program is the best program for regime B; the wrong program is the worst non-KEEP program in the development cards. KEEP denotes no control (no holding). The stale controller uses the stored program all day. The oracle is the scripted switch with hindsight knowledge of the regime (the best program in each phase). Phases correspond to the windows in Eq.~\eqref{eq:regime-windows}: A1 ($\mathcal W_{\mathrm{A1}}$, first hour), B ($\mathcal W_{\mathrm{B}}$, changed regime, 4\,h), and A2 ($\mathcal W_{\mathrm{A2}}$, last 3\,h; ``recovery phase'' or ``returned phase'' when the earlier regime returns). ``Full system'' denotes MEMBus with update, suspension, reactivation, and consolidation enabled. All tests are route-level, two-sided, exact sign-flip tests.

  \textbf{Design.} Each episode follows A (1\,h), B (4\,h), A (3\,h) with 96 ticks and 32 reviews. In B the program stored for A is at least 10\% worse than the best alternative while remaining better than not controlling. The regime-change time is set exogenously and never disclosed to any agent: the memory layer detects the shift from online evidence and updates its recommendation, while the LLM reads the evidence, executes, and can override clearly uncalibrated recommendations (without the memory layer, an LLM given a stale recommendation and live feedback waits 22.8\% more than the learner, 0/0/16 routes). While outdated context can degrade performance below uncontrolled baselines in general non-stationary settings~\citep{runding2026}, across seven perturbation families on closed-loop bus rings even mistuned headway-balancing programs remain preferable to uncontrolled bunching.

  \subsubsection{\texorpdfstring{Injected Erroneous Experience}{Injected Erroneous Experience}}
  \label{app:inject}

  To test robustness under actively misleading prior records, we evaluated two erroneous-experience injection scenarios. In both scenarios, the worst non-KEEP program on development data (\texttt{balance\_cap15} on 13 of 16 routes, \texttt{balance\_cap30} on 3) is forged to appear superior to every other program, including the regime-B optimum: its four development records are overwritten by those of the regime-B program with improvements magnified by $1.2\times$, card summaries are recomputed, and its forged development gain becomes $-$125 to $-$1390\,s (versus its true $-$112 to $+$16\,s). Both MEMBus and the ExpeL-style baseline receive the same forged evidence in their native representations: for MEMBus, the forged card becomes the compiled start card and carries the regime-B predicted queue trajectory; for the ExpeL-style agent, the four raw records are replaced and a matching ExpeL insight is added. All prompts, menus, executors, and windows match the 25-route experiment across the 16 screened routes $\times$ 3 target seeds (96 cells, 3,072 reviews, \texttt{gpt-6-luna}, 0 failed calls), compared against the uninjected runs on the same routes and seeds.

  \textbf{Injection at episode start.} When the forged card is present from the start of A1, both arms incur higher waiting over A1+B relative to their uninjected counterparts (MEMBus $+39.6\%$, $+265.7$\,s; ExpeL-style $+30.6\%$, $+200.1$\,s; both 1/0/15, Holm $1.8\times10^{-4}$), with primary A1+B contrast MEMBus $-$ ExpeL-style equal to $+31.4$\,s ($+3.05\%$, route bootstrap 95\% CI $[+1.02,+4.92]$, 3/0/13, $p=0.0050$, Holm $0.0050$ over three tests) before MEMBus recovers in A2 ($-16.65\%$, Table~\ref{tab:inject-start}).

  \begin{table}[!htbp]
    \centering
    \caption{Wrong experience injected at episode start: MEMBus $-$ ExpeL-style, relative difference in waiting (\%), 16 routes $\times$ 3 seeds.}
    \label{tab:inject-start}
    \scriptsize
    \setlength{\tabcolsep}{3pt}
    \begin{tabular}{@{}lrrrr@{}}
      \toprule
      Window & Difference (\%) & Better/tied/worse & $p$ & Role \\
      \midrule
      A1 and B (first 5\,h) & $+$3.05 & 3/0/13 & 0.0050 & primary \\
      A1 (1\,h) & 0.00 & 0/16/0 & --- & descriptive \\
      B (4\,h) & $+$3.52 & --- & 0.0050 & descriptive \\
      A2 (3\,h) & $-$16.65 & 12/0/4 & 0.0051 & descriptive, uncorrected \\
      Whole day & $-$3.42 (CI $-$7.64 to $+$0.51) & --- & 0.10 & descriptive, n.s. \\
      \midrule
      \multicolumn{5}{@{}l}{\textit{Injected $-$ not injected (same arm), A1+B / A2}} \\
      MEMBus & $+$39.6 / $+$48.6 & 1/0/15 (A1+B) & Holm $1.8\times10^{-4}$ & secondary \\
      ExpeL-style & $+$30.6 / $+$72.3 & 1/0/15 (A1+B) & Holm $1.8\times10^{-4}$ & secondary \\
      \bottomrule
    \end{tabular}
    \par\smallskip{\footnotesize A1 is identical (0.00\%) because both arms run the forged program on 100\% of A1 reviews under the same executor and seeds. Without injection, MEMBus $-$ ExpeL-style on the same routes and seeds is $-$0.14\%, $-$4.61\%, $-$10.19\%, and $-$5.98\% in A1, B, A2, and over the whole day.}
  \end{table}

  Inspecting per-review card states (48 cells per arm) explains this trajectory. MEMBus's suspension rule triggered on adverse evidence in 43 of 48 cells (89.6\%), with a median first-suspension review of 9 (range 7--13, roughly two hours into the episode). In 42 cells (87.5\%) the forged card was subsequently reactivated (1.58 times on average), giving forged-program selection shares of 100\%, 42.2\%, and 62.3\% in A1, B, and A2 (57.0\% overall) versus 100\%, 99.2\%, and 100\% (46 of 48 cells, 99.6\% overall) for the ExpeL-style agent; trajectory-agreement reactivation ($z\le 2.0$) matches observed queues against the card's stored prediction path, incurring $+3.5\%$ probing and switching overhead during B before reducing waiting by $16.65\%$ in A2.

  \textbf{Injection at the onset of phase B.} When the erroneous experience is injected at the 5th review (onset of B, coinciding with a 30\% demand increase across all 48 cells), the primary contrast (MEMBus $-$ ExpeL-style over B+A2 weighted by demand $\times$ duration) is $-68.7$\,s ($-7.63\%$, CI $[-16.58,+1.07]$, 8/0/8, $p=0.069$), with A2 waiting reduced by $19.56\%$ ($p=0.0097$; Table~\ref{tab:inject-b}).

  \begin{table}[!htbp]
    \centering
    \caption{Wrong experience injected at the start of phase B: relative difference in waiting (\%), 16 routes $\times$ 3 seeds.}
    \label{tab:inject-b}
    \scriptsize
    \setlength{\tabcolsep}{3pt}
    \resizebox{\linewidth}{!}{%
    \begin{tabular}{@{}llrrrr@{}}
      \toprule
      Contrast & Window & Difference (\%) & 95\% CI & Better/tied/worse & $p$ \\
      \midrule
      MEMBus $-$ ExpeL-style & B and A2 (primary) & $-$7.63 & $[-16.58,+1.07]$ & 8/0/8 & 0.069 \\
      MEMBus $-$ ExpeL-style & B & $-$0.26 & $[-8.07,+7.62]$ & 8/0/8 & 0.77 \\
      MEMBus $-$ ExpeL-style & A2 & $-$19.56 & $[-32.44,-6.82]$ & 10/0/6 & 0.0097 \\
      MEMBus $-$ ExpeL-style & A1 & $+$0.21 & $[-0.05,+0.59]$ & 1/6/9 & 0.17 \\
      MEMBus $-$ ExpeL-style & whole day & $-$7.21 & $[-15.65,+0.94]$ & 8/0/8 & 0.071 \\
      MEMBus, injected $-$ not injected & B and A2 & $+$5.28 & $[+0.08,+11.29]$ & 6/0/10 & 0.14 \\
      MEMBus, injected $-$ not injected & A2 & $+$16.57 & $[+6.28,+28.60]$ & 2/0/14 & 0.0071 \\
      ExpeL-style, injected $-$ not injected & B and A2 & $+$13.50 & $[-3.16,+32.10]$ & 6/0/10 & 0.35 \\
      ExpeL-style, injected $-$ not injected & A2 & $+$47.83 & $[+18.13,+79.38]$ & 4/0/12 & 0.028 \\
      \bottomrule
    \end{tabular}}
    \par\smallskip{\footnotesize The first row is the primary test; remaining rows are descriptive and unadjusted. Without injection on the same routes and seeds, MEMBus $-$ ExpeL-style is $-$6.18\% over B and A2 (13/0/3, $p=0.0020$). Leave-one-route-out primary differences range from $-$9.32\% to $-$5.39\%.}
  \end{table}

  In this second scenario, the forged card was retired in all 48 of 48 cells within three to four reviews after injection and never reactivated; in 47 of 48 cells, retirement occurred via \emph{supersession} when the initially suspended stored-program card matched two consecutive reviews at review 7 and resumed active status. After learner-level retirement, forged-program selection drops to 47.3\% in B and 44.6\% in A2, compared with 95.4\% and 92.0\% for ExpeL-style (MEMBus restores the stored program in 49.3\% of A2 reviews versus 5.9\% for ExpeL-style).

  \textbf{Diagnostic scope.} Under self-consistent forged records, trajectory-agreement reactivation delays phase-B switching before recovering in $\mathcal W_{\mathrm{A2}}$, whereas five of seven backbones override inconsistent recommendations in 85\%--100\% of reviews when card records remain genuine (Appendix~\ref{app:xmodel}).

  \subsubsection{\texorpdfstring{Results on 25 Routes, Ablation Ladder, and Management Removed}{Results on 25 Routes, Ablation Ladder, and Management Removed}}
  \label{app:ladder}

  Table~\ref{tab:all25} evaluates MEMBus against the same-interface ExpeL-style baseline across all 25 unscreened routes and a 7-route confirmation set; Table~\ref{tab:ladder} reports the additive mechanism ladder and leave-one-out contrasts; and Table~\ref{tab:mbstatic} compares full MEMBus against static cards with lifecycle management disabled.

  \begin{table}[!htbp]
    \centering
    \caption{MEMBus against the same-interface ExpeL-style memory under stale experience.}
    \label{tab:all25}
    \scriptsize
    \setlength{\tabcolsep}{3pt}
\resizebox{\linewidth}{!}{%
    \begin{tabular}{@{}lrrrr@{}}
      \toprule
      Set (routes $\times$ seeds) & Recovery phase \% & Whole day \% & W/T/L (day) & $p$ (day) \\
      \midrule
      25 routes not screened by a development rule ($\times$5, new seeds) & $-$5.88 ($p=0.0042$, Holm 0.0127) & $-$3.00 & 15/1/9 & 0.0054 \\
      \quad 16 routes selected by the development rule & $-$9.05 & $-$4.57 & --- & 0.0050 \\
      \quad 9 routes not selected & $-$0.24 & $-$0.22 & --- & 0.78 \\
      7 further confirmation routes ($\times$8) & $-$3.04 (6/1; $p=0.0469$, Holm 0.094) & $-$2.81 & 7/0/0 & 0.0156 \\
      \bottomrule
    \end{tabular}}
    \par\smallskip{\footnotesize Recovery phase: $\mathcal W_{\mathrm{A2}}$ (last 3\,h), fixed-cohort restricted mean wait, mean of route-relative differences. Within the 3-test hypothesis family, whole-day waiting is primary and recovery-phase waiting is secondary (Holm 0.0127). Manageable headroom relative to the hindsight-optimal switch is small on the 9 unselected routes (2.3\% for MEMBus and 2.5\% for ExpeL-style, versus 12.9\% on the 16 selected routes). On two-bus saturated routes SZM425 and SH152, heavy holding exceeds KEEP waiting for both agents, with MEMBus incurring roughly half the excess of ExpeL-style ($+$128\,s vs.\ $+$284\,s). In the recovery phase across 25 routes, MEMBus closes 62\% of the gap between ExpeL-style and the hindsight switch (ratio of totals, bootstrap 95\% CI 39\%--76\%) and saves 35.5\,s per passenger (bootstrap 95\% CI 14--59\,s; route median 9.8\,s). For magnitude context, the top two learned bus-holding policies in Table~2 of \citealp{wang2021asynchronous} (CAAC vs.\ MADDPG across three transfer routes) differ by 2.6\%--6.9\%. Holding volume is analyzed in Appendix~\ref{app:holding}.}
  \end{table}

  \begin{table}[!htbp]
    \centering
    \caption{Ablation ladder from a DiLu-style agent (16 routes $\times$ 5 seeds, \texttt{gpt-6-luna}, 23,040 reviews, 0 failures).}
    \label{tab:ladder}
    \scriptsize
    \setlength{\tabcolsep}{3pt}
    \resizebox{\linewidth}{!}{%
    \begin{tabular}{@{}lrrrr@{}}
      \toprule
      Contrast & Whole day \% (W/T/L) & Holm & Recovery phase \% & Holm \\
      \midrule
      \multicolumn{5}{@{}l}{\textit{Additive ladder (Holm over 6 tests)}} \\
      Static cards + recommendation $-$ DiLu-style & $-$2.64 (9/0/7) & 0.81 & --- & --- \\
      + update & $-$3.29 (14/0/2) & 0.0022 & $-$1.82 & 0.59 \\
      + suspend & $-$0.99 (10/0/6) & 0.81 & --- & --- \\
      + reactivate & $-$1.12 (9/0/7) & 0.81 & $-$7.12 & 0.063 \\
      + consolidate (full system) & $+$0.38 (7/0/9) & 0.81 & --- & --- \\
      Full system $-$ ExpeL-style & $-$3.72 (13/0/3) & 0.0345 & $-$9.31 (14/0/2) & 0.0037 \\
      \multicolumn{5}{@{}l}{\textit{Leave-one-out (Holm over 4 tests; positive: removing the mechanism worsens)}} \\
      Without suspend, reactivate and consolidate (bundle) & $+$2.08 (4/0/12) & 0.021 & $+$9.31 (0/0/16) & $1.2{\times}10^{-4}$ \\
      Without reactivation & $+$1.80 & 0.76 & $+$10.21 & 0.051 \\
      Without update & $-$0.78 & 1.0 & $+$3.56 & --- \\
      Without consolidation & $-$0.33 & 1.0 & --- & --- \\
      \bottomrule
    \end{tabular}}
    \par\smallskip{\footnotesize Mean whole-day waits (s): DiLu-style 701.4, static cards 669.8, +update 646.3, +suspend 637.7, +reactivate 630.8, full system 632.9, ExpeL-style 665.4, stale program 674.7, hindsight oracle 560.3, KEEP 1021.7. The two magnitudes 9.31\% (full system $-$ ExpeL-style in A2, $-$9.31\%; bundle removal, $+$9.31\%) are independent contrasts that coincide numerically. Updating also reduces phase-B waiting by 3.90\% (15/0/1, Holm 0.0005), and full system $-$ DiLu-style is $-$7.86\% over the day (13/0/3, descriptive). Because reactivation and consolidation are inactive without suspension, the three mechanisms form a coupled lifecycle bundle; with 16 routes (per-route recovery SD 52--63\,s, 80\%-power detectable effect 9\%--11\%), the scripted $2^4$ factorial in Appendix~\ref{app:factorial} isolates individual main effects and interactions without LLM sampling variance.}
  \end{table}

  \begin{table}[!htbp]
    \centering
    \caption{Lifecycle management ablation: MEMBus with and without online management modules (25 unscreened routes and 16-route screened subset).}
    \label{tab:mbstatic}
    \scriptsize
    \setlength{\tabcolsep}{3pt}
    \begin{tabular}{@{}>{\raggedright\arraybackslash}p{0.33\linewidth}>{\raggedright\arraybackslash}p{0.16\linewidth}>{\raggedright\arraybackslash}p{0.16\linewidth}>{\raggedright\arraybackslash}p{0.16\linewidth}>{\raggedright\arraybackslash}p{0.10\linewidth}@{}}
      \toprule
      Contrast & Whole day \% & Recovery phase \% & Phase B \% & $p$ \\
      \midrule
      16 screened routes: static cards $-$ ExpeL-style & $+$1.70 ($+$7.9\,s) & --- & --- & 0.26 \\
      16 screened routes: full system $-$ static cards & $-$5.9 (15/0) & $-$11.1 (15/0) & $-$4.3 (15/0) & 0.0001 each \\
      Full system $-$ static cards, 25 routes $\times$ 5 seeds, unscreened & $-$4.07 (18/4/3; Holm $3.2\times10^{-5}$) & $-$6.65 (17/4/4; $p=0.0003$, primary) & $-$3.36 (20/4/1; Holm $1.7\times10^{-5}$) & --- \\
      Static cards $-$ ExpeL-style, same 25 routes & $+$1.10 & $+$0.80 & $+$1.46 & n.s. \\
      \quad 9 unselected routes, full system $-$ static cards & $-$0.49 & $-$0.99 & --- & n.s. \\
      \bottomrule
    \end{tabular}
    \par\smallskip{\footnotesize Static cards share the exact static evidence, recommendation rule, cards, model, menu, and prompts of full MEMBus, differing only in disabling online update, suspension, reactivation, and consolidation. Across the 25-route comparison (4,000 reviews, 0 failures), phase A1 is bit-identical and leave-one-route-out recovery differences lie between $-$7.14\% and $-$5.76\%, saving 35.5\,s per passenger in $\mathcal W_{\mathrm{A2}}$. Restricting updates strictly to executed probe trials yields a $-$5.58\% recovery-phase bundle contrast.}
  \end{table}

  \subsubsection{\texorpdfstring{Full Factorial Over the Four Mechanisms}{Full Factorial Over the Four Mechanisms}}
  \label{app:factorial}

  Table~\ref{tab:factorial} reports the Shapley main effects from the complete $2^4$ factorial over update (U), suspension (S), reactivation (R), and consolidation (C) using the deterministic lifecycle learner (no LLM): 16 configurations $\times$ 16 ladder routes $\times$ 5 seeds $=1,280$ cells, all completed, with the 8 ladder-shared configurations (640 cells) matching bit-exactly (640/640). Route means over seeds form the unit of analysis; Shapley main effects are tested via exact two-sided route-level sign-flip tests with Holm adjustment ($m=4$), while pairwise Shapley interaction indices (SII) are exploratory.

  \begin{table}[!htbp]
    \centering
    \caption{Shapley main effects of the four lifecycle mechanisms (scripted learner, 16 routes $\times$ 5 seeds).}
    \label{tab:factorial}
    \scriptsize
    \setlength{\tabcolsep}{3pt}
    \resizebox{\linewidth}{!}{%
    \begin{tabular}{@{}llrrrrrr@{}}
      \toprule
      Window & Mechanism & Shapley (s) & Share of total & 95\% CI & Better/worse routes & $p$ & Holm ($m=4$) \\
      \midrule
      \multicolumn{8}{@{}l}{\textit{Whole day (all off 674.7\,s, all on 611.2\,s, total $-$63.5\,s)}} \\
      & Update & $-$11.0 & 17\% & $[-20.2,-2.1]$ & 12/4 & 0.0359 & 0.0359 \\
      & Suspension & $-$39.0 & 61\% & $[-60.8,-18.0]$ & 12/4 & 0.0038 & 0.0114 \\
      & Reactivation & $-$14.2 & 22\% & $[-23.6,-6.2]$ & 13/3 & 0.0027 & 0.0109 \\
      & Consolidation & $+$0.7 & $-$1\% & $[+0.2,+1.2]$ & 3/10 & 0.0161 & 0.0322 \\
      \multicolumn{8}{@{}l}{\textit{Recovery phase A2 (all off 460.8\,s, all on 386.7\,s, total $-$74.2\,s)}} \\
      & Update & $-$37.8 & 51\% & $[-53.9,-20.3]$ & 14/2 & 0.0016 & 0.0047 \\
      & Suspension & $+$12.5 & $-$17\% & $[-27.1,+50.7]$ & 7/9 & 0.54 & 0.54 \\
      & Reactivation & $-$50.3 & 68\% & $[-74.1,-30.5]$ & 16/0 & $<$0.0001 & 0.0001 \\
      & Consolidation & $+$1.5 & $-$2\% & $[+0.3,+2.7]$ & 3/10 & 0.0312 & 0.0625 \\
      \bottomrule
    \end{tabular}}
  \end{table}

  Whole-day main effects of update, suspension, and reactivation are all negative and Holm-significant. In the recovery phase $\mathcal W_{\mathrm{A2}}$, update ($-$37.8\,s) and reactivation ($-$50.3\,s, 16/16 routes) drive the reduction, whereas suspension alone increases recovery waiting ($+$126\,s relative to all-off when paired without reactivation) and exhibits strong complementarity with reactivation (negative SII with 95\% CI excluding zero in both windows). Update partially substitutes for suspension and reactivation (positive whole-day SII), and suspension plus reactivation without update attains the lowest scripted whole-day mean (563.9\,s, matching the leave-one-out arm of Table~\ref{tab:ladder}). Consolidation has a near-zero main effect ($+0.7$\,s whole day, $+1.5$\,s in A2; $<2\%$ of total effect).

  \subsubsection{\texorpdfstring{Regime-Resolved Behavior}{Regime-Resolved Behavior}}
  \label{app:regime}

\begin{figure}[t]
    \centering
    \includegraphics[width=\linewidth]{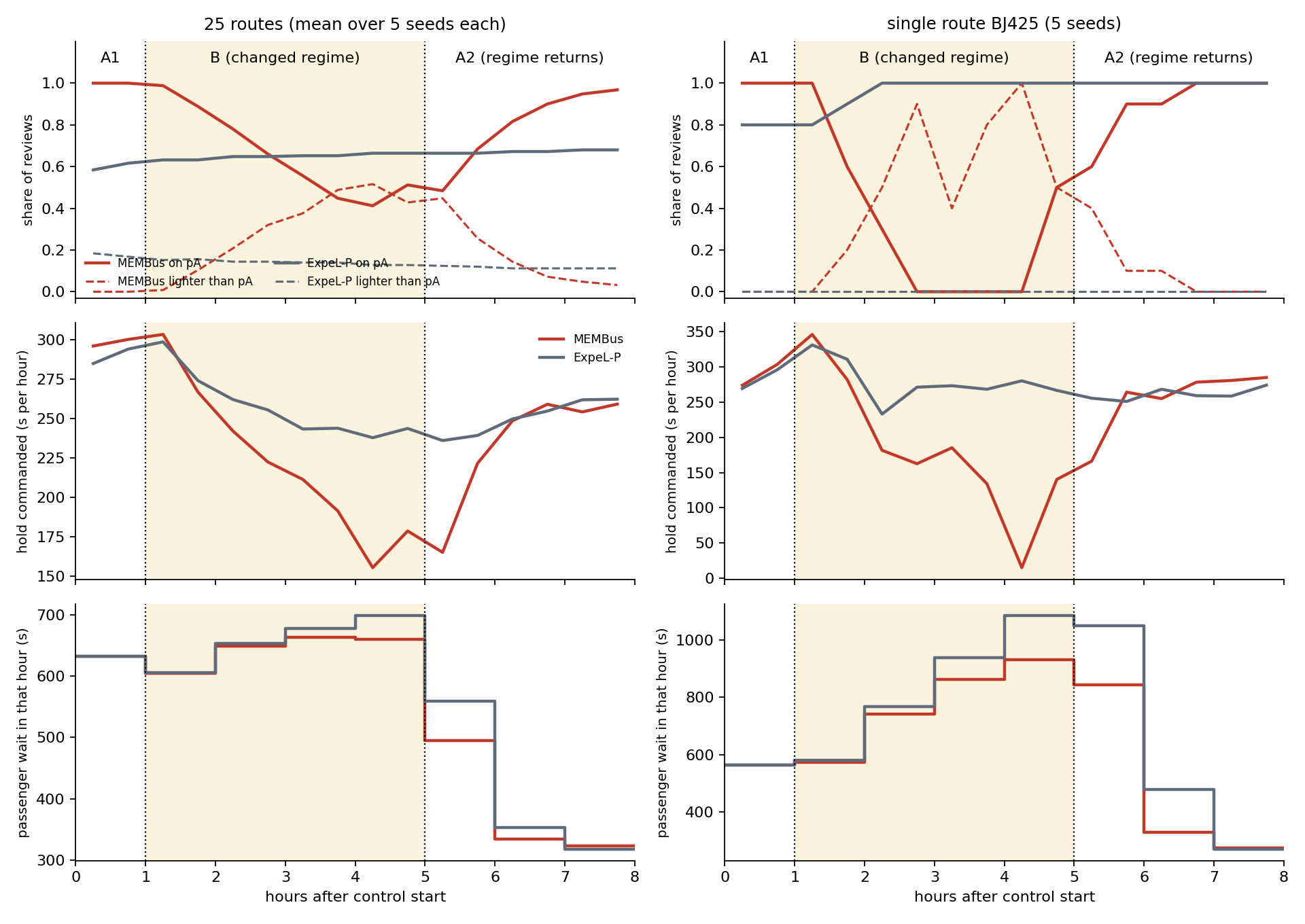}
    \caption{Regime trajectory across $\mathcal W_{\mathrm{A1}}$ (1\,h), $\mathcal W_{\mathrm{B}}$ (4\,h), and $\mathcal W_{\mathrm{A2}}$ (3\,h) over 25 routes $\times$ 5 seeds (left) and route BJ425 (right). Top: share of reviews on the stored program (pA, solid) and on lighter-holding programs (dashed). Middle: commanded holding (s/h). Bottom: hourly passenger waiting (s). Shaded: regime B; dotted lines: phase boundaries; MEMBus in red, ExpeL-style in gray.}
    \label{fig:regime}
  \end{figure}

  \textbf{Regime-resolved behavior.} Across the 25-route lifecycle experiment, MEMBus leaves the stored program after 2.47\,h in B versus 3.75\,h for the ExpeL-style agent (20 eligible routes, $p=0.0048$, Holm 0.0059) and returns to it in $\mathcal W_{\mathrm{A2}}$ in 98\% of runs that had left it versus 12\% for ExpeL-style (11 paired routes, $p=0.002$, Holm 0.006; unconditional A2 share 0.80 vs.\ 0.67, $p=0.16$). During $\mathcal W_{\mathrm{A1}}$, both agents achieve indistinguishable waiting ($+$0.07\%, $p=0.22$) while MEMBus commands 3.0\% more holding ($p=0.053$; $+$5.7\% on the 16 screened routes, Holm 0.0015), selecting the stored program in 100\% of A1 reviews versus 60\% for ExpeL-style. After the shift into $\mathcal W_{\mathrm{B}}$, MEMBus transitions toward lighter-holding programs (31\% vs.\ 14\% of B reviews), reducing holding with a lag ($-$0.4\% in hour 1 of B, $-$17\% and $-$31\% in hours 4 and 5, $-$19\% in the first hour of A2) and lowering phase-B waiting by 2.0\% across 25 routes (Holm 0.029; $-$4.6\% vs.\ a stale-program control on the 16 screened routes, 15/16). Figure~\ref{fig:regime} illustrates these phase dynamics.

  \subsubsection{\texorpdfstring{High-Saturation Regime-Shift Case Study}{High-Saturation Regime-Shift Case Study}}
  \label{app:limitedcase}

  \textbf{Target-seed evaluation on high-saturation disturbances.} Two route--disturbance configurations that met strict harm-below-KEEP criteria on development seeds were evaluated on five target seeds with KEEP, the stale controller, the hindsight oracle, a static ExpeL-style rule, and the full scripted learner (7-program menu). On CD99/A40h (40\% fleet withdrawal in B), 0/5 target seeds crossed the strict harm-vs.-KEEP threshold (harm ratio mean 0.89, range 0.85--1.01), whereas on 484/CL5 (saturation level 5), 1/5 seeds crossed it (mean 1.03, range 1.00--1.07; stale program 3\% and 8\% worse than KEEP in B and A2). Across all 10 seeds of both configurations, the full learner executed suspension, probing, and reactivation, reducing waiting by 20\% vs.\ the stale program on 484/CL5 (0.818 vs.\ 1.018 of KEEP; ExpeL-style 0.984, oracle 0.634) and by 7\% on CD99/A40h (0.722 vs.\ 0.774).

  \subsubsection{\texorpdfstring{Replication on Three Open-Weight Models}{Replication on Three Open-Weight Models}}
  \label{app:replic}

  \textbf{Lifecycle replication on three open-weight backbones} (16 routes $\times$ 3 seeds per model, 288 cells, 0 failed calls; full MEMBus minus ExpeL-style agent, route-level exact two-sided sign-flip tests, Holm-adjusted over recovery phase, whole day, and phase B within each model). Phase $\mathcal W_{\mathrm{A1}}$ is indistinguishable ($\pm 0.2$\,s), while phase $\mathcal W_{\mathrm{B}}$ ($p\in[3.7\times10^{-4},3.1\times10^{-3}]$), recovery phase $\mathcal W_{\mathrm{A2}}$, and whole-day waiting improve consistently across all three models:

  \begin{center}\scriptsize
  \setlength{\tabcolsep}{3pt}
  \begin{tabular}{@{}lrrrrrr@{}}
  \toprule
  Model & Recovery phase & W/T/L & Holm & Whole day & W/T/L & Holm \\
  \midrule
  glm-5.3-flash & $-$72.5\,s ($-$13.1\%) & 15/0/1 & $1.8\times10^{-4}$ & $-$59.5\,s ($-$8.0\%) & 15/0/1 & $2.4\times10^{-4}$ \\
  nemotron-3.5-super-vl & $-$60.4\,s ($-$11.3\%) & 15/0/1 & $2.7\times10^{-3}$ & $-$46.0\,s ($-$6.1\%) & 15/0/1 & $2.7\times10^{-3}$ \\
  nemotron-3-ultra-550b & $-$51.3\,s ($-$9.6\%) & 13/0/3 & $1.3\times10^{-3}$ & $-$45.3\,s ($-$6.0\%) & 13/0/3 & $1.3\times10^{-3}$ \\
  gpt-6-luna (same routes and seeds) & $-$59.7\,s ($-$10.2\%) & 13/0/3 & $p=2.9\times10^{-3}$ & $-$47.9\,s ($-$6.0\%) & 12/0/4 & $p=1.7\times10^{-3}$ \\
  \bottomrule
  \end{tabular}
  \end{center}
  These results confirm that the lifecycle recovery gain generalizes across diverse open-weight LLM backbones.

  \subsubsection{\texorpdfstring{Division of Labor and Cross-Replay of Program Sequences}{Division of Labor and Cross-Replay of Program Sequences}}
  \label{app:divlabor}

  \textbf{Program-selection behavior.} Across the 640 ladder runs (8 arms $\times$ 16 routes $\times$ 5 seeds $\times$ 32 reviews), among runs that left the stored program by the end of B, full MEMBus returns to it in $\mathcal W_{\mathrm{A2}}$ in 94\% (15 routes) versus 14\% for ExpeL-style (7 routes; 6 paired routes, $p=0.0625$), and leaves a stale program in B after 2.15\,h versus 4.00\,h (12/12 eligible routes, Holm 0.0015), with zero false suspensions in A1 and 133 reactivation events across 70 of 80 cells. In $\mathcal W_{\mathrm{A2}}$, the LLM departs from the memory layer's recommendation in only 11.1\% of reviews (107/960; 54/80 runs never depart), achieving similar gains over ExpeL-style in zero-departure runs ($-$58.4\,s) and departing runs ($-$49.7\,s; within-route $-$62.1 vs.\ $-$50.0\,s); on the three open-weight models, recommendation-following rates in $\mathcal W_{\mathrm{A2}}$ range from 59\% to 92\%.

  \textbf{Cross-replay of program sequences} (16 ladder routes $\times$ 5 seeds, 80 cells, deterministic zero-API replay; 80/80 update-only replays and spliced prefixes match bit-exactly). Relative to the ExpeL-style recovery-phase baseline wait of 450.7\,s, a Shapley decomposition ordering phase-B updates, phase-B suspension/probing, and returned-phase program choices yields:

  \begin{center}\scriptsize
  \setlength{\tabcolsep}{3pt}
  \resizebox{\linewidth}{!}{%
  \begin{tabular}{@{}>{\raggedright\arraybackslash}p{0.31\linewidth}>{\raggedright\arraybackslash}p{0.14\linewidth}>{\raggedright\arraybackslash}p{0.09\linewidth}>{\raggedright\arraybackslash}p{0.13\linewidth}>{\raggedright\arraybackslash}p{0.11\linewidth}>{\raggedright\arraybackslash}p{0.14\linewidth}@{}}
  \toprule
  Component of full system $-$ ExpeL-style & Effect (s) & Share & 95\% CI & Routes (better/worse) & $p$ (Holm) \\
  \midrule
  Total (full system $-$ ExpeL-style) & $-$55.6 ($-$12.3\%, ratio of totals) & 100\% & [$-$89.4, $-$27.8] & 16/0 & $<10^{-4}$ \\
  Evidence updates in phase B & $-$32.1 & 58\% & [$-$51.7, $-$14.7] & 13/3 & 0.0026 (0.0051) \\
  Suspension and probing in phase B & $-$6.2 & 11\% & [$-$13.5, $+$0.7] & 9/6 & 0.12 (0.12) \\
  Program choices in the returned phase & $-$17.3 & 31\% & [$-$28.3, $-$8.2] & 15/1 & 0.0002 (0.0006) \\
  \multicolumn{6}{@{}l}{\textit{Descriptive, uncorrected}} \\
  Update-only $-$ ExpeL-style / full $-$ update-only & $-$25.8 / $-$29.8 & 46\% / 54\% & & 11/5; 12/3 & 0.0525 / 0.0011 \\
  Path with phase B first: update / suspend, probe / returned-phase choices & $-$33.7 / $-$7.4 / $-$14.4 & 61\% / 13\% / 26\% & & & 0.0024 / 0.11 / 0.0007 \\
  Path with the returned phase first: returned-phase choices & $-$21.1 & 38\% & [$-$34.6, $-$10.3] & 14/2 & 0.0006 \\
  Whole day, same decomposition: update / suspend, probe / choices & $-$26.3 / $-$1.1 / $-$6.3 & 78\% / 3\% / 19\% & & & 0.0354 / 0.79 / 0.0005 (Holm) \\
  \bottomrule
  \end{tabular}}
  \end{center}

  Percentages here are ratios of totals ($-55.6/450.7=-12.3\%$), whereas $-9.31\%$ in Table~\ref{tab:ladder} is the mean of route-relative differences. Within $\mathcal W_{\mathrm{A2}}$, the update-only sequence is $+8.5$\,s slower than ExpeL-style ($p=0.08$), whereas full MEMBus is $-23.0$\,s faster than update-only (15/0, $p=0.0001$), confirming that suspension-enabled reactivation drives the returned-phase choice component (26\%--38\% of total recovery gain alongside 58\%--61\% from cleaner phase-B state carryover).

  \textbf{Composition of the returned phase (ticks 60--95, share of ticks).}

  \begin{center}\scriptsize
  \begin{tabular}{@{}lrrrr@{}}
  \toprule
  Arm & Original program & Phase-B program & KEEP & Other \\
  \midrule
  ExpeL-style & 68.5\% & 6.2\% & 0.0\% & 25.2\% \\
  Update only & 47.8\% & 7.5\% & 19.7\% & 25.0\% \\
  Full system & 78.0\% & 2.4\% & 10.7\% & 8.9\% \\
  \bottomrule
  \end{tabular}
  \end{center}

  Full MEMBus selects the original program 9.5 percentage points more often than ExpeL-style (CI [$-$12.3, $+$33.4], $p=0.45$, n.s.). ExpeL-style is bimodal across routes (80--100\% on 11 routes, 0--20\% on 5 routes: 484, BJ337, BJ402, BJ58, BJ556), yet the returned-phase gain is concentrated on the 11 routes where ExpeL-style stays on the original program ($-$23.8\,s, total $-$77.6\,s vs.\ $-$2.9\,s on the other 5), where full MEMBus reduces mistimed alternative programs from 25.2\% to 8.9\% and interposes brief KEEP transitions (10.7\%).

  \subsubsection{\texorpdfstring{Open-Loop Replay, Harmful Windows, and the Closed Loop}{Open-Loop Replay, Harmful Windows, and the Closed Loop}}
  \label{app:openloop}

  \textbf{Open-loop window replay versus closed-loop backlog.} In an open-loop replay of spliced backlog-free windows where a window is labeled harmful when the stored program waits at least 5\% longer than KEEP (median $+11.7\%$; 16.6\% of pooled windows), full MEMBus lowers the harmful-reuse rate to 50\% versus 74\% and 73\% for ExpeL-style and DiLu-style, waiting $+7.3\%$ above KEEP in harmful windows versus $+10.6\%$ for ExpeL-style/DiLu-style and $+14.9\%$ for pure stale reuse (15 routes, $p=0.0001$). Whereas spliced open-loop windows omit queue carryover (giving 0.818 vs.\ 0.870 of KEEP over 13 isolated windows under 2.25 vs.\ 1.03 recovery probing windows), closed-loop operation propagates phase-B queue backlogs into $\mathcal W_{\mathrm{A2}}$, so curtailing stale holding in $\mathcal W_{\mathrm{B}}$ yields net closed-loop waiting reductions.

  \subsubsection{\texorpdfstring{Online Learner and Monotone Load Drift}{Online Learner and Monotone Load Drift}}
  \label{app:drift}

  \textbf{Scripted online learner.} Across 17 routes (16 informative), the scripted learner waits 18.8\% less than stale cards (16/0/0, Holm $6{\times}10^{-5}$), 13.5\% less under a demand-level shift (17/0/0), and 16.0\% less than frozen experience under 7-day monotone load drift (8 routes $\times$ 3 seeds, 8/0/0).

  \textbf{Multi-day monotone load drift with the LLM} (8 routes $\times$ 3 seeds $\times$ 7 days, 504 cells, 0 failures; memory persisted across days; mean 7-day wait in s: no control 1725.6, frozen 1867.4, fixed program 1687.4, ExpeL-style append-only 1852.7, LLM update-only 1792.3, LLM full 1781.4; scripted update-only 1514.6, scripted full 1568.2, oracle 1482.1). Two-sided route-level sign-flip tests (with 8 routes, smallest attainable $p=0.0078$, so Holm over the 6-test family has floor $0.0469$):

  \begin{center}\scriptsize
  \begin{tabular}{@{}lrrrr@{}}
  \toprule
  Contrast & Difference (s) & Relative & W/T/L & $p$ (Holm) \\
  \midrule
  LLM update-only $-$ frozen & $-$75.1 & $-$4.0\% & 8/0/0 & 0.0078 (0.0469) \\
  LLM full $-$ frozen & $-$86.1 & $-$4.6\% & 8/0/0 & 0.0078 (0.0469) \\
  LLM update-only $-$ ExpeL-style & $-$60.4 & $-$3.3\% & 8/0/0 & 0.0078 (0.0469) \\
  LLM full $-$ ExpeL-style & $-$71.4 & $-$3.9\% & 8/0/0 & 0.0078 (0.0469) \\
  LLM full $-$ LLM update-only & $-$10.9 & $-$0.6\% & 5/0/3 & 0.40 (0.40) \\
  LLM update-only $-$ scripted update-only & $+$277.7 & $+$18.3\% & 0/0/8 & 0.0078 (0.0469) \\
  \multicolumn{5}{@{}l}{\textit{Secondary, uncorrected}} \\
  LLM full $-$ scripted full & $+$213.2 & $+$13.6\% & 0/0/8 & 0.0078 \\
  LLM update-only $-$ no control & $+$66.7 & $+$3.9\% & 3/0/5 & 0.18 \\
  LLM full $-$ no control & $+$55.8 & $+$3.2\% & 4/0/4 & 0.27 \\
  LLM update-only $-$ fixed program & $+$104.9 & $+$6.2\% & 1/0/7 & 0.023 \\
  LLM full $-$ fixed program & $+$94.0 & $+$5.6\% & 2/0/6 & 0.031 \\
  \bottomrule
  \end{tabular}
  \end{center}

  Under 7-day monotone overload drift, managed LLM memory reduces waiting by 3.3\%--4.6\% relative to frozen and append-only memory (8/0/0, Holm 0.0469) and beats KEEP on SH50, CD306, and CD99 ($+$13.6\%--18.3\% relative to its scripted learner counterpart overall). Outside probe-locked reviews, the LLM's recommendation-following rate moves from 1.00 on day 1 to 0.97, 0.91, 0.55, 0.11, 0.16, and 0.02 across days 2--7 for update-only, and to 1.00, 0.88, 0.65, 0.20, 0.23, and 0.23 for full MEMBus, concentrating the gap to the scripted learner on overloaded days 4--7 ($+162, +566, +591$, and $+631$\,s; $+18.6\%$--$25.2\%$ on days 5--7, 8/8 routes, unadjusted $p=0.0078$). Prompt length rises by only 1.7\%--2.7\% over the seven days ($\approx$24,000 tokens per review, within-day adherence correlation $-0.08$ to $-0.17$) while the single-card update-only arm also reaches 0.02, indicating that the LLM favors static development holding records over online KEEP recommendations during severe overload. Aligning LLM trust with online overload recommendations remains an open direction for long-horizon operational agents.

  \subsubsection{\texorpdfstring{Holding}{Holding}}
  \label{app:holding}

  \textbf{Holding volume decomposition.} Across the 32 lifecycle routes (25 unscreened plus 7 confirmation), MEMBus commands 8\% less total holding than ExpeL-style (10,077 vs.\ 11,002\,s; on the 10 routes with manageable headroom where MEMBus holds less, it waits 68.7\,s less, 10/0/0, $p=0.0020$, versus $+8.3$\,s, 2/11 better, on the 11 routes where it holds more; in $\mathcal W_{\mathrm{A1}}$ MEMBus holds 2.8\% more with identical waiting), whereas in steady state on 17 routes it holds slightly more (14,320 vs.\ 13,978\,s). Open-loop holding-equalization replays on the 11 lifecycle routes where MEMBus holds less ($\times$5 seeds) show that uniformly or randomly thinning ExpeL-style's holds to match MEMBus's total leaves ExpeL-style 4.9\% (Holm 0.008) or 4.0\% (Holm 0.028) slower over the day and 12.9\% / 8.4\% slower in $\mathcal W_{\mathrm{A2}}$ (Holm 0.002). Thinning ExpeL-style only in $\mathcal W_{\mathrm{B}}$ recovers 68\% of the whole-day gap while leaving MEMBus $-2.5\%$ better over the day (10/11, Holm 0.016) and $-10.0\%$ better in $\mathcal W_{\mathrm{A2}}$ (11/11, Holm 0.002); conversely, scaling MEMBus's holding up to ExpeL-style's total preserves a $-9.0\%$ recovery-phase advantage (10/11, Holm 0.002). Thus, reduced holding during $\mathcal W_{\mathrm{B}}$ mediates most of the phase-B advantage (across all 16 routes, unconditioned random thinning yields whole-day $p=0.136$), while timely reactivation governs $\mathcal W_{\mathrm{A2}}$ recovery at matched holding volume. Holding less indiscriminately is harmful: KEEP waits 438\,s longer than MEMBus (30/32 routes) and \texttt{balance\_cap15} waits 391\,s longer (31/32 routes).

  \subsection{\texorpdfstring{Running-Clock Execution}{Running-Clock Execution}}
  \label{app:async}

  Whereas the baseline protocol pauses the simulator during LLM inference, we re-evaluated the main 8-route $\times$ 5-seed comparison under a running simulation clock (360 cells, 0 failures): each request observes the state at call initiation, the active program continues executing at 300\,s ticks meanwhile, and the returned selection takes effect at the first tick at or after request time plus measured wall-clock latency (1 wall-clock second $= 1$ simulated second). Measured delays had median 48--55\,s, p90 158--170\,s, and p99 335--410\,s (including shared-gateway queueing over an 8--12\,s median HTTP service time), with 2.1--2.6\% of calls exceeding 300\,s and 34\% of ticks continuing the prior program. Under the running clock, MEMBus waits 437.9\,s versus 424.9\,s synchronously ($+3.05\%$, CI $[0.0,+6.8]$, 1/1/6, $p=0.070$, Holm 0.50; $+3.1\%$ under $2\times\text{delay}+60$\,s stress), ExpeL-style shifts by $+2.4\%$ ($+2.2\%$ stress), and DiLu by $-0.2\%$, preserving the exact controller ordering $\text{MEMBus} < \text{ExpeL-style} < \text{ReAct-style} < \text{DiLu} < \text{KEEP}$. Detailed breakdowns are given in Appendix~\ref{app:async-data}.

  \subsection{\texorpdfstring{Offline Reinforcement-Learning Baselines}{Offline Reinforcement-Learning Baselines}}
  \label{app:rl}

  \textbf{Offline-trained RL suite.} Alongside DQN and CQL (MLP 12--64--64--7, $\gamma=0.9$, soft target update 0.01, CQL $\alpha=1.0$, Adam $5\times10^{-4}$, batch 256, 0.29--0.50M transitions, $4.1\times10^{6}$--$1.4\times10^{7}$ gradient steps), which appear in Table~\ref{tab:baseline-comparison}, we trained DDPG, SAC, MADDPG, and CAAC offline directly on the development seeds of the 8 main-table timetable-grounded routes (MLP 256--256, Adam $3\times10^{-4}$, $\tau=0.005$, $\gamma=0.9$, 200k steps; checkpoints selected solely on development seeds; 160 target runs, all completed), complementing the frozen transferred policies evaluated on the auxiliary 10-route subset in Appendix~\ref{app:supporting-evaluations}.

  \begin{center}\scriptsize
  \setlength{\tabcolsep}{4pt}
  \begin{tabular}{@{}lrrrrrrr@{}}
  \toprule
  & DDPG & SAC & MADDPG & CAAC & DQN & CQL & MEMBus \\
  \midrule
  8\,h wait (s) & 433.4 & 459.5 & 422.7 & 461.3 & 472.6 & 460.8 & 424.9 \\
  Baseline $-$ MEMBus (s) & $+$8.5 & $+$34.6 & $-$2.2 & $+$36.4 & $+$47.7 & $+$35.9 & --- \\
  $p$ (unadjusted) & 0.56 & 0.047 & 0.76 & 0.078 & 0.10 & 0.0078 & --- \\
  \bottomrule
  \end{tabular}
  \end{center}

  In stationary 8\,h operation on the 8 main routes, MEMBus matches offline MADDPG ($-2.2$\,s, $p=0.76$) and DDPG ($+8.5$\,s, $p=0.56$), achieves lower waiting than CQL ($+35.9$\,s, $p=0.0078$) and SAC ($+34.6$\,s, unadjusted $p=0.047$), and has lower point-estimate waiting than CAAC and DQN ($+36.4$\,s and $+47.7$\,s, n.s.). On a separate 40-cell evaluation with disjoint seeds, a scripted variant without dispatch programs waits 8\%--25\% less than each of the six learned policies (Holm $p=0.0469$, the floor attainable with 8 routes; evaluated as a separate arm without cross-seed subtraction).

  \subsection{\texorpdfstring{Voyager-Style and MemGPT-Style Baselines}{Voyager-Style and MemGPT-Style Baselines}}
  \label{app:vgmg}

  For transit control, our Voyager-style baseline adapts curriculum and critic selection over route programs, and our MemGPT-style baseline manages tiered memory buffers over structured fleet states. Against the strengthened Voyager-style agent (with critic and curriculum), MEMBus waits 3.53\% less across 25 routes (484.8 vs.\ 502.5\,s, 17/4/4, $p=0.0094$, Holm 0.0283) and 18.5\% less than an unguided Voyager-style variant (22/1/2, Holm $1.3{\times}10^{-5}$); on 8 routes the differences are $-$4.3\% (6/2/0) and $-$14.3\%, while on \texttt{glm-5.3-flash} MEMBus is equivalent within TOST $\pm 2\%$ to the initial Voyager-style variant and statistically level ($-1.2\%$, n.s.) with the strengthened variant (which is itself 15.5\% better than the initial variant and 2.7\% slower than ExpeL-style, unadjusted $p=0.018$). For MemGPT-style at the per-bus interface, MEMBus waits 60.2\% less on \texttt{gpt-6-luna} (8/0/0) and 46.5\% less on \texttt{glm-5.3-flash}; enforcing a final-decision-only last stage (MG2) yields $-$60.2\% and $-$51.4\% on two additional backbones (8/0/0 each, $p=0.0078$, with 16/40 and 27/40 cells encountering model format fallbacks), matching per-bus ReAct.

  \subsection{\texorpdfstring{Further Models}{Further Models}}
  \label{app:xmodel}

  \begin{table}[!htbp]
    \centering
    \small
    \caption{MEMBus against ReAct and DiLu across seven LLM backbones on the 8 main routes at $h=8$.}
    \label{tab:crossmodel-main}
    \setlength{\tabcolsep}{4pt}
    \begin{tabular}{@{}lrrrr@{}}
      \toprule
      Model & ReAct (s) & DiLu (s) & MEMBus $-$ ReAct & MEMBus $-$ DiLu \\
      \midrule
      gpt-6-luna (main) & 1072.1 & 1068.3 & $-$60.4\% (8/0/0) & $-$60.2\% (8/0/0) \\
      Nemotron Bolt & 1077.6 & 1066.0 & $-$59.9\% (7/0/0) & $-$60.1\% (8/0/0) \\
      MiniMax M3.1 & 1052.6 & 1032.4 & $-$59.6\% (8/0/0) & $-$58.8\% (8/0/0) \\
      Nemotron Super VL & 1060.4 & 1066.2 & $-$59.9\% (8/0/0) & $-$60.1\% (8/0/0) \\
      Step 5 & 1143.7 & 1086.3 & $-$62.8\% (8/0/0) & $-$60.9\% (8/0/0) \\
      DeepSeek V4.1 & 867.6 & 889.0 & $-$51.0\% (8/0/0) & $-$52.2\% (8/0/0) \\
      GLM 5.3 & 772.6 & 846.4 & $-$45.0\% (8/0/0) & $-$49.8\% (8/0/0) \\
      \bottomrule
    \end{tabular}
    \par\smallskip{\footnotesize MEMBus waits 424.9\,s on every backbone (each followed every recommendation); Nemotron Bolt's ReAct covers 7 of the 8 routes (35 cells due to context overflow on one route). Figure~\ref{fig:heatmap} and Tables~\ref{tab:x17}--\ref{tab:x17core} evaluate the 17-route confirmation set across seven models (\texttt{gpt-6-luna}, \texttt{glm-5.3-flash}, \texttt{nemotron-3.5-super-vl}, \texttt{nemotron-3-ultra-550b}, \texttt{laguna-xs-2.1}, \texttt{nemotron-3.5-bolt-e4b}, and \texttt{nemotron-3-nano-omni}), whereas MiniMax M3.1, Step 5, DeepSeek V4.1, and GLM 5.3 are evaluated on the 8 main routes (see Appendix~\ref{app:s11-disclosures} for model pre-check criteria).}
  \end{table}

  \begin{table}[!htbp]
    \centering
    \caption{MEMBus $-$ baseline at $h=8$ on 17 confirmation routes $\times$ 5 seeds within each model (original-interface harnesses and same-interface agents).}
    \label{tab:x17}
    \scriptsize
    \setlength{\tabcolsep}{3pt}
    \begin{tabular}{@{}lrrrrr@{\hspace{8pt}}rrr@{}}
      \toprule
      & \multicolumn{5}{c}{Original interface} & \multicolumn{3}{c}{Same interface (-P)} \\
      Model & ReAct & DiLu & ExpeL & RAG & Reflexion & ExpeL-P & RAG-P & ReAct-P \\
      \midrule
      glm-5.3-flash & $-$39.8 & $-$48.6 & $-$17.8 & $-$16.3 (16/1/0) & $-$57.0 & $+$0.5 & $-$6.6 & $-$17.3 \\
      nemotron-3.5-super-vl & $-$59.1 & $-$59.3 & $-$51.0 & $-$59.1 & $-$58.4 & $-$0.4 & $-$17.6 & $-$52.6 \\
      nemotron-3-ultra-550b & $-$54.6 & $-$59.7 & $-$41.2 & $-$46.9 & $-$56.3 & $+$0.4 & $-$16.9 & $-$22.9 \\
      laguna, bolt, nano-omni & \multicolumn{8}{c}{core subset, see Table~\ref{tab:x17core}} \\
      \bottomrule
    \end{tabular}
    \par\smallskip{\footnotesize Relative differences (\%) as MEMBus $-$ baseline in waiting (negative: MEMBus waits less); 17/0/0 unless shown; Holm over 5 tests within model $7.6{\times}10^{-5}$ (smallest attainable $p=1.5{\times}10^{-5}$); 2,635 evaluated cells (with schema-invalid replies in original-interface harnesses defaulting to KEEP in 446 cells). When the recommendation is calibrated, MEMBus follows it in 100\% of reviews on \texttt{glm}, \texttt{super-vl}, and \texttt{ultra}, yielding identical MEMBus outcomes across all three (512.9\,s, CV 0.444, served 98.73\%). Under a corrupted-recommendation stress test (replacing the recommendation with the lightest-holding program while keeping card evidence genuine), 5 of the 7 evaluated backbones override the bad recommendation in 85\%--100\% of reviews (\texttt{gpt-6-luna} 91--95\%, reducing waiting by $-$54.0\% on 8 routes, Holm 0.031, and $-$50.6\% on 17 routes, Holm $10^{-4}$; \texttt{glm} $-$58.3\%, \texttt{super-vl} $-$57.8\%, \texttt{ultra} $-$47.9\%, 17/0/0, Holm $3.05\times10^{-5}$), whereas \texttt{bolt} (5.1\%) and \texttt{nano-omni} (0\%) copy the recommendation.}
  \end{table}

  \begin{table}[!htbp]
    \centering
    \caption{MEMBus $-$ baseline at $h=8$ on the 17-route $\times$ 3-seed core subset for three smaller/specialized backbones.}
    \label{tab:x17core}
    \scriptsize
    \setlength{\tabcolsep}{3pt}
    \resizebox{\linewidth}{!}{%
    \begin{tabular}{@{}lrrrrr@{}}
      \toprule
      Model & ReAct & DiLu & ExpeL-P & Follows rec.\ & Bad-recommendation override \\
      \midrule
      laguna-xs-2.1 & $-$57.5 & $-$53.4 & $+$7.1 (5/0/12, n.s.) & 0.32 & 84.6\% \\
      nemotron-3.5-bolt-e4b & $-$59.9 (16/0/0) & $-$59.7 & $-$57.5 & high & 5.1\% \\
      nemotron-3-nano-omni & $-$59.6 & $-$59.7 & $-$38.4 & 1.00 & 0\% \\
      \bottomrule
    \end{tabular}}
    \par\smallskip{\footnotesize Relative differences (\%) in waiting; 17/0/0 unless shown; Holm over 3 tests within each model ($4.6{\times}10^{-5}$; \texttt{laguna} vs.\ ExpeL-P Holm 0.114). On \texttt{bolt}, ReAct uses 16 of 17 routes (48 pairs, $p=3.1{\times}10^{-5}$) due to context overflow on one route. ``Follows rec.'' denotes the share of reviews following a calibrated recommendation: models that adhere closely to compiled recommendations (\texttt{bolt}, \texttt{nano-omni}) achieve large gains over raw-record ExpeL-P (17/0/0) while exhibiting lower override rates under corrupted recommendations, whereas \texttt{laguna} overrides frequently (0.32 adherence) and performs comparably to ExpeL-P.}
  \end{table}

  \subsubsection{\texorpdfstring{Cross-Model Comparison}{Cross-Model Comparison}}
  \label{app:xmodel-fig}

  Figure~\ref{fig:xmodel-main} summarizes the steady-state contrasts of Tables~\ref{tab:x17} and~\ref{tab:x17core} across all seven backbones (panel a) alongside the lifecycle replication of Appendix~\ref{app:replic} (panel b). In steady state, MEMBus reduces waiting by 40--60\% relative to ReAct and 48--60\% relative to DiLu across every model, matches ExpeL-P on strong reasoning models, and substantially outperforms ExpeL-P on smaller models (\texttt{bolt}, \texttt{nano-omni}) that benefit directly from compiled card recommendations. In the recurring-regime lifecycle setting, MEMBus reduces recovery-phase waiting by 9.6--13.1\% relative to ExpeL-P across \texttt{glm-5.3-flash}, \texttt{nemotron-3.5-super-vl}, and \texttt{nemotron-3-ultra-550b}, matching the $-10.2\%$ improvement observed on \texttt{gpt-6-luna}. On the single-seed diagnostic sweep of Figure~\ref{fig:heatmap} for \texttt{laguna}, \texttt{bolt}, and \texttt{nano-omni}, two cells exceeding the 5\% call-failure threshold (\texttt{laguna} RAG at 11.5\% and \texttt{nano-omni} Reflexion on 16/17 routes) are shaded gray.

  \textbf{Multi-metric weighting sensitivity.} Across all 286 convex combinations of the five evaluation metrics (waiting, headway CV, served fraction, token usage, and bunching events on a 0.1-step simplex grid), MEMBus ranks first among the original-interface LLM controllers (ReAct, DiLu, ExpeL, RAG, and Reflexion) across all 17 confirmation routes and three evaluated backbones (token-free rule and RL baselines are compared per metric in Table~\ref{tab:permetric}).

  \subsection{\texorpdfstring{Simulator Realism and Robustness}{Simulator Realism and Robustness}}
  \label{app:realism}

  This subsection evaluates supply alignment against published route catalogs and mobile-app timetables (queried 2 November 2026) alongside controller sensitivity under time-varying dispatch and perturbed simulation dynamics.

  \subsubsection{\texorpdfstring{Supply Fidelity}{Supply Fidelity}}
  We compared simulated free-running loop times against mobile-app route-planner travel times across the 25 timetable-grounded routes (8 main and 17 confirmation; Table~\ref{tab:realism-fid}). Planner in-vehicle estimates exclude terminal layover and provide a geometric and free-flow speed reference. SH50 and CD50 use construction-speed planner estimates (marked $^{*}$, excluded from summary quantiles), and CD61 lacks a standalone planner query.

  \begin{table}[!htbp]
    \centering
    \caption{Simulated supply against the App timetable (direction A$\to$B).}
    \label{tab:realism-fid}
    \scriptsize
    \setlength{\tabcolsep}{2.5pt}
    \begin{tabular}{@{}lrrrrrrrr@{}}
      \toprule
      Route & $H_{\mathrm{am}}$ & Loop $L$ & Planned & $L$/pl. & Buses & $\lceil\mathrm{pl.}/H_{\mathrm{am}}\rceil$ & Stops & Length km \\
      \midrule
      \multicolumn{9}{@{}l}{\textit{Main table (8 routes)}} \\
      598 & 10.9 & 44.6 & 48 & 0.93 & 4 & 5 & 29/28 & 14.9/14.9 \\
      546 & 7.5 & 46.9 & 50 & 0.94 & 6 & 7 & 30/29 & 16.5/16.5 \\
      484 & 13.3 & 34.1 & 39 & 0.87 & 3 & 3 & 21/20 & 13.6/13.5 \\
      SH50 & 8.6 & 47.1 & 56$^{*}$ & 0.84 & 6 & 7 & 26/--- & 18.5/18.3 \\
      CD306 & 13.3 & 49.8 & 51 & 0.98 & 4 & 4 & 31/30 & 16.5/16.5 \\
      CD50 & 6.0 & 64.6 & 69$^{*}$ & 0.94 & 11 & 12 & 40/--- & 22.5/22.6 \\
      CD99 & 13.3 & 38.4 & 40 & 0.96 & 3 & 3 & 21/20 & 13.0/13.1 \\
      K12 & 5.7 & 54.1 & 61 & 0.89 & 9 & 11 & 27/27 & 24.1/24.2 \\
      \midrule
      \multicolumn{9}{@{}l}{\textit{Confirmation routes (17)}} \\
      BJ2 & 7.1 & 44.0 & 45 & 0.98 & 6 & 7 & 25/24 & 15.0/15.1 \\
      BJ337 & 7.5 & 38.8 & 41 & 0.95 & 5 & 6 & 22/21 & 13.9/14.1 \\
      BJ342 & 12.0 & 46.4 & 50 & 0.93 & 4 & 5 & 28/29 & 16.5/16.5 \\
      BJ386 & 15.0 & 68.9 & 74 & 0.93 & 5 & 5 & 41/40 & 24.8/24.8 \\
      BJ402 & 12.0 & 49.4 & 54 & 0.91 & 4 & 5 & 29/28 & 18.1/18.1 \\
      BJ425 & 8.6 & 51.5 & 57 & 0.90 & 6 & 7 & 32/31 & 19.1/19.0 \\
      BJ55 & 13.3 & 24.4 & 25 & 0.98 & 2 & 2 & 16/15 & 8.2/8.2 \\
      BJ556 & 8.6 & 33.8 & 33 & 1.03 & 4 & 4 & 20/19 & 10.7/10.8 \\
      BJ58 & 8.6 & 31.5 & 37 & 0.85 & 4 & 5 & 17/16 & 13.5/13.3 \\
      BJ597 & 8.0 & 51.8 & 53 & 0.98 & 6 & 7 & 32/31 & 17.6/17.6 \\
      BJ663 & 9.2 & 53.2 & 58 & 0.92 & 6 & 7 & 32/31 & 19.5/19.4 \\
      BJ671 & 9.2 & 62.8 & 75 & 0.84 & 7 & 9 & 31/30 & 26.3/27.2 \\
      BJ820 & 10.9 & 85.2 & 103 & 0.83 & 8 & 10 & 39/38 & 38.2/38.1 \\
      BJ9 & 10.9 & 51.3 & 57 & 0.90 & 5 & 6 & 27/26 & 20.2/20.4 \\
      BJ997 & 15.0 & 77.7 & 91 & 0.85 & 5 & 7 & 43/42 & 32.3/32.8 \\
      CD56 & 6.0 & 63.6 & 67 & 0.95 & 11 & 12 & 40/38 & 21.0/21.1 \\
      CD61 & 10.9 & 63.0 & --- & --- & 6 & --- & 40/--- & 21.4/21.6 \\
      \bottomrule
    \end{tabular}
    \par\smallskip{\footnotesize $H_{\mathrm{am}}$: 07:00--09:00 headway in minutes ($120\text{ min}/\text{departures}$ A$\to$B). Simulated loop $L$: KEEP loop time (min) on the one-direction ring. $^{*}$: construction-speed estimate, excluded from loop-ratio quantiles. Catalog stop counts exceed planner ``stops passed'' by 1 because planner counts omit the origin terminal.}
  \end{table}

  Relative to planner trip times (Table~\ref{tab:realism-fid}), simulated one-way loop times are a median 7.2\% shorter across the 22 routes with independent planner values (IQR $-$11.0\% to $-$4.3\%, max $-$17.2\%), and catalog road lengths match published route lengths within a median $-$0.1\% (24 of 25 routes within 1\%; max deviation $-$3.4\% on BJ671). Because midday (09:00--16:00) timetable headways exceed AM-peak headways by a median factor of 1.35 (range 0.83 on BJ55 to 1.89 on CD50; evening-to-AM median ratio 1.14), holding $H_{\mathrm{am}}$ constant over an 8\,h horizon sustains peak-density supply and demand throughout midday, motivating the time-varying timetable check below.

  \subsubsection{\texorpdfstring{Timetable-Driven Variant of the Testbed}{Timetable-Driven Variant of the Testbed}}
  We implemented an optional time-varying timetable mode in which fleet size and passenger arrival rate track the empirical 06:30--14:30 timetable in 30-minute bins (simulation $t=0$ mapped to 06:00; bin headway $H_{\mathrm{bin}}$ computed over a 2-hour sliding window; demand scaled by $\mathrm{clip}(H_{\mathrm{am}}/H_{\mathrm{bin}},0.35,1.0)$ times the scenario multiplier; active fleet $\mathrm{clip}(\mathrm{round}(n_{\mathrm{am}}H_{\mathrm{am}}/H_{\mathrm{bin}}),2,n_{\mathrm{am}})$ adjusted via terminal withdrawals/insertions starting 60\,s after $t_0$, averaging $0.78\times$ the constant-AM-peak baseline). Across 16 routes $\times$ 3 seeds (672 cells, deterministic scripted arms: KEEP, stale program, static cards, and full scripted MEMBus; route-level exact sign-flip tests, Holm over 3 contrasts), the lifecycle recovery and whole-day gains remain significant (Table~\ref{tab:realism-tt}).

  \begin{table}[!htbp]
    \centering
    \caption{Timetable-driven variant: scripted MEMBus against the stale program and KEEP, relative waiting difference (\%), 16 routes $\times$ 3 seeds.}
    \label{tab:realism-tt}
    \scriptsize
    \setlength{\tabcolsep}{3pt}
    \resizebox{\linewidth}{!}{%
    \begin{tabular}{@{}lllrrrrr@{}}
      \toprule
      Testbed & Scenario & Window & vs.\ stale (= static cards) & vs.\ KEEP & KEEP (s) & Stale (s) & MEMBus (s) \\
      \midrule
      Timetable-driven & Lifecycle & whole day & $-$7.5 [$-$10.4, $-$4.7], 13/1/2, Holm 0.0006 & $-$25.8, Holm $9\times10^{-5}$ & 1096.0 & 900.7 & 821.0 \\
      Timetable-driven & Lifecycle & A2 & $-$13.1 [$-$18.5, $-$7.8], 13/1/2, Holm 0.0012 & $-$32.3, Holm 0.0005 & --- & --- & --- \\
      Constant-$H_{\mathrm{am}}$ & Lifecycle & whole day & $-$9.0 [$-$11.3, $-$6.7], 15/1/0, Holm 0.0001 & $-$40.1, Holm $9\times10^{-5}$ & 1032.9 & 684.0 & 618.4 \\
      Constant-$H_{\mathrm{am}}$ & Lifecycle & A2 & $-$14.7 [$-$19.9, $-$9.9], 15/1/0, Holm 0.0001 & $-$60.8, Holm $9\times10^{-5}$ & --- & --- & --- \\
      Timetable-driven & Stationary & whole day & $+$0.1 [0.0, $+$0.4] & $-$48.1, Holm $9\times10^{-5}$ & 1040.5 & 531.3 & 532.2 \\
      Constant-$H_{\mathrm{am}}$ & Stationary & whole day & 0.0 (identical) & $-$62.0, Holm $9\times10^{-5}$ & 1077.1 & 406.9 & 406.9 \\
      \bottomrule
    \end{tabular}}
    \par\smallskip{\footnotesize W/T/L counts routes where scripted MEMBus is better / tied / worse than the stale program (in scripted runs without an LLM, static cards coincide bit-exactly with the stale program).}
  \end{table}

  In Table~\ref{tab:realism-tt}, scripted MEMBus reduces lifecycle whole-day waiting by 7.5\% (Holm 0.0006, better in 39 of 48 route--seed blocks) and $\mathcal W_{\mathrm{A2}}$ waiting by 13.1\% (Holm 0.0012) under time-varying fleet and demand, closely matching the constant-$H_{\mathrm{am}}$ testbed ($-$9.0\% and $-$14.7\%; between-testbed difference $p=0.12$, 11 of 16 routes less negative, n.s.). Per-route lifecycle differences against the stale program in the timetable-driven variant (\%): 484 $0.0$, 546 $-16.9$, 598 $-9.7$, BJ2 $-9.8$, BJ337 $-7.9$, BJ342 $-6.8$, BJ402 $+0.7$, BJ425 $-17.0$, BJ556 $-6.1$, BJ58 $-2.9$, BJ597 $-15.8$, BJ663 $+0.8$, BJ671 $-8.4$, CD306 $-1.5$, CD56 $-7.7$, and K12 $-11.5$. In stationary operation without regime shifts, scripted MEMBus and static cards coincide as expected, while both outperform KEEP by 48.1\%--62.0\%. \textbf{Withdrawn passengers.} In this variant, withdrawing off-peak buses removes passengers currently onboard (averaging 173 passengers per cell, roughly 1\% of total boardings, excluded thereafter) and re-inserts vehicles at the origin terminal.

  \subsubsection{\texorpdfstring{Robustness to Simulator Assumptions}{Robustness to Simulator Assumptions}}
  \label{app:robust}
  We further evaluated sensitivity to uncalibrated simulator assumptions on the 16 ladder routes ($\times$2 target seeds $\times$ 4 scripted arms $\times$ 7 settings $= 896$ runs, all completed) under a recurring-regime schedule where A1 runs for 1\,h after $t_0$, B runs until absolute simulation time $18{,}000$\,s (rather than relative $t_0+18{,}000$\,s in Eq.~\eqref{eq:regime-windows}), and A2 spans the remaining ${\approx}4.1$\,h of the 8\,h control horizon. Programs and cards remain frozen from the unperturbed development seeds, while evaluation-day dynamics undergo one perturbation at a time: (a) per-passenger boarding/alighting/crowding dwell scaled by $\times 0.7$ or $\times 1.5$; (b) dwell jitter doubled ($0.2\to 0.4$); (c) link running-time noise added ($\mathrm{CV}=0.10$); (d) uniform stop weights replaced by skewed TransitLM origin--destination stop marginals (log SD 1.1--2.1, mean-normalized); and (e) deterministic fluid arrivals replaced by compound Poisson batch arrivals (geometric batch mean 3, identical mean rate).

  \begin{table}[!htbp]
    \centering
    \caption{Scripted MEMBus against stale program, static ExpeL-style policy, and KEEP under perturbed simulator assumptions: relative waiting difference (\%, route mean; W/T/L; exact two-sided $p$, 16 routes $\times$ 2 seeds).}
    \label{tab:robust}
    \scriptsize
    \setlength{\tabcolsep}{2.5pt}
    \resizebox{\linewidth}{!}{%
    \begin{tabular}{@{}lllllll@{}}
      \toprule
      & \multicolumn{3}{c}{Recovery phase A2 (${\approx}4.1$\,h)} & \multicolumn{3}{c}{Whole day} \\
      \cmidrule(lr){2-4}\cmidrule(l){5-7}
      Setting & vs.\ stale & vs.\ static ExpeL-style & vs.\ KEEP & vs.\ stale & vs.\ static ExpeL-style & vs.\ KEEP \\
      \midrule
      Baseline (unperturbed) & $-$15.1 (14/0/2; 0.0004) & $-$6.9 (14/0/2; 0.0079) & $-$61.3 (16/0/0; $3{\times}10^{-5}$) & $-$8.7 (15/0/1; $6{\times}10^{-5}$) & $-$1.6 (11/0/5; 0.070) & $-$39.8 (16/0/0; $3{\times}10^{-5}$) \\
      Dwell $\times0.7$ & $-$0.1 (4/11/1; 0.63) & $+$0.8 (2/2/12; 0.34) & $-$64.6 (16/0/0; $3{\times}10^{-5}$) & $-$1.8 (5/11/0; 0.063) & $+$1.0 (5/2/9; 0.80) & $-$54.0 (16/0/0; $3{\times}10^{-5}$) \\
      Dwell $\times1.5$ & $-$9.4 (16/0/0; $3{\times}10^{-5}$) & $-$8.7 (16/0/0; $3{\times}10^{-5}$) & $+$3.1 (4/0/12; 0.010) & $-$6.3 (16/0/0; $3{\times}10^{-5}$) & $-$5.7 (16/0/0; $3{\times}10^{-5}$) & $+$3.6 (5/0/11; 0.028) \\
      Dwell jitter $\times2$ & $-$16.6 (14/1/1; $2{\times}10^{-4}$) & $-$9.0 (13/0/3; 0.0010) & $-$56.5 (16/0/0; $3{\times}10^{-5}$) & $-$9.8 (14/1/1; $1{\times}10^{-4}$) & $-$3.9 (12/0/4; 0.0021) & $-$36.0 (16/0/0; $3{\times}10^{-5}$) \\
      Running-time noise 0.10 & $-$21.0 (15/0/1; $9{\times}10^{-5}$) & $-$12.4 (14/0/2; 0.0010) & $-$56.0 (16/0/0; $3{\times}10^{-5}$) & $-$11.5 (16/0/0; $3{\times}10^{-5}$) & $-$4.6 (13/0/3; 0.0074) & $-$33.1 (16/0/0; $3{\times}10^{-5}$) \\
      TransitLM-weighted demand & $-$4.3 (15/0/1; $6{\times}10^{-5}$) & $-$2.8 (15/0/1; $2{\times}10^{-4}$) & $-$0.7 (10/0/6; 0.14) & $-$4.3 (16/0/0; $3{\times}10^{-5}$) & $-$2.4 (15/0/1; $6{\times}10^{-5}$) & $+$0.1 (8/0/8; 0.93) \\
      Clustered arrivals & $-$18.3 (16/0/0; $3{\times}10^{-5}$) & $-$10.6 (15/0/1; $1{\times}10^{-4}$) & $-$42.8 (16/0/0; $3{\times}10^{-5}$) & $-$10.0 (16/0/0; $3{\times}10^{-5}$) & $-$4.5 (13/0/3; 0.0011) & $-$28.2 (15/0/1; $6{\times}10^{-5}$) \\
      \bottomrule
    \end{tabular}}
    \par\smallskip{\footnotesize Negative values indicate lower waiting under scripted MEMBus. Mean whole-day waiting (s; MEMBus / stale / static ExpeL-style / KEEP): baseline 604.0 / 669.9 / 624.0 / 1016.1; dwell $\times0.7$ 406.1 / 415.4 / 404.6 / 904.4; dwell $\times1.5$ 2754.8 / 2930.1 / 2911.5 / 2667.4; jitter $\times2$ 670.1 / 747.7 / 704.7 / 1044.0; running-time noise 698.8 / 795.1 / 738.7 / 1050.7; TransitLM-weighted demand 3902.1 / 4071.4 / 3994.6 / 3904.4; clustered arrivals 769.3 / 855.9 / 809.9 / 1084.6.}
  \end{table}

  \textbf{Findings and boundary conditions (Table~\ref{tab:robust}).} Scripted MEMBus outperforms both the stale program and the static ExpeL-style policy with $p\le 0.05$ in 6 of 7 settings during A2 (and in 6/7 vs.\ stale, 5/7 vs.\ static over the whole day), with amplified gains under higher dwell ($\times 1.5$), link running-time noise, and clustered batch arrivals. Under lighter dwell ($\times 0.7$), bunching instability weakens so 11 of 16 routes execute identical sequences across arms (n.s.). Conversely, under severe overload perturbations (dwell $\times 1.5$ and uncalibrated high-skew TransitLM stop weights, where mean waits reach 2,755\,s and 3,902\,s), any positive holding saturates capacity relative to KEEP ($+3.6\%$ under dwell $\times 1.5$, $p=0.028$; $+0.1\%$ under TransitLM weights, $p=0.93$), so online lifecycle management reduces excess waiting relative to static holding policies under severe capacity saturation. Separate cruising-speed perturbations yielded mixed signs across route subsets ($+10.9\%$ and $-13.7\%$, both n.s.).

  \subsection{\texorpdfstring{Running-Clock Execution Results}{Running-Clock Execution Results}}
  \label{app:async-data}

  \begin{table}[!htbp]
    \centering
    \scriptsize
    \caption{Running-clock results: mean waiting at $h=8$ (s, route mean; 8 routes $\times$ 5 seeds), paused clock against running clock.}
    \label{tab:async-data}
    \setlength{\tabcolsep}{3pt}
    \begin{tabular}{@{}lrrrrrr@{}}
      \toprule
      Arm & Paused & Running & $\Delta$\% [95\% CI] & W/T/L & Holm & Stress $\Delta$\% \\
      \midrule
      MEMBus & 424.9 & 437.9 & $+$3.05 [0.0, $+$6.8] & 1/1/6 & 0.50 & $+$3.1 \\
      ExpeL-style (same interface) & 429.8 & 440.0 & $+$2.4 [$-$0.9, $+$5.9] & --- & 0.81 & $+$2.2 \\
      ReAct-style (same interface) & 676.0 & 739.0 & $+$9.3 [$-$6.3, $+$26.6] & --- & 0.61 & $-$6.9 \\
      DiLu & 1062.1 & 1060.4 & $-$0.2 & --- & 1.0 & $+$0.2 \\
      No control (KEEP) & 1066.0 & --- & --- & --- & --- & --- \\
      \bottomrule
    \end{tabular}
    \par\smallskip{\footnotesize Running clock: each answer is applied after its wall-clock delay. Stress: delay $=2\times$ wall clock $+60$\,s. Holm over 8 tests. Route-level exact two-sided sign-flip tests; W/T/L reported for MEMBus.}
  \end{table}

  Table~\ref{tab:async-data} details the running-clock comparison summarized in Appendix~\ref{app:async}: wall-clock delays had median 48--55\,s, p90 158--170\,s, p99 335--410\,s, and max 612--815\,s (HTTP service median 8--12\,s plus gateway queueing; 0 failed calls; 0.67\% HTTP 429 retries, all succeeded; 0 superseded answers under normal clock and 2--4 under stress). Synchronous MEMBus re-runs matched stored baselines bit-exactly on 40/40 cells, isolating the $+13.0$\,s shift as pure actuation latency, and Kendall's $\tau=1$ rank concordance holds across paused, running, and stress clocks.

  \paragraph{One-tick-late actuation (LAG1).} Because 97.4--97.9\% of measured wall-clock delays are under 300\,s, applying every decision deterministically at the next 300\,s tick (with tick 0 running KEEP; 0 failed calls, $d=0$ stub verified bit-identical to synchronous runs) provides a provider-independent test of one-tick actuation delay across the 25-route lifecycle benchmark and three open-weight models (Table~\ref{tab:lag1}).

  \begin{table}[!htbp]
    \centering
    \scriptsize
    \caption{One-tick-late actuation (LAG1): recovery phase $\mathcal W_{\mathrm{A2}}$ relative waiting difference (\%), with synchronous baseline in the final column.}
    \label{tab:lag1}
    \setlength{\tabcolsep}{3pt}
    \begin{tabular}{@{}lrrrr@{}}
      \toprule
      Comparison (full system $-$ reference) & One tick late \% & W/T/L & Holm & Synchronous \% \\
      \midrule
      \multicolumn{5}{@{}l}{\textit{gpt-6-luna, 25 routes $\times$ 5 seeds}} \\
      vs.\ ExpeL-style agent & $-$7.95 & 21/0/4 & $1.9{\times}10^{-4}$ & $-$5.88 \\
      vs.\ management off & $-$7.87 & 19/3/3 & $1.5{\times}10^{-5}$ & $-$6.65 \\
      \multicolumn{5}{@{}l}{\textit{whole day: vs.\ ExpeL-style $-$4.06\%, vs.\ management off $-$3.98\%}} \\
      \multicolumn{5}{@{}l}{\textit{Open-weight models, 16 routes $\times$ 3 seeds, vs.\ ExpeL-style agent}} \\
      glm-5.3-flash & $-$12.31 & 14/0/2 & $2.8{\times}10^{-4}$ & $-$13.13 \\
      nemotron-3.5-super-vl & $-$11.81 & 15/0/1 & $2.8{\times}10^{-4}$ & $-$11.30 \\
      nemotron-3-ultra-550b & $-$10.09 & 15/0/1 & $2.8{\times}10^{-4}$ & $-$9.61 \\
      \bottomrule
    \end{tabular}
    \par\smallskip{\footnotesize Route-level exact two-sided sign-flip tests; Holm over 2 tests (\texttt{gpt-6-luna}) and over the 3 open-weight models.}
  \end{table}

  As shown in Table~\ref{tab:lag1}, one-tick actuation delay causes no statistically significant degradation relative to synchronous execution (largest shift $+1.55\%$), preserving significant recovery-phase improvements across all four evaluated models.

\end{document}